\documentclass[10pt,twocolumn,letterpaper]{article}

\usepackage[pagenumbers]{cvpr}
\usepackage{graphicx, amsmath, amssymb, caption, subcaption, multirow, overpic, textpos, indentfirst}
\usepackage[table]{xcolor}
\usepackage[british, american]{babel}
\definecolor{cvprblue}{rgb}{0.21,0.49,0.74}
\usepackage[breaklinks,colorlinks,allcolors=cvprblue]{hyperref}
\usepackage{placeins}
\usepackage{tabularx}
\usepackage{afterpage}
\usepackage{caption}
\usepackage{xcolor}
\usepackage{pifont}
\usepackage{microtype}
\usepackage{algorithm}
\usepackage{algpseudocode}
\usepackage{listings}
\usepackage{overpic}
\usepackage{wrapfig}
\usepackage{colortbl}
\usepackage{tabularx}
\usepackage{verbatim}
\usepackage{nicefrac}
\usepackage{empheq} 

\def\paperID{***}
\def\confName{CVPR}
\def\confYear{2026}

\newcolumntype{x}[1]{>{\centering\arraybackslash}p{#1pt}}
\newcolumntype{y}[1]{>{\raggedright\arraybackslash}p{#1pt}}
\newcolumntype{z}[1]{>{\raggedleft\arraybackslash}p{#1pt}}

\newcommand{\authorskip}{\hspace{1.35mm}}

\newcommand{\g}[1]{\textcolor{gray}{#1}}

\input{cmd}

\begin{document}
\title{
\Large Bidirectional Normalizing Flow: From Data to Noise and Back}
\author{
 Yiyang Lu$^{1,2,*,\dagger,\ddagger}$
 \authorskip Qiao Sun$^{1,*,\dagger}$
 \authorskip Xianbang Wang$^{1,*}$
 \authorskip Zhicheng Jiang$^1$
 \authorskip Hanhong Zhao$^1$
 \authorskip Kaiming He$^1$\\ [2mm]
 \normalsize $^{*}$Equal technical contribution 
 \qquad $^{\dagger}$Project lead\\ [2mm]
 $^1$MIT\qquad $^2$Tsinghua University
\vspace{-4mm}
}
\maketitle
\renewcommand{\thefootnote}{\fnsymbol{footnote}}
\footnotetext[3]{Work done as an intern at MIT.}

\begin{abstract}
Normalizing Flows (NFs) have been established as a principled framework for generative modeling.
Standard NFs consist of a forward process and a reverse process: the forward process maps data to noise, while the reverse process generates samples by inverting it.
Typical NF forward transformations are constrained by explicit invertibility, ensuring that the reverse process can serve as their exact analytic inverse.
Recent developments in TARFlow and its variants have revitalized NF methods by combining Transformers and autoregressive flows, but have also exposed causal decoding as a major bottleneck.
In this work, we introduce Bidirectional Normalizing Flow (\textbf{BiFlow}), a framework that removes the need for an exact analytic inverse. 
BiFlow learns a reverse model that approximates the underlying noise-to-data inverse mapping, enabling more flexible loss functions and architectures.
Experiments on ImageNet demonstrate that BiFlow, compared to its causal decoding counterpart, improves generation quality while accelerating sampling by up to two orders of magnitude.
BiFlow yields state-of-the-art results among NF-based methods and competitive performance among single-evaluation (``1-NFE'') methods.
Following recent encouraging progress on NFs, we hope our work will draw further attention to this classical paradigm.
\end{abstract}


\vspace{-22pt}
\section{Introduction}
\label{sec:intro}

\begin{figure}
    \centering
    \begin{subfigure}[t]{0.49\linewidth}
        \centering
        \includegraphics[width=\linewidth,clip,page=1]{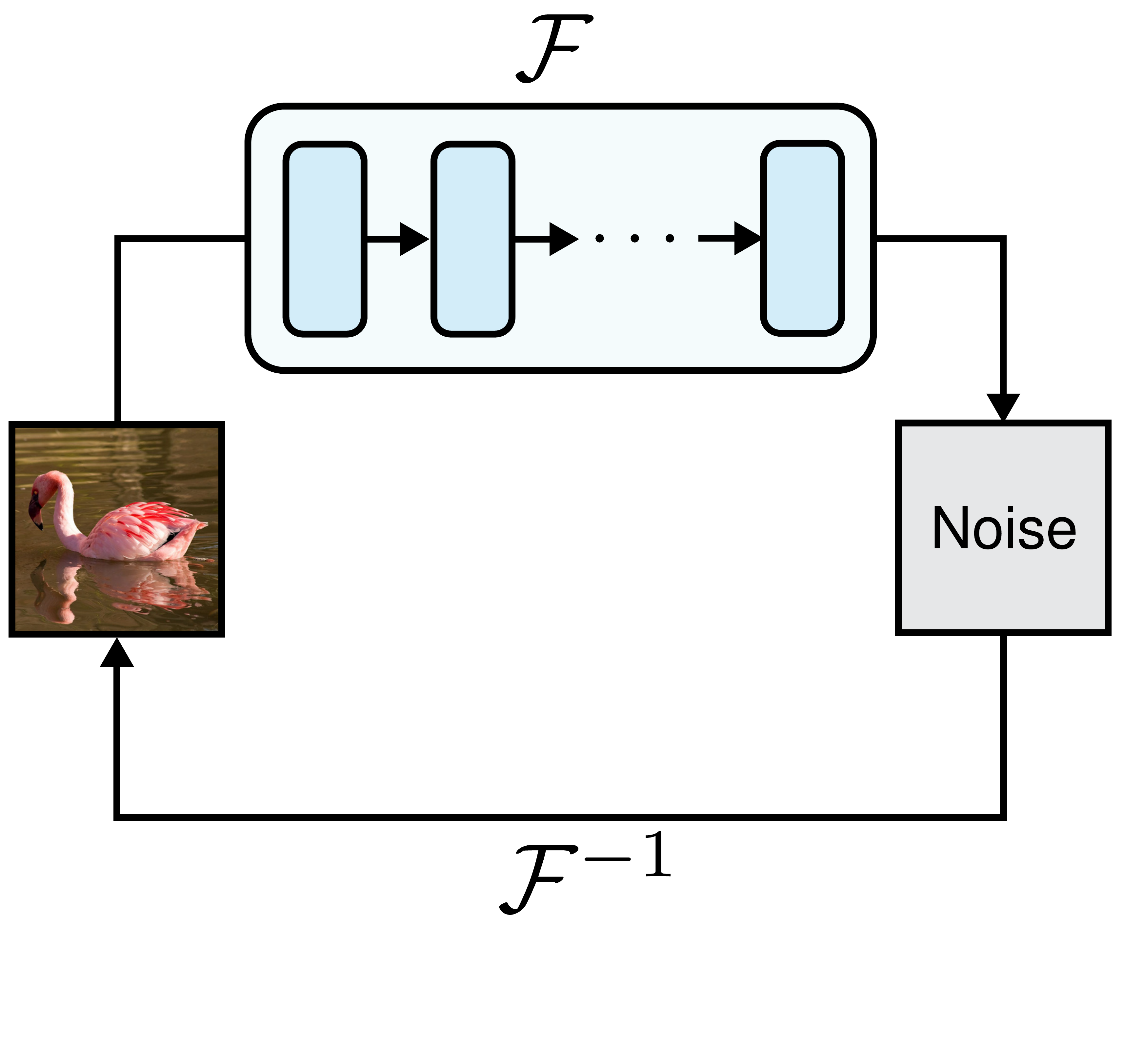}
        \caption{\centering Standard Normalizing Flow:\\explicit inverse.}
        \label{fig:teaser_nf}
    \end{subfigure}
    \begin{subfigure}[t]{0.49\linewidth}
        \centering
        \includegraphics[width=\linewidth,clip,page=2]{imgs/teaser.pdf}
        \caption{\centering Bidirectional Normalizing Flow:\\\textbf{\textit{learned}} inverse.}
        \label{fig:teaser_biflow}
    \end{subfigure}
    \vspace{-0.5em}
    \caption{
        Conceptual comparison between standard Normalizing Flows and our proposed Bidirectional Normalizing Flow (BiFlow). Instead of constraining the forward model $\mcal F$ to be explicitly invertible and using its \textit{exact analytic inverse} for generation, BiFlow introduces a learnable reverse model $\mcal G$ that approximates this inverse through our hidden alignment objective. This design frees BiFlow from architectural constraints and enables flexible loss design, allowing for efficient generation with improved quality in a single forward pass.
    }
    \vspace{-1.0em}
    \label{fig:teaser}
\end{figure}

Normalizing Flows (NFs) are a long-standing family of generative models~\cite{nf,nice,arflow}. 
They contain two processes: a \textit{forward process} that learns to transform data into noise, and a \textit{reverse process} that generates samples by inverting this transformation. 
A notable property of NFs is that the underlying flow trajectories from data to noise are \textit{learned} rather than imposed. 
This differs from their modern continuous-time counterparts~\cite{neuralode}, such as Flow Matching (FM)~\cite{fm,recitifiedflow, stochasticflow}, whose ground-truth trajectories are predetermined via time-scheduling. 
However, this advantage of NFs comes at the cost of increased learning difficulty, typically leading to more demanding constraints on forward architectures and objective formulations.

The standard NF paradigm~\cite{nf,nice} requires the reverse process to be the exact analytic inverse of the forward process (\cref{fig:teaser_nf}). This requirement restricts the range of forward model architectures that can be employed, as the model must be \textit{explicitly invertible} and its Jacobian determinant must be computable, tractable, and differentiable. Existing work on NFs~\cite{nf,nice,householder,arflow,maf,glow} have largely focused on designing compound forward functions that satisfy these requirements. Despite these diverse attempts, NF-based methods remain limited in their ability to use powerful, general-purpose architectures (\eg, U-Nets~\cite{unet} or Vision Transformers~\cite{vit}), in contrast to many modern generative model families.

\begin{figure}[t]
    \centering
    \vspace{-0.5em}
    \includegraphics[width=0.98\linewidth]{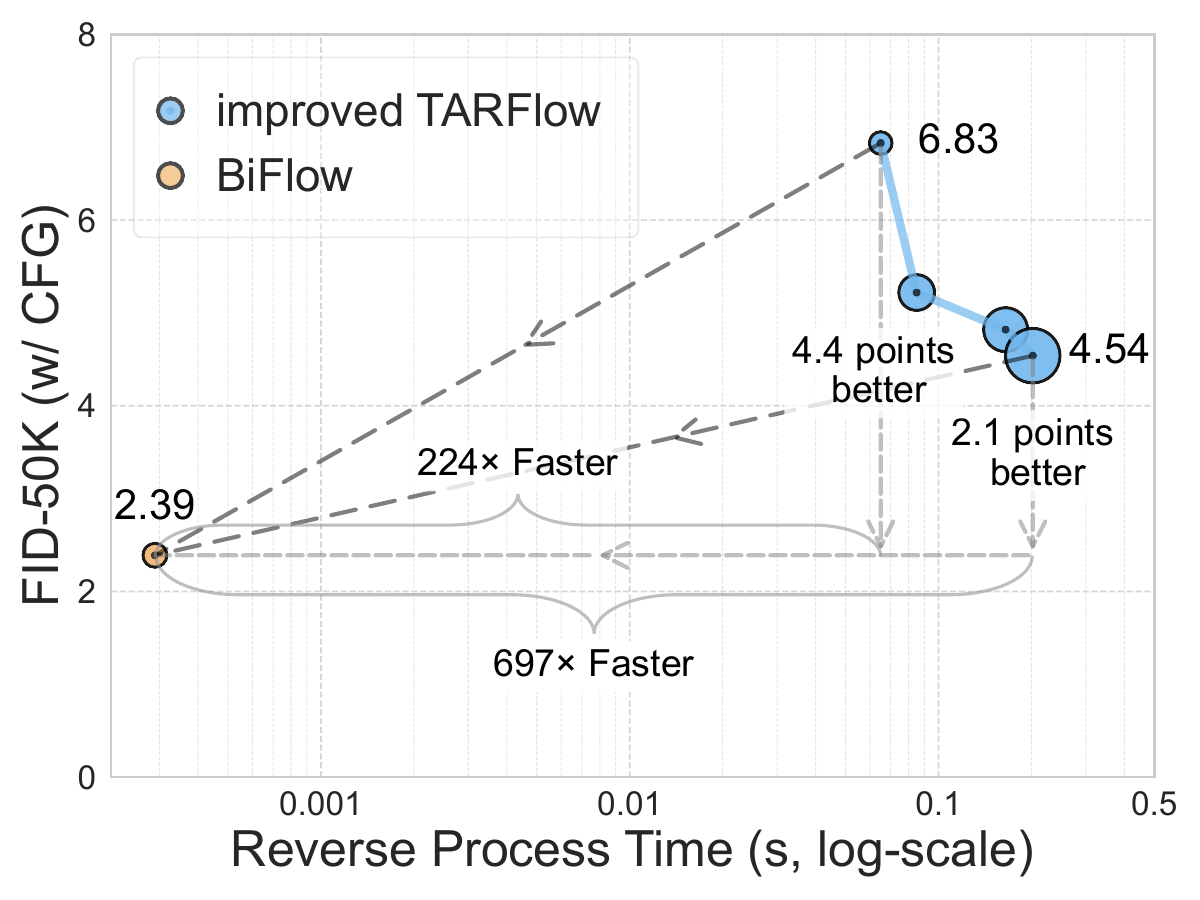}
    \vspace{-1em}
    \caption{BiFlow surpasses our improved TARFlow baseline by a wide margin in generation quality, despite using a base-size model versus an extra-large model, and it achieves markedly faster sampling as well. The $x$-axis denotes the wall-clock time (log scale) for generating one image on 8 v4 TPU cores. VAE decoding is omitted from this figure; comprehensive inference cost comparison appears in \cref{tab:biflow_vs_itarflow}.}
    \label{fig:biflow_vs_itarflow}
    \vspace{-1em}
\end{figure}

Recently, the gap between NFs and other generative models has been largely closed by TARFlow~\cite{tarflow} and its extensions~\cite{starflow}.
TARFlow has effectively integrated Transformers \cite{attention} with autoregressive flows \cite{arflow,maf} into the NF paradigm.
This design allows NF methods to benefit from the powerful Transformers, substantially mitigating a major limitation of traditional NFs.
However, to maintain computable and tractable Jacobian determinants, TARFlow decomposes the forward process into a long chain (\eg, thousands of steps) of autoregressive operations. The resulting explicit inverse therefore requires a large number of causal steps at inference time, which is difficult to parallelize.
This design not only slows down sampling, but also retains the undesirable architectural constraints during inference, \eg, the reverse model cannot perform feedforward, non-causal attention.

In this work, we introduce \textit{Bidirectional Normalizing Flow} (\textbf{BiFlow}), a framework in which both the forward and reverse processes are learned.
In our framework, the designs of the forward and reverse processes are \textit{decoupled}: the forward process can be any NF model $\mathcal F_\theta$ that is computable, tractable, and easy to learn (\eg, an improved TARFlow), while the reverse process learns a separate model $\mathcal G_\phi$ to approximate its inverse (\cref{fig:teaser_biflow}).
In contrast to the explicit inverse, our reverse model is highly flexible: it can be a feedforward, non-causal Transformer that is both expressive and efficient to run, naturally enabling high-quality, single function evaluation (1-NFE) generation.

Learning the reverse model $\mathcal G_\phi$ is not merely a form of distillation, even though we use a pre-trained forward model $\mathcal F_\theta$: in fact, our learned reverse model $\mathcal G_\phi$ can \textit{outperform} the explicit inverse of $\mathcal F_\theta$. Compared to distilling the noise-to-data trajectories, we find that \textit{aligning} the intermediate hidden states yields results even better than the explicit inverse. In addition, our learnable reverse model can naturally eliminate the extra step of score-based denoising in TARFlow, simplifying and accelerating inference while improving quality. Such a ``what-you-see-is-what-you-get'' property further enables the use of \textit{perceptual} loss~\cite{lpips}, which is impossible or difficult to leverage with an explicit inverse. Putting these factors together, our learned reverse model can substantially outperform its explicit-inverse counterpart.

We report competitive results on the ImageNet 256$\times$256 generation. Comparing with an improved TARFlow (which is also the forward model for BiFlow), BiFlow achieves an FID of 2.39 using a DiT-B size~\cite{dit} model, while being two orders of magnitude faster (see \cref{fig:biflow_vs_itarflow}; detailed in \cref{tab:biflow_vs_itarflow}).
This not only sets a new state-of-the-art result among NF-based methods, but also represents a strong 1-NFE result in comparison with other generative model families.

Following the progress established by TARFlow and extensions, our work on BiFlow further unleashes the potential of NFs as a strong competitor among modern generative model families. 
Our findings indicate that the NF principle of learning the forward trajectories, rather than pre-scheduling them, can be advantageous and need not introduce inference-time limitations. 
Considering that modern Flow Matching methods are continuous-time NFs with pre-scheduled trajectories, we hope our study will shed light on the potential synergy among these related methods.

\section{Related Work}\label{sec:related}

\paragraph{Normalizing Flows.}
Normalizing Flows (NFs) have long served as a principled framework for probabilistic generative modeling. Over the past decade, extensive research has focused on enhancing the expressivity and scalability of NFs under the constraint of invertible transformations. Planar flows~\cite{nf} and NICE~\cite{nice} pioneered the use of simple reversible mappings to construct deep generative models. Real NVP~\cite{realnvp} and Glow~\cite{glow} extended this framework with non-volume-preserving transformations and convolutional architectures. IAF~\cite{arflow} and MAF~\cite{maf} introduced autoregressive flows to improve expressivity while maintaining tractable likelihoods. TARFlow~\cite{tarflow} and STARFlow~\cite{starflow} further revitalized the NF family by incorporating Transformer into autoregressive flows. They demonstrated significant gains in generation quality and scalability, reaffirming NFs as a competitive paradigm in modern generative modeling.

Despite these advances, standard NFs still inherit limitations from their \textit{invertibility} requirement. In particular, autoregressive flow formulations impose strict causal ordering and sequential dependencies, which constrain architectural design and lead to slow inference.

\paragraph{Continuous Normalizing Flows.}
Continuous Normalizing Flows (CNFs)~\cite{ffjord,rnode,steer} generalize discrete flows by modeling transformations as continuous-time dynamics governed by ordinary differential equations (ODEs) \cite{neuralode}. CNFs enable more flexible architectures and tractable likelihood computation via numerical ODE simulations. FM \cite{fm,recitifiedflow,sd3} reformulates the explicit maximum-likelihood training objective into an equivalent implicit objective.
Diffusion models \cite{ddpm,ddim,adm} can be interpreted as a special case of Flow Matching with stochastic dynamics, achieving impressive fidelity and scalability. Despite their empirical success, the implicit formulation of FM and diffusion models sacrifices the learnable bidirectional mapping that characterizes NFs.

\section{Background: Normalizing Flows}
\label{sec:background}

\begin{figure}[t]
\centering
\vspace{-1em}
\includegraphics[width=0.98\linewidth]{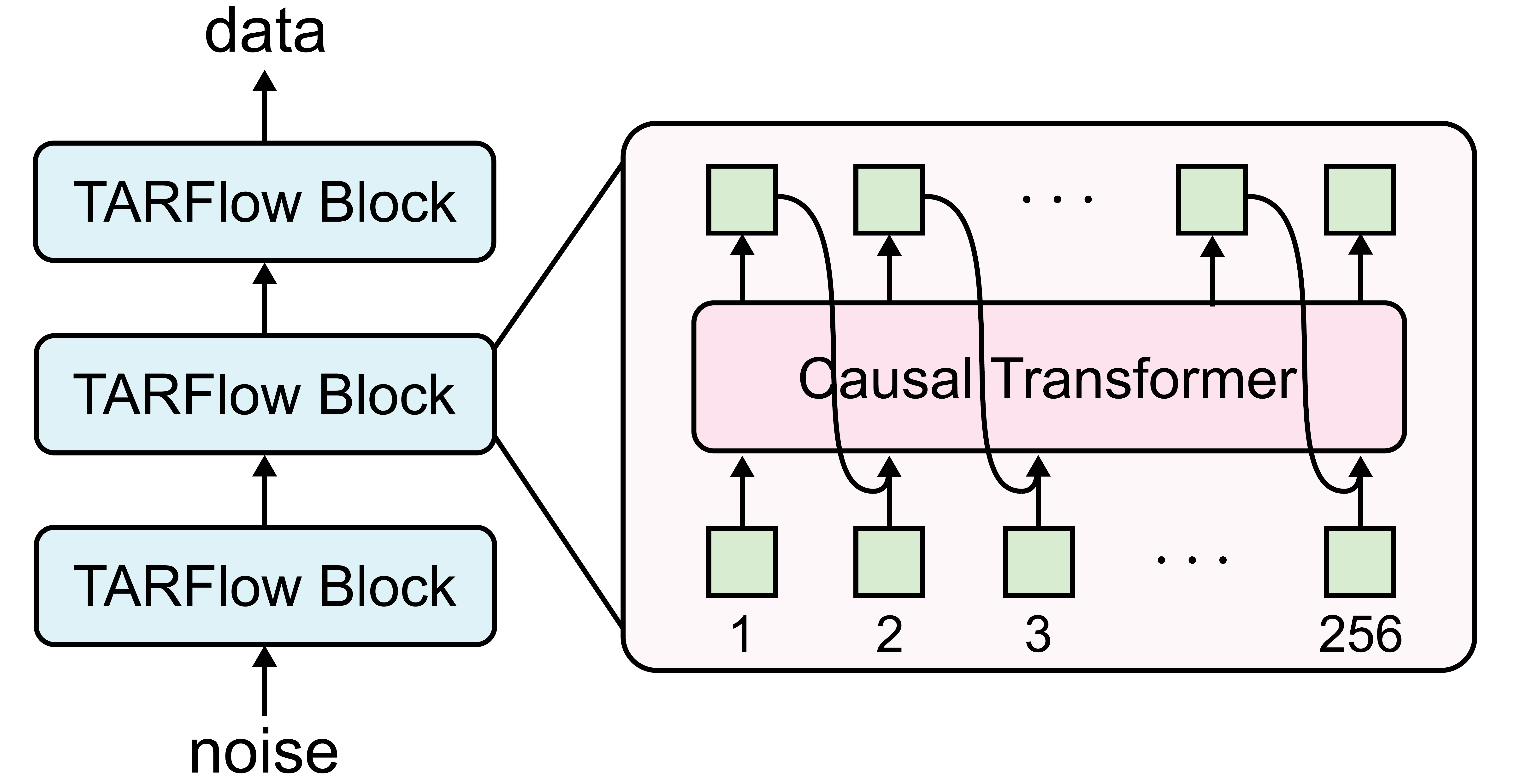}
\vspace{-0.5em}
\caption{Illustration on the \textbf{autoregressive inference} process of TARFlow. In each block, each token is transformed one by one, depending on previous tokens. This is repeated for a sequence with length 256 for a 32$\times$32 input with patch size 2, and is further repeated for all blocks (\eg, 8 blocks). Altogether, TARFlow inference requires 8$\times$256 sequential function evaluations.}
\label{fig:tarflow}
\vspace{-0.5em}
\end{figure}

Normalizing Flows (NFs) are a class of generative models that establish a bijective transformation between a Gaussian prior distribution $p_0$ and a complex data distribution $p_{\text{data}}$. 
An NF consists of a forward process and a reverse process. Given a data sample $x \in \mathbb{R}^D \sim p_{\text{data}}$, the forward process $\mathcal{F}$ maps it into the Gaussian prior space $z = \mathcal{F}(x)$.
The model assigns the data likelihood $p(x)$ through the \textit{change-of-variables} formula. Training is performed by optimizing $\mcal F$ to maximize the log-likelihood $\log p(x)$ over data samples. 

Classical NF requires the forward process $\mathcal F$ to be explicitly invertible for exact likelihood computation and efficient sampling.
Once trained, its exact inverse, $\mathcal F^{-1}$, can be used for generation by transforming Gaussian noise back to the data space, \ie, $x=\mathcal F^{-1}(z)$ where $z\sim p_0$.

In practice, to enhance expressiveness, the forward process is commonly constructed as a composition of multiple simpler bijective transformations $\mcal F := f_{B-1} \circ \cdots \circ f_1\circ f_0$ ($\circ$ denotes function composition). Under this formulation, the log-likelihood objective becomes
\begin{align}
\label{eq:change_of_variables}
\log p(x)=\log p_0(z) + \sum_{i}\log \left|\det\dfrac{\partial f_i(x^i)}{\partial x^i}\right|,
\end{align}
with $x^0=x$ and $x^{i+1}=f_i(x^{i})$. Here, $\det(\cdot)$ denotes the determinant operator.
Designing transformations that yield \textit{computable} and \textit{differentiable} determinant has been a key consideration in prior NF formulations.
This requirement motivates specialized designs such as affine coupling \cite{nice,realnvp} and autoregressive flows \cite{arflow,maf}, which preserve tractable Jacobians.

Importantly, while the log-determinant term in \cref{eq:change_of_variables} requires the forward process $\mathcal F$ to be \textit{invertible}, it does not necessitate an \textit{explicitly invertible} formulation. The explicit inverse is only required at \textit{inference} time, where we need to map samples from prior back to the data space.

\paragraph{TARFlow.}
TARFlow~\cite{tarflow} integrates Transformer architectures into autoregressive flows (AF), substantially improving their expressiveness and scalability. 
The core idea in AF is to further decompose each sub-transformation $f_i$, parameterized by a block, into $T$ steps, where $T$ denotes the sequence length of the input tokens. 
Each step transforms the $i$-th token only conditioned on its predecessors, which can naturally be realized through Transformer layers with causal masks. 
To capture bidirectional context, AF flips the sequence order in alternating blocks. By combining expressive Transformer architectures with autoregressive flows, TARFlow successfully revives NF to remain competitive with today's state-of-the-art generative models.

However, AF parameterization introduces asymmetry between training and sampling. Similar to next-token-prediction language models, although likelihood evaluation and training can be parallelized efficiently, sampling must proceed \textit{sequentially} due to the autoregressive nature, as illustrated in \cref{fig:tarflow}. 
In practice, this requires performing, \eg, thousands of (8$\times$256) inverse transformations \textit{one after another}, resulting in substantial inference latency.

\section{Bidirectional Normalizing Flow}
\label{sec:method}

\setlength{\fboxsep}{1pt}
\begin{figure*}[!t]
    \centering
    \begin{subfigure}[t]{0.32\linewidth}
        \centering
        \includegraphics[width=\linewidth,page=1]{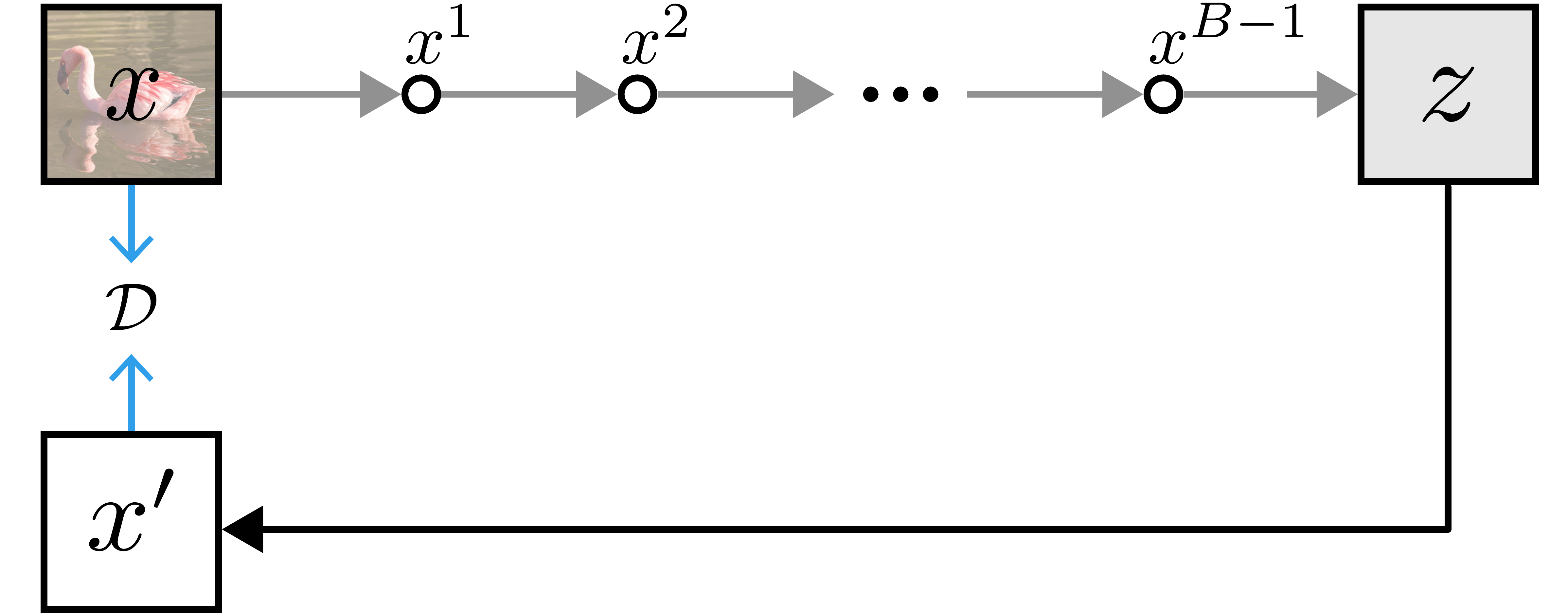}
        \caption{Naive Distillation.}
        \label{fig:naive_distill}
    \end{subfigure}
    \hfill
    \begin{subfigure}[t]{0.32\linewidth}
        \centering
        \includegraphics[width=\linewidth,page=2]{imgs/reverse.pdf}
        \caption{Hidden Distillation.}
        \label{fig:hidden_distill}
    \end{subfigure}
    \hfill
    \begin{subfigure}[t]{0.32\linewidth}
        \centering
        \includegraphics[width=\linewidth,page=3]{imgs/reverse.pdf}
        \caption{Hidden Alignment.}
        \label{fig:hidden_align}
    \end{subfigure}
    \vspace{-0.5em}
    \caption{Comparison of three approaches for learning the reverse process. Each $\boldsymbol{\circ}$ marks a position where the model returns to the same dimension as input $x$. Blue arrows with $\mcal D$ refer to distance loss terms. Our hidden alignment strategy (\cref{fig:hidden_align}) combines the strengths of \cref{fig:naive_distill} and \cref{fig:hidden_distill}, leveraging the entire trajectory for supervision without repeatedly returning to input space.}
    \vspace{-0.5em}
    \label{fig:reverse_learning_strategies}
\end{figure*}


We propose a Bidirectional Normalizing Flow (BiFlow) framework, which has:
(i) a forward model $\mathcal{F}_\theta$ that transforms data samples into pure noise, and (ii) a \emph{learnable}, separate reverse model $\mathcal{G}_\phi$ that approximates its inverse, mapping noise back to the data space.
Training is performed in two stages: first, similar to classical NF, we train the forward model using maximum likelihood estimation; then, keeping the forward model fixed, we train the reverse model to approximate its inverse mapping.

Notably, our reverse model $\mcal G_{\phi}$ is \textit{not} constrained by explicit invertibility. As a result, this allows us to design the reverse model with arbitrary architectures (\eg, bidirectional attention-based Transformers) and training objectives. 
Next, we discuss the formulation, objectives, and learning dynamics of the reverse process.

\subsection{Learning to Approximate the Inverse}
\label{subsec:reverse_learning}

Given a pre-trained forward model $\mcal F_{\theta}$, our goal is to optimize a reverse model $\mcal G_{\phi}$ that approximates its inverse. We consider three strategies: (i) naive distillation; (ii) hidden distillation; (iii) hidden alignment, as approaches to learning the reverse model. \cref{fig:reverse_learning_strategies} illustrates the differences among these methods, as we describe next.

\paragraph{Naive Distillation.}
A straightforward strategy is to impose a direct distillation loss:
\begin{align*}
    \mcal L_{\text{naive}}(x) = \mcal D\bigl(x, x'\bigr),
\end{align*}
where $x$ is a data sample, $x'=\mcal G_{\phi}(\mcal F_\theta(x))$ is the reconstructed data, and $\mcal D$ denotes a distance metric (\eg, L2 distance). The reverse model is trained to minimize the reconstruction error on data samples (see \cref{fig:naive_distill}).

This simple approach provides supervision only at the \textit{final} output, which may be insufficient for effectively training the reverse model. Directly mapping pure noise to data in one step is highly under-constrained, making it difficult for the reverse network to learn a reliable inverse from a single reconstruction loss.

\paragraph{Hidden Distillation.}
A typical NF is composed of a sequence of simple sub-transformations, \ie, $\mcal F_\theta = f_{B-1} \circ \cdots \circ f_1 \circ f_0$, where each $f_i$ is a transformation block and $B$ denotes the total number of blocks.
We can strengthen the training signal by leveraging the full sequence of intermediate states generated along the forward trajectory.

As illustrated in \cref{fig:hidden_distill}, starting from $x\sim p_{\text{data}}$, the forward model produces a trajectory of intermediate hidden states $\{x^i\}$
with $z=\mathcal F_\theta(x)$ as the final output prior. Analogously, we also design the reverse model to be composed of $B$ blocks, generating a reverse trajectory $\{h^i\}$ from $z$.
We distill the reverse model by enforcing the two trajectories to be close. Formally, the loss is defined as:
{\setlength{\belowdisplayskip}{4pt}
\begin{align*}
    \mathcal L_{\text{hidden}}(x)=\sum_{i}\, \mcal D\bigl(x^i, h^i\bigr),
\end{align*}
}where $h^0$ corresponds to the reconstructed output $x'$. Optionally, each term can be assigned a distinct weighting factor. This formulation encourages the reverse model to \mbox{invert} each sub-transformation individually, which could help guide the reverse model to invert the mapping $\mathcal{F}_\theta$ step by step. The intermediate hidden states $\{x^i\}$ serve as auxiliary supervision for learning the correspondence between $x$ and $z$.

Although this hidden distillation strategy provides more supervision than naive distillation, it introduces structural constraints on model design. Since each intermediate state $x^i$ has the same dimensionality as the input, the reverse model is forced to repeatedly project features down to the input space and then back up into the hidden space. This rigid requirement restricts architectural flexibility, ultimately limiting the model's effectiveness.

\paragraph{Hidden Alignment.}
\label{para:hidden_align} 
We propose a more flexible strategy, termed hidden alignment. Crucially, it leverages the full forward trajectory for supervision while relaxing the restrictive requirement in hidden distillation that intermediate hidden states must lie in the input space.

As shown in \cref{fig:hidden_align}, we extract intermediate hidden states $\{h^i\}$ from the reverse model $\mathcal G_\phi$. Unlike hidden distillation, which enforces each $h^i$ to directly match its input-space counterpart $x^i$, we introduce a set of learnable projection heads $\{\varphi_i\}$ to align the projected representations $\varphi_i(h^i)$ with the corresponding forward states $x^i$. The training objective then becomes:
{
\setlength{\belowdisplayskip}{4pt}
\begin{align}\label{eq:hidden_align_loss}
\mathcal L_{\text{align}} (x) = \sum_{i} \mcal D\bigl(x^i, \varphi_i(h^i)\bigr),
\end{align}
}where $h^0=x'$ and $\varphi_0$ is the identity mapping.

This simple modification allows the reverse model to benefit from full trajectory supervision while maintaining architectural and representational flexibility. By decoupling the representation space from the input token space, hidden alignment avoids the potential semantic distortion caused by repeated projections. 

\subsection{Eliminating Score-based Denoising}
\label{subsec:denoise}

\begin{figure}
    \vspace{-0.1em}
    \centering
    \begin{subfigure}[t]{0.98\linewidth}
    \includegraphics[width=0.98\linewidth,page=1]{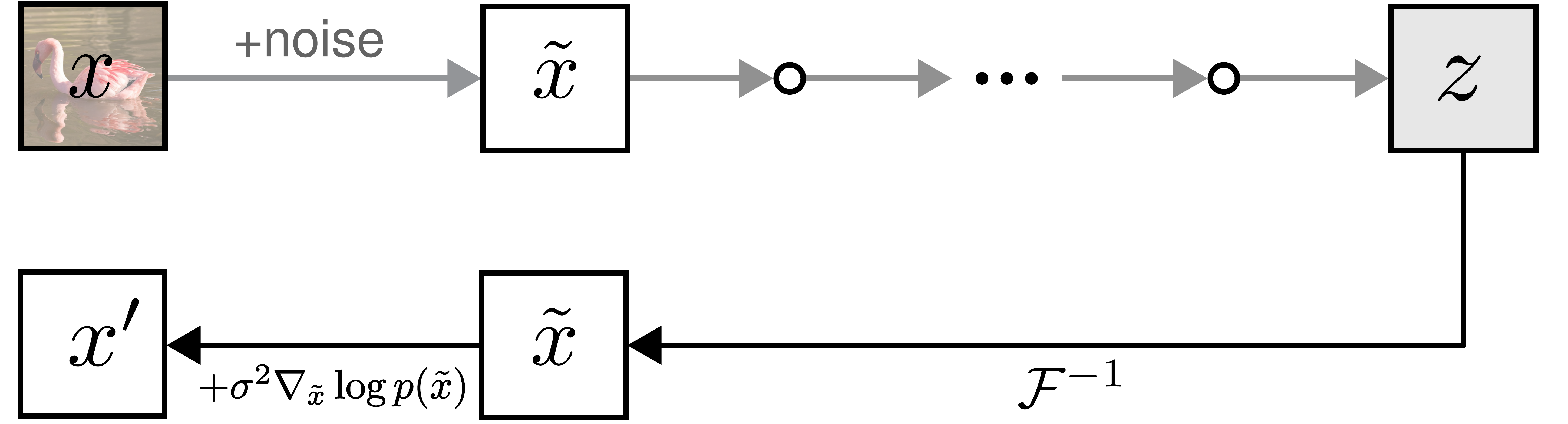}
    \vspace{0.25em}
    \caption{TARFlow: explicit denoising.}
    \label{fig:tarflow_denoising}
    \end{subfigure}
    \begin{subfigure}[t]{0.98\linewidth}
    \vspace{0.5em}
    \includegraphics[width=0.98\linewidth,page=2]{imgs/denoise.pdf}
    \caption{BiFlow: learned denoising.}
    \label{fig:biflow_denoising}
    \end{subfigure}
    \vspace{-0.5em}
    \caption{Incorporating the \textbf{denoising step} into our hidden alignment framework. The reverse model is extended with an additional block dedicated to denoising. Our learned denoising eliminates the need for calculating the score function through a whole forward-backward pass, incurring only a single additional block forward.}
    \label{fig:denoising}
    \vspace{-1.0em}
\end{figure}

Existing state-of-the-art NFs such as TARFlow~\cite{tarflow} deviate from standard flow-based modeling in that they learn a noise-perturbed distribution and then denoise the output. Specifically, during training, TARFlow takes a noise-perturbed input $\tilde{x} = x + \sigma\epsilon$, where $\epsilon\sim\mathcal{N}(\mathbf{0},\mathbf{I})$, and during inference, TARFlow first generates $\tilde{x}=\mathcal{F}^{-1}_\theta(z)$, then performs an additional \textit{score-based denoising} step:
\begin{equation}
    x\leftarrow\tilde{x}+\sigma^2\,\nabla_{\tilde{x}} \log p(\tilde{x}),
\end{equation}
as illustrated in \cref{fig:tarflow_denoising}, where the score term is computed via a forward-backward pass. This post-processing almost doubles the inference cost, becoming a clear computational bottleneck for efficient generation.

\paragraph{Learned Denoising.}
We eliminate the explicit score-based denoising step by integrating denoising directly into the reverse model. As illustrated in \cref{fig:biflow_denoising}, we extend the forward trajectory from $\tilde{x}$ to $z$ by appending the clean data $x$ at its start, and extend the reverse model with one additional block $h^0\rightarrow x'$ that learns denoising jointly with the inverse. The resulting reverse network, with one extra block for denoising, maps $z$ to a clean sample $x'$ in a single pass.
As such, our reverse model directly learns the \textit{correspondence} between $z$ and the \textit{clean data} $x$ directly, rather than the noisy data $\tilde{x}$.

The training process follows the same objective as \cref{eq:hidden_align_loss}, with a reconstruction loss on $(x, x')$ and hidden alignment losses on intermediate states. By integrating denoising into the reverse process itself, BiFlow achieves a unified learned formulation for generation, where inverse and denoising are seamlessly coupled within a single direct generative model, eliminating the need for any extra refinement step.

\subsection{Distance Metric}
\label{subsec:distance}

BiFlow provides a flexible supervised-learning framework for tackling the generation problem. This flexibility stems from two key properties of BiFlow: (i) \textit{1-NFE generation} — the learned reverse model produces a sample $x'$ in a single forward pass, so generated samples are directly accessible during training; and (ii) \textit{explicit pairing} — the forward process establishes a direct correspondence between data $x$ and noise $z$, serving as training pairs for the reverse model. Together, these properties realize a \emph{what-you-see-is-what-you-get} training regime: generated samples are available for immediate loss evaluation and backpropagation, enabling rich semantic supervision signals.

Our framework is highly flexible in the choice of loss functions: almost \textit{any distance metric} can be used, and multiple metrics can be combined. Our default choice for the distance metric $\mathcal D$ in \cref{eq:hidden_align_loss} is simply mean squared error (MSE). To enhance realism, we further apply perceptual loss at the final VAE-decoded image, while intermediate hidden states remain aligned by MSE. In this work, we adopt both VGG~\cite{vgg} and ConvNeXt V2~\cite{convnextv2} feature spaces for perceptual loss (our implementation for VGG features follows LPIPS~\cite{lpips}).
As in prior work~\cite{ect,icm,mf}, all loss terms can be adaptively re-weighted during training. Details are provided in Appendix~\ref{app:loss}.

\subsection{Norm Control}
\label{subsec:norm}

The intermediate states produced by the forward model are unconstrained under the NF formulation, often exhibiting large norm fluctuations across blocks (see \cref{fig:norm_wo_clip}). These variations can lead to imbalanced supervision when using magnitude-sensitive losses such as MSE for reverse-model training. To mitigate this issue, we introduce two complementary norm-control strategies applied to the forward and reverse models to ensure stable and consistent supervision strength (details in Appendix~\ref{app:loss}).

On the forward model, we clip the output parameters of each transformation $f_i$ within a fixed range $[-c, c]$, limiting excessive scaling and stabilizing intermediate state norms without compromising expressiveness.
On the reverse model, we normalize each intermediate state before performing hidden alignment, which equalizes the contribution across trajectory depth and promotes scale-invariant learning.

\subsection{BiFlow with Guidance}
\label{subsec:guidance}

Classifier-free guidance (CFG)~\cite{cfg} was originally proposed for diffusion models to control the trade-off between sample diversity and fidelity. 
Due to its effectiveness, it has been widely adopted in diffusion-based generative models.
Following this success, recent Normalizing Flows~\cite{tarflow,starflow} and autoregressive models~\cite{var,mar} also incorporate CFG to further improve generation quality.

CFG can be seamlessly integrated into BiFlow's inference process by extrapolating conditional and unconditional predictions of $\mcal G_\phi$ at each hidden state $h^i$, \ie,
\begin{align}
\label{eq:cfg}
h^{i+1}=(1+w_i)\,\mathcal G_\phi^i(h^i\mid\mathbf{c}) - w_i\,\mathcal G_\phi^i(h^i),
\end{align}
where $\mathbf{c}$ is the class condition and $w_i$ is the guidance scale (our $w$ definition follows the original CFG formulation~\cite{cfg}, \ie, $w=0$ is w/o CFG). 
The subscript $i$ indicates that $w_i$ can differ among blocks, supporting CFG interval~\cite{interval}. More results are provided in Appendix~\ref{app:guidance_exp}.

Directly applying CFG doubles the computational cost during inference, since each guided block requires two forward passes. To alleviate this, following~\cite{gft,mg}, we incorporate CFG into the training stage, enabling inference with only one function evaluation (1-NFE) while preserving the benefits of guidance. Additionally, to retain the flexibility of adjusting guidance scales at inference time, we allow the reverse model to leverage CFG scale as \emph{condition}~\cite{guidancedist,imf}. By training the model with a range of guidance scales, BiFlow can generate outputs corresponding to various guidance strengths within a single forward pass. Further details are provided in Appendix~\ref{app:guidance}.

\section{Experiments}
\label{sec:exp}

\paragraph{Experiment Settings.} Our experiments are conducted on class-conditional ImageNet~\cite{imagenet} generation at 256$\times$256 resolution. We evaluate Fr\'echet Inception Distance (FID)~\cite{fid} and Inception Score (IS)~\cite{improvedgan} on 50000 generated images. Following~\cite{ldm,sd3,starflow}, we implement our models on the latent space of a pre-trained VAE tokenizer. For ImageNet 256$\times$256, the tokenizer maps images to a 32$\times$32$\times$4 latent representation, serving as the input and output domain of our models.

\paragraph{Improved TARFlow as Baseline.}
Our BiFlow framework builds upon TARFlow~\cite{tarflow} as our forward model. 
We introduce several modifications to the original TARFlow to enhance stability and performance. Specifically, we replace additive conditioning with in-context conditioning~\cite{dit} and apply the norm control strategy in \cref{subsec:norm}, while omitting STARFlow-specific components such as deep-shallow design, decoder finetuning, and customized CFG.
We denote this enhanced version as \textit{improved TARFlow} (iTARFlow). As shown in the table below, it achieves substantial gains over the original TARFlow, both with or without CFG, establishing a strong \textit{baseline} for BiFlow.

\begin{table}[ht]
\centering
\vspace{-1em}
\tablestyle{2pt}{1.1}
\begin{tabular}{y{100}|x{40}x{40}x{30}}
\multirow{2}{*}{Method} & \multicolumn{2}{c}{FID ($\downarrow$)} & \multirow{2}{*}{\# Params} \\
& w/o CFG & w/ CFG & \\
\shline
latent TARFlow-B/2~\footnotemark & \baseline{59.43} & \baseline{10.89} & 118M \\
\hline
$+$ in-context conditioning & 53.87 & 8.25 & 120M\\
$+$ 160 epochs $\to$ 960 epochs & 45.48 & 7.05 & 120M \\
$+$ norm control (iTARFlow) & \textbf{44.46} & \textbf{6.83} & 120M\\
\end{tabular}
\vspace{-1.6em}
\end{table}

\footnotetext{The latent TARFlow-B/2 is our TARFlow reproduction in VAE latent.}

\paragraph{Configurations.} Our reverse model adopts a ViT backbone with modern Transformer components~\cite{rope,rmsnorm} and multi-token in-context conditioning~\cite{imf}. We name our model as BiFlow-B/2, where B/2 indicates a base-sized model with patch size 2, resulting in a sequence length of 256. In our ablation studies, we choose an iTARFlow as our forward model and train the reverse model with the forward model fixed. Unless otherwise specified, our ablations employ the adaptive-weighted MSE, while final comparisons in \cref{tab:imagenet256} incorporate perceptual distance mentioned in \cref{subsec:distance} for optimal performance. Details are provided in Appendix~\ref{app:impl}.

\begin{table}[t]
\centering
\vspace{-0.5em}
\tablestyle{4pt}{1.1}
\begin{tabular}{y{80}|x{55}x{55}}
 & FID ($\downarrow$) & attention \\
\shline
exact inverse & \baseline{44.46} & \baseline{causal} \\
\hline
naive distillation & \fidimproved{43.41}{1.05} & bidirect\\
hidden distillation & \fidworsened{55.00}{10.54} & bidirect\\
hidden alignment & \fidimproved{\textbf{36.93}}{7.53} & bidirect\\
\end{tabular}
\vspace{-0.8em}
\caption{\textbf{Reverse learning method}. Naive distillation can \textit{exceed} the exact inverse with a simple MSE objective. Our hidden alignment yields the best result among the three strategies.
(Settings: BiFlow-B/2, 160 epochs, adaptive weighted MSE loss, w/o CFG)}
\label{tab:reverse_learning}
\vspace{-1.8em}
\end{table}

\subsection{Ablation: Learning to Approximate the Inverse}

We evaluate three strategies for learning the reverse model, as described in \cref{subsec:reverse_learning}, and report generation quality (FID in \cref{tab:reverse_learning}) as well as reconstruction error (see Appendix~\ref{app:reverse_learning}).

The naive distillation approach, trained with a simple MSE objective, already \textit{outperforms} the exact inverse baseline, indicating that a learned reverse model is a \textit{practical and competitive} alternative to the analytic inverse.

Hidden distillation supervises the reverse model using the entire forward trajectory. However, repeated projections between representation and input spaces cause information loss and limit architectural expressiveness. This results in degraded performance compared to the naive distillation.

Our proposed hidden alignment method removes the repeated projections inherent in hidden distillation while retaining full trajectory-level supervision, thereby preserving both architectural flexibility and representational richness. It achieves the best performance among the three strategies and surpasses the exact inverse by a clear margin in generation quality. These results collectively demonstrate that hidden alignment is an effective and robust strategy for learning an approximate inverse in BiFlow.

\subsection{Other Ablations}

We ablate several key design choices in BiFlow and analyze their impact on performance in \cref{tab:ablations}.

\paragraph{BiFlow with Guidance.}
BiFlow is conditioned on the CFG scale and learns across a range of CFG scales during training. This enables 1-NFE inference while preserving the benefits and flexibility of guidance. As shown in \cref{subtab:cfg}, compared to standard CFG approach, our training-time CFG mechanism reduces inference cost by half while achieving better FID.

\paragraph{Learned Denoising.}  
\cref{subtab:denoise} demonstrates the effectiveness of our \emph{learned} denoising strategy.
By jointly training denoising with the inverse, our learned one-block denoiser improves generation quality over the score-based denoising used in TARFlow. Moreover, our approach introduces only a single additional block, whereas TARFlow's score-based denoising requires an extra forward-backward pass (incurring 15.8$\times$ flops). This substantially reduces inference overhead.

\begin{table}[t]
\centering
\vspace{-0.5em}
\begin{minipage}{\linewidth}
\begin{subtable}{\linewidth}
\centering
\tablestyle{6pt}{1.1}
\begin{tabular}{y{75}|x{48}x{48}}
 & FID, w/o CFG & FID, w/ CFG\\
\shline
inference-time CFG & 36.93 $_{(\text{NFE}=1)}$ & 6.90 $_{(\text{NFE}=2)}$\\
\hline
$\rightarrow$ training-time CFG & \textbf{31.88} $_{(\text{NFE}=1)}$ & \textbf{6.79} $_{(\text{NFE}=1)}$\\
\end{tabular}
\caption{\textbf{BiFlow with guidance}. Conditioning on the CFG scale during training improves FID both w/ and w/o CFG while preserving flexible 1-NFE inference. The baseline w/o CFG is final results in \cref{tab:reverse_learning}.}
\label{subtab:cfg}
\vspace{0.2em}
\end{subtable}
\begin{subtable}{\linewidth}
\centering
\tablestyle{6pt}{1.1}
\begin{tabular}{y{75}|x{48}x{48}}
 & FID, w/o CFG & FID, w/ CFG \\
\shline
learned denoise & \textbf{31.88} & \textbf{6.79} \\
\hline
$\rightarrow$ no denoise & 100.51 & 26.20 \\
$\rightarrow$ score-based denoise & 42.62 & 10.98 \\
\end{tabular}
\caption{\textbf{Learned denoising}. Our learned denoising scheme is effective. Compared to score-based denoising in TARFlow, it eliminates an extra forward-backward calculation, and unifies the denoising step into our framework.}
\label{subtab:denoise}
\vspace{0.2em}
\end{subtable}
\begin{subtable}{\linewidth}
\centering
\tablestyle{6pt}{1.1}
\begin{tabular}{y{75}|x{48}x{48}}
 & FID, w/o CFG & FID, w/ CFG\\
\shline
norm control: clip & \textbf{31.88} & \textbf{6.79} \\
\hline
norm control: none & 45.54 & 12.33 \\
norm control: traj. & 34.88 & 8.03 \\
\end{tabular}
\caption{\textbf{Norm control}. Either clipping the forward model's output or normalizing the forward trajectory improves generation quality by ensuring balanced supervision strength across blocks.}
\label{subtab:norm}
\vspace{0.2em}
\end{subtable}
\begin{subtable}{\linewidth}
\centering
\tablestyle{6pt}{1.1}
\begin{tabular}{y{75}|x{48}x{48}}
 & FID, w/o CFG & FID, w/ CFG \\
\shline
MSE & 31.88 & 6.79 \\
\hline
+ LPIPS & 14.15 & 4.91 \\
+ LPIPS + ConvNeXt & \textbf{2.46} & \textbf{2.46} \\
\end{tabular}
\caption{\textbf{Distance metric}. Our framework enables a flexible design of distance metrics. Incorporating perceptual distance improves generation quality.}
\label{subtab:distance}
\end{subtable}
\end{minipage}
\vspace{-0.5em}
\caption{\textbf{Ablation study on ImageNet 256$\times$256 generation.} FID-50K with 1-NFE is reported by default. (Settings: \textbf{BiFlow-B/2}, 160 epochs. By default: adaptive weighted MSE loss without perceptual loss, training-time CFG.)}
\label{tab:ablations}
\vspace{-0.5em}
\end{table}

\paragraph{Norm Control.} 
We introduce two norm control strategies in \cref{subsec:norm} and evaluate their effectiveness in \cref{subtab:norm}. Applying either strategy alleviates imbalance in MSE loss across blocks, thereby enhancing performance. We provide visualizations of the norm statistics in Appendix~\ref{app:improved_tarflow_norm_exp}.

\paragraph{Distance Metric.} 
Our framework supports various distance metric designs. 
As shown in \cref{subtab:distance}, incorporating perceptual distance~\cite{lpips,convnextv2} at the image end can largely improve generation quality. Notably, when both VGG and ConvNeXt features are used for the perceptual loss, the optimal guidance scale in \cref{eq:cfg} for this model is close to 0.0, resulting in performance similar to no-CFG setting. This suggests these features already provide strong class-discriminative information. More results are provided in Appendix~\ref{app:adpwgt}.

\paragraph{Scaling Behavior.} 
We investigate the scaling behavior of BiFlow under different distance metrics, using iTARFlow of corresponding size as forward models. 
We summarize preliminary results in the table below.

\begin{table}[H]
\vspace{-0.8em}
\centering
\tablestyle{2pt}{1.05}
\begin{tabular}{y{80}|x{40}x{40}}
\textbf{FID, w/ CFG} & B & XL\\
\shline
MSE & 6.79 & 4.61 \\
+ LPIPS & 4.91 & 3.36 \\
+ LPIPS + ConvNeXt & 2.46 & 2.57 \\
\end{tabular}
\vspace{-1.2em}
\end{table}

Overall, BiFlow exhibits clear gains from increased model capacity when trained \textit{without} the ConvNeXt-based perceptual loss. However, after incorporating ConvNeXt features, further scaling yields diminishing returns, with FID improvements gradually saturating. We hypothesize this behavior may be related to overfitting, as evidenced by an increase in FID during training. A comprehensive investigation of BiFlow's scaling behavior is left for future work.

\subsection{BiFlow vs. improved TARFlow}
\label{subsec:biflow_vs_itarflow}

\begin{table}[t]
\centering
\vspace{-0.8em}
\tablestyle{4pt}{1.15}
\setlength{\tabcolsep}{4pt}
\resizebox{\linewidth}{!}{
\begin{tabular}{x{30}|x{40}|x{40}x{40}x{40}x{40}}
\shline
 & \textbf{BiFlow} & \multicolumn{4}{c}{\textbf{improved TARFlow}} \\
& B/2 & B/2 & M/2 & L/2 & XL/2 \\
\hline
FID & \textbf{2.39} & 6.83 & 5.22 & 4.82 & 4.54 \\
\# Params & 133M & 120M & 296M & 448M & 690M \\
Gflops & 38 & 152 & 363 & 552 & 836 \\
\hline
\multicolumn{6}{l}{\textbf{Wall-clock time (ms)}} \\
TPU  & 0.29+1.3 & 65+1.3 & 85+1.3 & 165+1.3 & 202+1.3 \\
GPU & 2.15+2.7 & 129+2.7 & 208+2.7 & 349+2.7 & 400+2.7 \\
CPU & 80+240 & 9040+240 & 16200+240 & 20400+240 & 26300+240\\
\hline
\multicolumn{6}{l}{\textbf{Wall-clock speedup, BiFlow-B/2 \vs iTARFlow: (VAE excluded, see also \cref{fig:biflow_vs_itarflow})}} \\
TPU & - & 224$\times$ & \g{293$\times$} & \g{569$\times$} & \g{697$\times$} \\
GPU & - & 60$\times$ & \g{97$\times$} & \g{162$\times$} & \g{186$\times$} \\
CPU & - & 113$\times$ & \g{203$\times$} & \g{255$\times$} & \g{329$\times$} \\
\hline
\multicolumn{6}{l}{\textbf{Wall-clock speedup, BiFlow-B/2 \vs iTARFlow: (VAE included)}} \\
TPU & - & 42$\times$ & \g{54$\times$} & \g{105$\times$} & \g{128$\times$} \\
GPU & - & 27$\times$ & \g{43$\times$} & \g{73$\times$} & \g{83$\times$} \\
CPU & - & 29$\times$ & \g{51$\times$} & \g{65$\times$} & \g{83$\times$} \\
\shline
\end{tabular}
}
\vspace{-0.7em}
\caption{
\textbf{Comparison between BiFlow and iTARFlow baseline}.
We report both generation quality (FID-50K) and inference cost per image.
All wall-clock time measurements are reported as ``\textit{generator} + \textit{VAE decoding}''.
Compared to iTARFlow, BiFlow achieves one to two orders of magnitude faster sampling on TPU, GPU, and CPU, while attaining superior generation quality. (The VAE decoder contains 49M parameters and requires 308 Gflops.)
}
\label{tab:biflow_vs_itarflow}
\vspace{-1.7em}
\end{table}

\begin{table*}[t]
\centering
\begin{minipage}[b]{0.36\textwidth}
  \renewcommand{\arraystretch}{1.05}
  \small
  \setlength{\tabcolsep}{4pt}
  \centering
  \resizebox{!}{0.086\textheight}{
  \begin{tabular}{@{}lcccc@{}}
    \toprule
    {Method} & {\# Params} & NFE & {FID($\downarrow$)} & IS($\uparrow$) \\
    \midrule
    \multicolumn{5}{@{}l}{\textit{\textbf{Autoregressive Normalizing Flow}}} \\
    ~~TARFlow-XL/8@pix~\cite{tarflow} & 1.3B & $\star$ & 5.56 & -\\
    ~~STARFlow-XL/1~\cite{starflow} & 1.4B & $\star$ & 2.40 & 
    - \\
    \multicolumn{5}{@{}l}{\textit{\textbf{Autoregressive Normalizing Flow (our impl.)}}} \\
    ~~iTARFlow-B/2 & 120M & $\star$ & 6.83 & 226.2 \\
    ~~iTARFlow-M/2 & 296M & $\star$ & 5.22 & 255.5 \\
    ~~iTARFlow-L/2 & 448M & $\star$ & 4.82 & 254.8 \\
    ~~iTARFlow-XL/2 & 690M & $\star$ & 4.54 & 259.3 \\
    \midrule
    \multicolumn{4}{@{}l}{\textit{\textbf{1-NFE Normalizing Flow}}} \\
    ~~BiFlow-B/2 (Ours) & 133M & 1 & \textbf{2.39} & \textbf{303.0} \\
    \bottomrule
  \end{tabular}
  }
\end{minipage}
\hspace{2pt}
\textcolor[rgb]{0.5,0.5,0.5}{
\begin{minipage}[b]{0.30\textwidth}
  \renewcommand{\arraystretch}{1.05}
  \small
  \setlength{\tabcolsep}{4pt}
  \centering
  \resizebox{!}{0.11\textheight}{
  \begin{tabular}{@{}lcccc@{}}
    \toprule
    {Method} & {\# Params} & NFE & {FID($\downarrow$)} & IS($\uparrow$) \\
    \midrule
    \multicolumn{4}{@{}l}{\textit{\textbf{GANs}}} \\
    ~~BigGAN-deep~\cite{biggan}    & 112M & 1 & 6.95 & 202.6 \\
    ~~GigaGAN~\cite{gigagan}       & 569M & 1 & 3.45 & 225.5 \\
    ~~StyleGAN-XL~\cite{stylegan} & 166M & 1 & {2.30} & 265.1 \\
    \midrule
    \multicolumn{4}{@{}l}{\textit{\textbf{1-NFE diffusion/flow matching from scratch}}} \\
    ~~iCT-XL/2~\cite{icm} & 675M & 1 & 34.24 & - \\
    ~~Shortcut-XL/2~\cite{shortcut} & 675M & 1 & 10.60 & - \\
    ~~MeanFlow-XL/2~\cite{mf} & 676M & 1 & 3.43 & 247.5 \\
    ~~TiM-XL/2~\cite{tim} & 664M & 1 & 3.26 & 210.3 \\
    ~~$\alpha$-Flow-XL/2+~\cite{alphaflow} & 676M & 1 & 2.58 & - \\
    ~~iMF-XL/2~\cite{imf} & 610M & 1 & 1.72 & 282.0 \\
    \midrule
    \multicolumn{4}{@{}l}{\textit{\textbf{1-NFE diffusion/flow matching (distillation)}}} \\
    ~~$\pi$-Flow-XL/2~\cite{piflow} & 675M & 1 & 2.85 & - \\
    ~~DMF-XL/2+~\cite{dmf} & 675M &  1 & 2.16 & - \\
    ~~FACM-XL/2~\cite{facm} & 675M & 1 & 1.76 &  290.0 \\
    \bottomrule
  \end{tabular}
  }
\end{minipage}
}
\hspace{1pt}
\textcolor[rgb]{0.5,0.5,0.5}{
\begin{minipage}[b]{0.30\textwidth}
  \renewcommand{\arraystretch}{1.05}
  \small
  \setlength{\tabcolsep}{4pt}
  \centering
  \resizebox{!}{0.11\textheight}{
  \begin{tabular}{@{}lcccc@{}}
    \toprule
    {Method} & {\# Params} & NFE & {FID($\downarrow$)} & IS($\uparrow$) \\
    \midrule
    \multicolumn{4}{@{}l}{\textit{\textbf{autoregressive/masking}}} \\
    ~~MaskGIT~\cite{maskgit}     & 227M & $\star$ & 6.18 & 182.1 \\
    ~~RCG, conditional~\cite{rcg} & 512M & $\star$ & 2.12 & 267.7 \\
    ~~VAR-$d30$~\cite{var}  & 2.0B & $\star$ & 1.92 & 323.1 \\
    ~~MAR-H~\cite{mar}     & 943M & $\star$ & 1.55 & 303.7 \\
    ~~RAR-XXL~\cite{rar} & 1.5B & $\star$ & 1.48 & 326.0 \\
    ~~xAR-H~\cite{xar}    & 1.1B & $\star$ & 1.24 & 301.6 \\
    \midrule
    \multicolumn{4}{@{}l}{\textit{\textbf{Multi-NFE diffusion/flow matching}}} \\
    ~~ADM-G~\cite{adm} & 554M & 250$\times$2 & 4.59 & - \\
    ~~LDM-4-G~\cite{ldm}      & 400M & 250$\times$2 & 3.60 & 247.7 \\
    ~~DiT-XL/2~\cite{dit} & 675M & 250$\times$2 & 2.27 & 278.2 \\
    ~~SiT-XL/2~\cite{sit} & 675M & 250$\times$2 & 2.06 & 252.2 \\
    ~~JiT-G/16~\cite{jit} & 2B & 100$\times$2 & 1.82 & 292.6 \\
    ~~SiT-XL/2\,+\,REPA~\cite{repa}   & 675M & 250$\times$2 & 1.42 & 305.7 \\
    ~~LightningDiT-XL/1~\cite{vavae} & 675M & 250$\times$2 & 1.35 & 295.3 \\
    ~~DDT-XL/2~\cite{ddt} & 675M & 250$\times$2 & 1.26  & 310.6 \\
    ~~DiT$^\text{DH}$-XL\,+\,RAE~\cite{rae} & 839M & 50$\times$2 & 1.13 & 262.6 \\
    \bottomrule
  \end{tabular}
  }
\end{minipage}
}
\caption{\textbf{System-level comparison on ImageNet 256\boldmath${\times}$256 class-conditional generation}. All results are reported with CFG if applicable. \textbf{Left:} Comparison with Normalizing Flow models. \textbf{Middle:} Other 1-NFE generative models, including GANs and diffusion/flow matching-based models. \textbf{Right:} Other families of generative models. Our BiFlow model is trained for 350 epochs with perceptual distance. In all tables, $\times$2 indicates the use of CFG incurs double NFEs. 
$\star$: All AR-based methods, including AR Normalizing Flow (left) and other AR models (right), involve a large number of forward evaluations, yet each evaluation is on one or a very few tokens. For example, for standard left-to-right order AR, the \textit{average} NFE of the entire AR process is roughly 1 (or 2$\times$ w/ CFG), that is, ${K}$ evaluations with a $\frac{1}{K}$ fraction of tokens each. In addition, TARFlow/iTARFlow has an extra NFE of 2 due to the score-based denoising post-processing. {\scriptsize Results of \cite{icm} is collected from \cite{imm}; results of \cite{tarflow} is collected from \cite{starflow}; TARFlow-XL/8@pix denotes TARFlow on pixel-space with patch size 8.}
}
\vspace{-0.5em}
\label{tab:imagenet256}
\end{table*}

We compare our learned reverse model (BiFlow) with the exact analytic inverse baseline (improved TARFlow) of the forward process. In \cref{tab:biflow_vs_itarflow}, we benchmark in terms of generation quality (FID score) and inference efficiency (flops and wall-clock time for generating a single image). Details of our benchmarking setup are provided in Appendix~\ref{app:impl}.

Experiments show that our BiFlow-B/2 surpasses the exact inverse of the improved TARFlow-XL/2 baseline in generation quality. Remarkably, BiFlow requires only a single function evaluation (1-NFE), compared to 256$\times$2 sequential decoding steps for the autoregressive inference of the exact analytic inverse — resulting in up to a 42$\times$ speedup for models of similar size on TPU.

\paragraph{\textit{Why can a learned inverse outperform the exact inverse?}}
Our reverse model $\mcal G_\phi$ is trained to reconstruct real images directly, rather than to replicate synthetic samples produced by the exact inverse as in conventional distillation. This encourages its predictions to align more closely with the true data distribution. In addition, $\mcal G_\phi$ is optimized \textit{end-to-end} with the forward map fixed, learning to directly transform noise into clean data. This \textit{joint} optimization can help the model to learn a stable and globally consistent mapping.

\paragraph{\textit{Why is a learned inverse significantly faster than the \mbox{exact} inverse?}}
From an algorithmic perspective, two key improvements reduce the computational cost of BiFlow. First, BiFlow eliminates the score-based denoising step required by the exact inverse of TARFlow, removing a major computational bottleneck. Second, we integrate CFG into the training stage, effectively halving the inference cost compared to applying CFG during sampling.
Together, these two improvements reduce the flops by roughly 4$\times$.

From an architectural perspective, the autoregressive design of TARFlow imposes inherent limitations on parallelism during inference. Our bidirectional attention Transformer design allows for \textit{fully parallelized} computation across the sequence dimension, which leads to significant speedups on modern accelerators. Notably, due to the efficiency of BiFlow, the VAE decoder has become a dominant computational overhead, which is outside the scope of this work.

\subsection{Comparison with Prior Works}

In \cref{tab:imagenet256}, we provide system-level comparisons with previous methods on class-conditional ImageNet 256$\times$256 generation. We categorize prior works into three groups: Normalizing Flows (\cref{tab:imagenet256}, left), 1-NFE generative models (\cref{tab:imagenet256}, middle), and other families of generative models (\cref{tab:imagenet256}, right). All our models are trained to convergence.

\paragraph{Comparison with Normalizing Flows.} \cref{tab:imagenet256} (left) compares BiFlow with previous state-of-the-art Normalizing Flows models. Our BiFlow-B/2, with only 133 million parameters, achieves an FID of 2.39 in a single function evaluation (1-NFE), establishing a new state-of-the-art among Normalizing Flows.
In contrast, STARFlow uses thousands of sequential decoding steps due to their autoregressive sampling process. 
It yields a similar FID score with about 10$\times$ parameters and more than 400$\times$ inference wall-clock time (see \cref{tab:inference_speed} for details).

More broadly, BiFlow represents a significant advancement in Normalizing Flows, demonstrating that direct and efficient generation can coexist with high fidelity. 

\paragraph{Comparison with Other Generative Models.} We compare BiFlow with other generative model families, especially 1-NFE methods. As shown in \cref{tab:imagenet256}, BiFlow offers an excellent balance between generation quality and sampling efficiency. These results demonstrate that BiFlow achieves performance on par with leading 1-NFE generative models.

\begin{figure}[t]
    \centering
    \includegraphics[width=0.98\linewidth]{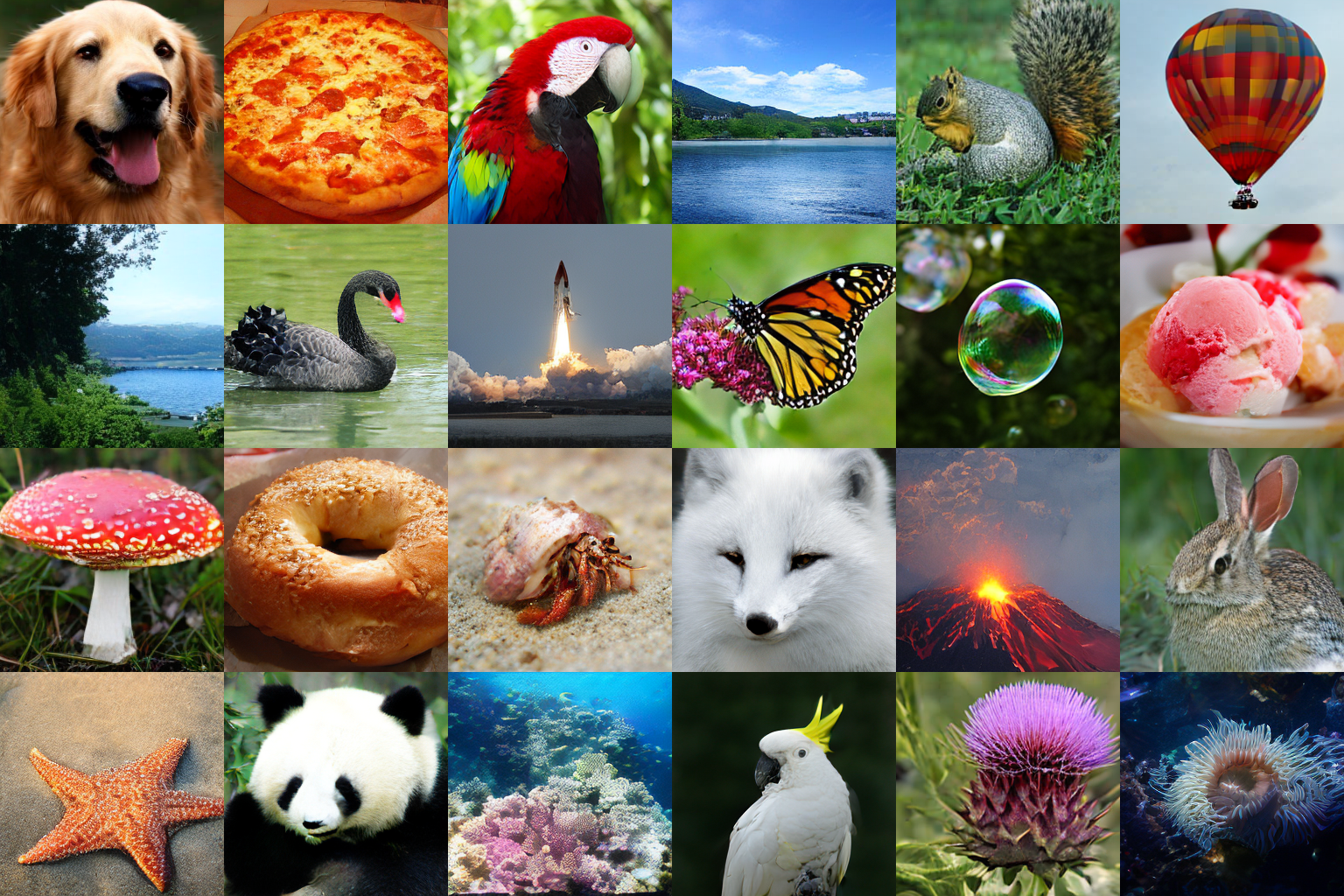}
    \vspace{-0.2em}
    \caption{\textbf{1-NFE Generation Results}. We show selected samples generated by our BiFlow-B/2 model with guidance scale 2.0 on ImageNet 256$\times$256. BiFlow achieves high-fidelity generation with only a single function evaluation (1-NFE) from noise.}
    \vspace{-1.5em}
\end{figure}

\section{Conclusion}
\label{sec:conclusion}
This work revisits one of the oldest, yet most principled, foundations of generative modeling — Normalizing Flows — and redefines its boundaries. We challenge the conventional wisdom that the reverse process must be the exact analytic inverse of the forward process, and demonstrate that the long-held constraint is unnecessary. 
By introducing a learnable reverse model, BiFlow pushes Normalizing Flows from analytically invertible mappings to trainable bidirectional systems, from autoregressive sampling to fully parallelized, efficient 1-NFE generation, and from an implicit generative model towards a \textit{direct generative model}. Experiments demonstrate that BiFlow achieves competitive generation quality among Normalizing Flows, while delivering up to two orders of magnitude faster inference than its explicit inverse counterpart.
We hope this work can serve as a step toward rethinking and expanding the scope of Normalizing Flows, inspiring future research on direct, flexible, and efficient NF-based generation.

\clearpage
\appendix

\section{Implementation Details}
\label{app:impl}

\begin{table}[t]
\tablestyle{4pt}{1.1}
\resizebox{\linewidth}{!}{
\begin{tabular}{y{60}|x{40}|x{25}x{25}x{25}x{25}}
\multirow{2}{*}{config} & BiFlow & \multicolumn{4}{c}{improved TARFlow}\\
& B/2 & B/2 & M/2 & L/2 & XL/2 \\
\shline
\# Params (M) & 133 & 120 & 296 & 448 & 690 \\
block  & 9 & 8 & 10 & 12 & 15\\
layer & 8 & 8 & 9 & 9 & 9 \\
hidden dim & 384 & 384 & 512 & 576 & 640 \\
attn heads & 6 & 6 & 8 & 9 & 10 \\
patch size & 2 & \multicolumn{4}{c}{2} \\
class tokens & 8 & \multicolumn{4}{c}{---}\\
guidance tokens & 4$^\dagger$ / 1 & \multicolumn{4}{c}{---} \\
\hline
epochs & 160$^\dagger$ / 350 & 960 & 640 & 640 & 480 \\
batch size & 256 & \multicolumn{4}{c}{256} \\
learning rate & 4e-4 & 4e-4 & 4e-4 & 2e-4 & 1e-4 \\
lr schedule & constant & \multicolumn{4}{c}{constant} \\
lr warmup & 10 epochs & \multicolumn{4}{c}{10 epochs} \\
optimizer & Adam & \multicolumn{4}{c}{Adam \cite{adam}} \\
Adam $(\beta_1, \beta_2)$ & (0.9, 0.95) & \multicolumn{4}{c}{(0.9, 0.95)} \\
weight decay & 0.0 & \multicolumn{4}{c}{0.0} \\ 
dropout & 0.0 & \multicolumn{4}{c}{0.0} \\
ema decay & 0.9999 & \multicolumn{4}{c}{0.9999} \\
label drop & 0.1 & \multicolumn{4}{c}{0.1} \\
\hline
adaptive weight $p$  & 1 & \multicolumn{4}{c}{---} \\
$w_{\text{VGG}}$ & 1.0$^\dagger$ / 0.8 & \multicolumn{4}{c}{---} \\
$w_{\text{ConvNeXt}}$ & 0.4$^\dagger$ / 0.6 & \multicolumn{4}{c}{---} \\
clip range $c$ & --- & 1.0 & 1.0 & 3.0 & 3.0 \\
noise level $\sigma$ & --- & \multicolumn{4}{c}{0.3} \\
\end{tabular}
}
\caption{\textbf{Configurations and training hyperparameters on ImageNet 256$\times$256}. $^\dagger$ indicates the setting in the ablation study.
}
\label{tab:config}
\end{table}

We implement all experiments using the JAX framework~\cite{jax} on Google TPU hardware. All reported results are obtained on TPU v4, v5p, and v6e cores. 
The configurations and training hyperparameters for improved TARFlow and BiFlow are provided in \cref{tab:config}.
For the MSE-only ablation in \cref{sec:exp}, we employ adaptive weighting with exponent $p=2$; for all other experiments we use $p=1$ (see Appendix~\ref{app:adpwgt} for detailed ablations). 

\paragraph{FID Evaluation.}
For generative evaluation, we compute the Fréchet Inception Distance (FID)~\cite{fid} between 50,000 generated images and training images, without applying any data augmentation. We use the Inception-V3 model~\cite{inceptionv3} provided by StyleGAN3~\cite{stylegan3}, converted into a JAX-compatible implementation. We sample 50 images per class for all 1000 ImageNet classes, following the protocol in \cite{rae}.

\paragraph{Inference Cost Evaluation.}
In \cref{fig:biflow_vs_itarflow} and \cref{tab:biflow_vs_itarflow}, we report inference cost across three hardware configurations: GPU, TPU, and CPU. For all metrics, we report the average per-image runtime in seconds, averaged over multiple runs to ensure stability. All measurements include the overhead of CFG and VAE decoding time when applicable.
We also provide a comparison with prior Normalizing Flow models~\cite{tarflow, starflow} in \cref{tab:inference_speed}. 
All autoregressive models utilize KV-cache to accelerate inference, and Gflops in \cref{tab:biflow_vs_itarflow} is estimated using JAX's \texttt{cost\_analysis} function.

For TPU wall-clock time, all models are evaluated using a pre-compiled JAX sampling function on 8 TPU v4 cores. Reported times exclude compilation overhead. We use a local device batch size of 10 for model inference, and 200 for VAE decoding.

For GPU wall-clock time, all models are re-implemented in PyTorch and evaluated on a single NVIDIA H200 GPU with a batch size of 128. The VAE decoding time is obtained with \texttt{torch.compile} optimization. 

For CPU wall-clock time, we reuse the PyTorch implementation on a single AMD EPYC 7B12 node (120 physical CPU cores and 400 GB RAM). We use a smaller batch size of 64 for most models; however, TARFlow and STARFlow are restricted to a batch size of 4 due to efficiency concerns. We observe that batch size has a negligible impact on per-image CPU inference time. All other experimental settings remain consistent with the GPU evaluation.

\begin{table}[t]
\centering
\begin{minipage}{\linewidth}
\begin{subtable}{\linewidth}
\centering
\tablestyle{4pt}{1.2}
\setlength{\tabcolsep}{4pt}
\resizebox{\linewidth}{!}{
\begin{tabular}{y{85}|x{40}x{40}z{30}x{30}}
Method & \# Params & Time (ms) & Speed & VAE?\\
\shline
TARFlow-XL/8@pix~\cite{tarflow} & 1.3B & 1192 & $1\times$ & \ding{55} \\
STARFlow-XL/1~\cite{starflow} & 1.4B  & 677+1.3 & 1.76$\times$ & \ding{51} \\
\hline
iTARFlow-B/2 & 120M & 65+1.3 & 18.0$\times$ & \ding{51} \\
iTARFlow-M/2 & 296M & 85+1.3 & 13.8$\times$ & \ding{51} \\
iTARFlow-L/2 & 446M & 165+1.3 & 7.17$ \times$ & \ding{51} \\
iTARFlow-XL/2 & 675M & 202+1.3 & 5.86$\times$ & \ding{51} \\
\hline
BiFlow-B/2 (Ours) & 133M & \textbf{0.29+1.3} & $\boldsymbol{750\times}$ & \ding{51} \\
\end{tabular}
}
\caption{TPU inference time comparison, benchmarked on 8 TPU v4 cores with a pre-compiled JAX sampling function.}
\label{subtab:tpuspeed}
\vspace{0.5em}
\end{subtable}
\begin{subtable}{\linewidth}
\centering
\tablestyle{4pt}{1.2}
\setlength{\tabcolsep}{4pt}
\resizebox{\linewidth}{!}{
\begin{tabular}{y{85}|x{40}x{40}z{30}x{30}}
Method & \# Params & Time (ms) & Speed & VAE?\\
\shline
TARFlow-XL/8@pix~\cite{tarflow} & 1.3B & 3452 & $1\times$ & \ding{55} \\
STARFlow-XL/1~\cite{starflow} & 1.4B  & 2193+2.7 & $1.57\times$ & \ding{51} \\
\hline
iTARFlow-B/2 & 120M & 129+2.7 & $26.2\times$ & \ding{51} \\
iTARFlow-M/2 & 296M & 208+2.7 & $16.4\times$ & \ding{51} \\
iTARFlow-L/2 & 446M & 349+2.7 & $9.82\times$ & \ding{51} \\
iTARFlow-XL/2 & 675M & 400+2.7 & $8.57\times$ & \ding{51} \\
\hline
BiFlow-B/2 (Ours) & 133M & \textbf{2.15+2.7} & $\boldsymbol{712\times}$ & \ding{51} \\
\end{tabular}
}
\caption{GPU inference time comparison, benchmarked on 1 NVIDIA H200 core with PyTorch and \texttt{torch.compile} optimization if beneficial.}
\label{subtab:gpuspeed}
\vspace{0.5em}
\end{subtable}
\begin{subtable}{\linewidth}
\centering
\tablestyle{4pt}{1.2}
\setlength{\tabcolsep}{4pt}
\resizebox{\linewidth}{!}{
\begin{tabular}{y{85}|x{40}x{40}z{30}x{30}}
Method & \# Params & Time (ms) & Speed & VAE?\\
\shline
TARFlow-XL/8@pix~\cite{tarflow} & 1.3B & 512000 & $1\times$ & \ding{55} \\
STARFlow-XL/1~\cite{starflow} & 1.4B  & 276700+240 & $1.85\times$ & \ding{51} \\
\hline
iTARFlow-B/2 & 120M & 9040+240 & $55.2\times$ & \ding{51} \\
iTARFlow-M/2 & 296M & 16200+240 & $31.1\times$ & \ding{51} \\
iTARFlow-L/2 & 446M & 20400+240 & $24.8\times$ & \ding{51} \\
iTARFlow-XL/2 & 675M & 26300+240 & $19.3\times$ & \ding{51} \\
\hline
BiFlow-B/2 (Ours) & 133M & \textbf{80+240} & $\boldsymbol{1600\times}$ & \ding{51} \\
\end{tabular}
}
\caption{CPU inference time comparison, benchmarked on 1 AMD EPYC 7B12 node with 120 physical CPU cores and 400GB RAM. We reuse the PyTorch implementations with \texttt{torch.compile} optimization if beneficial.}
\label{subtab:cpuspeed}
\end{subtable}
\end{minipage}
\caption{\textbf{Comparison of NF models' inference wall-clock time on TPU, GPU, and CPU}. The wall-clock time is evaluated per image on average in milliseconds. All models include the overhead of CFG at inference time, as well as the VAE decoding time when applicable. All autoregressive models utilize KV-cache to accelerate inference. See Appendix~\ref{app:impl} for further details.}
\label{tab:inference_speed}
\end{table}

\section{Method Details}

\subsection{Pseudocode}
\label{app:code}

We provide the pseudocode for training our BiFlow model with hidden alignment in Alg.~\ref{alg:train}, as well as the 1-NFE sampling procedure in Alg.~\ref{alg:sample}.

In the algorithm, the forward model $\mcal F$ produces the entire forward trajectory \texttt{xs}, \ie, $\tilde{x}, x^1, x^2, \ldots, x^{B-1}$, along with the prior $z=x^B$.
Similarly, the reverse model $\mcal G$ outputs the sequence of intermediate hiddens states \texttt{hs} as reverse trajectory: $h^{B-1}, h^{B-2}, \ldots, h^0$, along with the reconstructed clean input \texttt{x\_prime}. 

The final loss function consists of alignment loss between forward and reverse hidden states in \cref{eq:hidden_align_loss}, and reconstruction loss between the clean input \texttt{x} and reconstructed output \texttt{x\_prime}.

\definecolor{codeblue}{rgb}{0.25,0.5,0.5}
\definecolor{codekw}{rgb}{0.85, 0.18, 0.50}
\definecolor{codesign}{RGB}{0, 0, 255}
\definecolor{codefunc}{rgb}{0.85, 0.18, 0.50}
\definecolor{keywordblue}{rgb}{0.18, 0.50, 0.85}

\lstdefinelanguage{PythonFuncColor}{
  language=Python,
  keywordstyle=\color{blue},
  commentstyle=\color{codeblue},
  stringstyle=\color{orange},
  showstringspaces=false,
  basicstyle=\ttfamily\small,
  literate=
    {*}{{\color{codesign}* }}{1}
    {-}{{\color{codesign}-}}{1}
    {+}{{\color{codesign}+ }}{1}
    {patchify}{{\color{codefunc}patchify}}{1}
    {concat}{{\color{codefunc}concat}}{1}
    {flip}{{\color{codefunc}flip}}{1}
    {randn}{{\color{codefunc}randn}}{1}
    {randn_like}{{\color{codefunc}randn\_like}}{1}
    {adaptive_mean}{{\color{codefunc}adaptive\_mean}}{1}
    {stopgrad}{{\color{codefunc}stopgrad}}{1}
    {metric}{{\color{codefunc}metric}}{1}
    {mse}{{\color{codefunc}mse}}{1}
    {perceptual}{{\color{codefunc}perceptual}}{1}
    {append}{{\color{codefunc}append}}{1}
}

\lstset{
  language=PythonFuncColor,
  backgroundcolor=\color{white},
  basicstyle=\fontsize{9pt}{9.9pt}\ttfamily\selectfont,
  columns=fullflexible,
  breaklines=true,
  captionpos=b,
}

\begin{algorithm}[t]
\caption{{BiFlow}: Training.}
\label{alg:train}
\begin{lstlisting}
# x: training batch, (N, H, W, C)
# F: forward model (B blocks), frozen
# G: reverse model (B +1 blocks)
# phi: projection heads

# noise injection
e = randn_like(x)
x_tilde = x + noise_level * e

# get forward trajectory xs and prior z
xs, z = F(x_tilde)

# get reverse trajectory hs and reconstructed x'
hs, x_prime = G(z)

# project hidden into input space
for i in range(B):
    hs[i] = phi[i](hs[i])

# compute loss
loss_align = mse(xs, hs)
loss_recon = metric(x, x_prime)
loss = loss_align + loss_recon
        
\end{lstlisting}
\end{algorithm}

\begin{algorithm}[t]
\caption{{BiFlow}: 1-NFE Sampling.
}
\label{alg:sample}
\begin{lstlisting}
e = randn(x_shape)
_, x = G(e)
\end{lstlisting}
\end{algorithm}

\subsection{BiFlow with Guidance}
\label{app:guidance}

\paragraph{Training-time CFG.} As discussed in \cref{subsec:guidance}, to enable guided sampling within a single forward pass (1-NFE), we directly train a guided reverse model $\mcal G_{\phi}^{\text{cfg}}$ defined as
\begin{align*}
    \mcal G_{\phi}^{i,\text{cfg}}(h^i\mid\mathbf{c}) = (1+w_i) \mcal G_{\phi}^i(h^i\mid\mathbf{c}) - w_i\mcal G^i_{\phi}(h^i).
\end{align*}
where $w_i$ is the guidance scale at block $i$.
The unconditional output of $\mcal G_{\phi}^{i,\text{cfg}}$ matches that of the original $\mcal G_{\phi}^i$. Therefore, the unguided block output can be expressed as
\begin{align}
\label{eq:mg}
h^{i+1}=\frac{\mcal G_{\phi}^{i,\text{cfg}}(h^i\mid\mathbf{c})+w_i\mcal G^{i,\text{cfg}}_{\phi}(h^i)}{1+w_i}.
\end{align}
During training, we compute our hidden-alignment loss directly on $h^{i+1}$ from \cref{eq:mg}.
At inference time, this formulation allows us to use $\mcal G_{\phi}^{i,\text{cfg}}$ directly, producing guided samples with only a 1-NFE forward pass. We add stop gradient to the unconditional output to stabilize training.

\paragraph{Guidance conditioning.}
To retain the ability to adjust the CFG scale at inference time, we explicitly condition the reverse model on the guidance scale~\cite{guidancedist,imf}, \ie, $\mcal G_{\phi}^{i,\text{cfg}}(h^i \mid \mathbf{c}, w_i)$.
During training, we sample $w_i$ from a uniform distribution $\mcal U(0, w_{\max})$ and apply \cref{eq:mg} to compute the unguided output $h^{i+1}$ for hidden alignment loss.

\begin{figure}[t]
    \centering
    \vspace{-1em}
    \includegraphics[width=0.9\linewidth,clip]{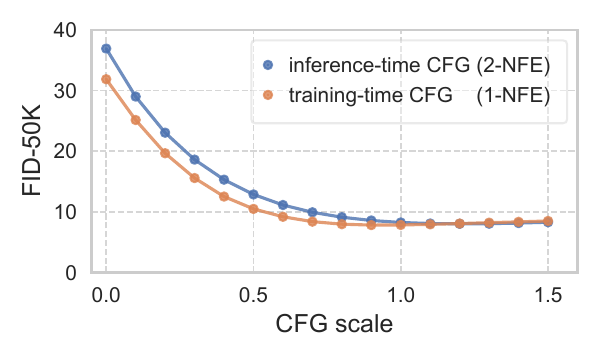}
    \vspace{-1.5em}
    \caption{
        \textbf{Comparison between training-time and inference-time CFG of BiFlow}. Training-time CFG achieves similar or better performance compared to inference-time CFG while requiring only half inference compute and retaining full post-hoc tuning flexibility. (Settings: BiFlow-B/2, 160 epochs, MSE-only baseline.)
    }
    \vspace{-0.5em}
    \label{fig:cfg}
\end{figure}

We compare this training-time CFG scheme with the more conventional inference-time CFG in \cref{fig:cfg}. Training-time CFG achieves similar (even better) performance while preserving the 1-NFE efficiency and the flexibility to sweep CFG scales at inference time.

\subsection{Distance Metric}
\label{app:loss}

\paragraph{Adaptive weighting.}
We adopt the adaptive loss reweighting strategy from~\cite{ect,icm,mf}. Given a prediction $x$ and target $y$, the adaptive-weighted distance is defined as:
\begin{align*}
\label{eq:adpwgt}
\mcal D_p = \mathrm{sg}(w_p)\cdot \mcal D(x,y), \quad w_p=(\mcal D(x, y)+c)^{-p}, 
\end{align*}
where $c$ is a small constant and $p \geq 0$ is adaptive weight. $\text{sg}(\cdot)$ denotes the stop-gradient operator. We apply adaptive weighting to all loss terms in our training objective.

\paragraph{VGG feature.}
For the perceptual loss based on VGG features, we follow the LPIPS formulation~\cite{lpips}. Since our model operates in the latent space of a pre-trained VAE, we decode the predicted latent $x'$ back into image space and compute the LPIPS loss against the ground-truth image.

\paragraph{ConvNeXt feature.}
In addition to VGG features, we incorporate ConvNeXt V2~\cite{convnextv2} (ImageNet-22K pre-trained, base-size) as a complementary perceptual feature extractor. Similar to the LPIPS, both the reconstructed image $x'$ and the ground-truth image $x$ are passed through the ConvNeXt network after VAE decoding. The perceptual distance is computed using the extracted features, excluding the final classification head.

\paragraph{Usage of loss terms.}
In the ablation studies in \cref{sec:exp}, we use only the adaptively weighted MSE loss unless otherwise noted. In \cref{tab:imagenet256}, we combine all three loss terms:
\begin{align*}
\mcal L(x)&=\sum_{i}\mcal L_{\text{MSE}}(x^i, \varphi_i(h^i))\\
&+ w_{\text{VGG}}\mcal L_{\text{VGG}}(x, x') + w_{\text{ConvNeXt}}\mcal L_{\text{ConvNeXt}}(x, x') ,
\end{align*}
where $w_{\text{VGG}}$ and $w_{\text{ConvNeXt}}$ are tunable hyperparameters. We observe that the final performance is particularly sensitive to $w_{\text{ConvNeXt}}$. Concrete weights are specified in Appendix~\ref{app:impl}.

\paragraph{Normalized Trajectory.}
As described in \cref{subsec:norm}, the reverse model is trained to align with a normalized forward trajectory. Specifically, we pre-compute the squared norm $\|x_i\|^2$ of each trajectory point and average it over the entire dataset. During reverse model training, the intermediate trajectory points are divided by $\sqrt{\mathbb{E}[\|x_i\|^2]}$, ensuring scale consistency across different blocks.

We do not use normalized trajectories in any experiments except the one in \cref{subtab:norm}, as we observe no significant difference when combined with iTARFlow. Nonetheless, normalized trajectories are worth noting for scenarios where one wishes to use a pre-trained NF model without clipping.

\section{Additional Experiments}
\label{app:exp}

\begin{table}[t]
\centering
\tablestyle{4pt}{1.1}
\begin{minipage}{1.0\linewidth}
\centering
\begin{tabular}{y{75}|x{40}x{40}}
 & MSE & LPIPS \\
\shline
naive distillation & 0.115 & 0.331 \\
hidden distillation & 0.156 & 0.392 \\
hidden alignment & \textbf{0.111} & \textbf{0.321} \\
\end{tabular}
\caption{\textbf{Reverse learning methods: reconstruction fidelity.}
We report MSE and LPIPS between the original sample $x$ and the reconstructed sample $x'$ produced by the learned reverse model. Among the three strategies for approximating the inverse transformation, the hidden alignment method achieves the most accurate reconstruction. The corresponding FIDs are shown in \cref{tab:reverse_learning} (Settings: BiFlow-B/2, 160 epochs, no CFG, adaptive-weighted MSE loss only. All three rows share the same forward model.)}
\vspace{-0.5em}
\label{tab:reverse_loss}
\end{minipage}
\end{table}

\subsection{Learning to Approximate the Inverse}
\label{app:reverse_learning}

In \cref{tab:reverse_learning}, we compare the empirical performance of the three reverse learning approaches introduced in \cref{sec:method}. Here, we further provide quantitative results on their reconstruction fidelity in \cref{tab:reverse_loss}. Specifically, we evaluate the reconstruction distance $\mcal D(x, x')$ using MSE and LPIPS (VGG-based) as metrics.

We observe that the proposed hidden-alignment strategy achieves the lowest regression loss across both metrics. This indicates that hidden alignment provides a more accurate mapping between $x$ and $x'$, leading to a better-behaved reverse learning process.

\subsection{BiFlow with Guidance}
\label{app:guidance_exp}

\begin{table}[t]
\tablestyle{4pt}{1.1}
\begin{tabular}{x{30}|x{30}x{30}x{30}}
$w_d\backslash w$ & $0.5$ & $0.6$ & $0.7$ \\
\shline
0.5 & 10.51 & 9.45 & 8.77 \\
0.6 & 10.21 & 9.21 & 8.59 \\
0.7 & 9.93 & 8.99 & 8.41 \\
\hline
3.5 & 7.05 & 6.87 & 6.89 \\
4.0 & 6.94 & 6.81 & 6.87 \\
4.5 & 6.88 & \textbf{6.79} & 6.87 \\
5.0 & 6.86 & 6.80 & 6.89 \\
\end{tabular}
\caption{\textbf{Separate guidance scale for the denoising block}. BiFlow eliminates the score-based denoising step in TARFlow by learning a dedicated denoising block, jointly trained with other blocks. This denoising block serves a different purpose from the rest NF reverse process. Using a separate, larger guidance scale for the denoising block improves sample quality. (Settings: BiFlow-B/2, 160 epochs, adaptively-weighted MSE only, training-time CFG. FID w/o CFG: 31.88.)}
\vspace{-1em}
\label{tab:separate_denoise_cfg}
\end{table}

As discussed in \cref{subsec:denoise}, the additional denoising block in our reverse model functions as a dedicated denoiser, while the preceding $B$ blocks focus on inverting the forward sub-transformations. This structure naturally motivates applying CFG differently across these two components. We empirically validate this design choice in \cref{tab:separate_denoise_cfg}.

For training-time CFG, we use a shared guidance scale across all blocks, sampling $w$ from a simple uniform prior $\mathcal{U}[0, 0.5]$. In ablation studies that use MSE loss only (\cref{sec:exp}), we decouple the guidance scales for the inverse blocks and the denoising block, since the optimal pair $(w, w_d)$ typically satisfies $w_d \gg w$. In this case, we sample $w\sim\mathcal{U}[0,1]$ and $w_d\sim\mathcal{U}[0,8]$.

\subsection{Distance Metric}
\label{app:adpwgt}

In \cref{subsec:distance}, we discuss different choices of distance metrics for training BiFlow. We ablate the choice of perceptual distance terms in \cref{tab:perceptual_ablation}. First, we compare different feature extractors, including a ResNet-101 \cite{resnet} pre-trained for classification and a DINOv2-B model~\cite{dinov2}, as reported in \cref{tab:feature_model}. For ResNet-101, we extract features by removing the final MLP head, following the same procedure as ConvNeXt. Among all tested feature extractors, ConvNeXt achieves the best empirical performance. We further evaluate the combination of VGG and ConvNeXt features in \cref{tab:w_convnext}. The results indicate that using both features together yields better FID scores than using either one individually.

Furthermore, we study the effect of adaptive weighting in the MSE loss in \cref{tab:adp_wt}. MSE with adaptive weighting consistently outperforms the naive MSE loss.

\begin{table}[t]
\centering
\begin{subtable}{\linewidth}
\centering
\tablestyle{6pt}{1.1}
\begin{tabular}{y{75}|x{48}x{48}}
feature model & FID, w/o CFG & FID, w/ CFG \\
\shline
none & 31.88 & 6.79 \\
VGG + ResNet & 9.69 & 4.34 \\
VGG + DINO & 9.33 & 4.36 \\
ConvNeXt + DINO & 3.19 & 3.19 \\
VGG + ConvNeXt & \textbf{2.46} & \textbf{2.46} \\
\end{tabular}
\caption{Feature model ablation.}
\vspace{0.4em}
\label{tab:feature_model}
\end{subtable}
\begin{subtable}{\linewidth}
\centering
\tablestyle{6pt}{1.1}
\begin{tabular}{x{30}x{40}|x{48}x{48}}
$w_{\text{VGG}}$ & $w_{\text{ConvNeXt}}$ & FID, w/o CFG & FID, w/ CFG \\
\shline
0.0 & 0.0 & 31.88 & 6.79 \\
1.0 & 0.0 & 16.97 & 5.31 \\
0.0 & 0.4 & 2.62 & 2.62 \\
1.0 & 0.4 & \textbf{2.46} & \textbf{2.46} \\
\end{tabular}
\caption{VGG and ConvNeXt weight ablation.}
\label{tab:w_convnext}
\end{subtable}
\vspace{-1.5em}
\caption{\textbf{Ablation on perceptual loss for BiFlow-B/2}. FID-50K with/without CFG are reported. (Settings: BiFlow-B/2, 160 epochs, training-time CFG, weight for two perceptual losses are 1.0 and 0.4 by default.)}
\label{tab:perceptual_ablation}
\end{table}

\begin{table}[t]
\centering
\tablestyle{6pt}{1.05}
\begin{tabular}{x{72}|x{48}x{48}}
adaptive weight $p$ & FID, w/o CFG & FID, w/ CFG \\
\shline
0.0 & 38.23 & 7.49 \\
\hline
0.5 & 34.74 & 7.08 \\
1.0 & 34.43 & \textbf{6.70} \\
2.0 & \textbf{31.88} & 6.79 \\
\end{tabular}
\caption{\textbf{Ablation on adaptive weighting}. Adaptive weighted MSE loss works better than naive MSE ($p$ = 0.0). (Settings: BiFlow-B/2, 160 epochs, adaptive weighted MSE only, training-time CFG.)}
\label{tab:adp_wt}
\end{table}

\begin{figure}[t]
    \centering
    \begin{subfigure}[t]{0.98\linewidth}
        \centering
        \includegraphics[width=\linewidth,clip,page=1]{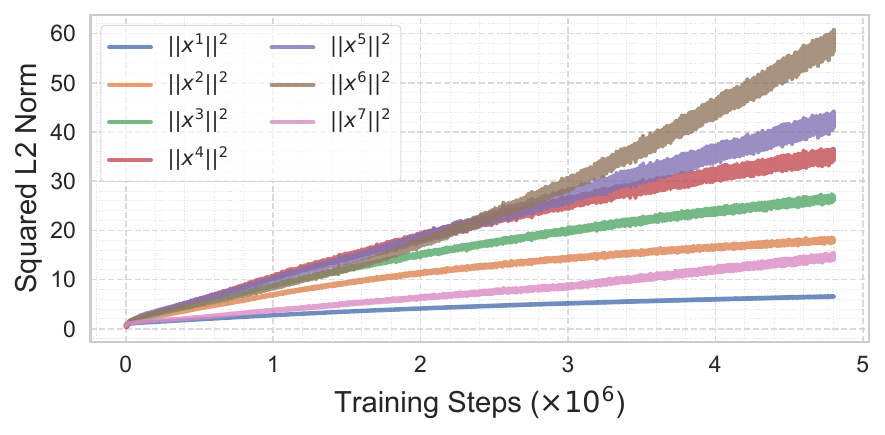}
        \caption{\centering Improved TARFlow w/o clipping.}
        \label{fig:norm_wo_clip}
    \end{subfigure}
    \hfill
    \begin{subfigure}[t]{0.98\linewidth}
        \centering
        \includegraphics[width=\linewidth,clip,page=2]{imgs/norm.pdf}
        \caption{\centering Improved TARFlow w/ clipping.}
        \label{fig:norm_w_clip}
    \end{subfigure}
    \vspace{-0.5em}
    \caption{Comparison of intermediate states' norm during TARFlow training between training with and without clipping. \textbf{Left}: without clipping, the norms at different blocks diverge significantly, and continue to increase as training proceeds. \textbf{Right}: with clipping, the norms are well controlled within a reasonable range, stabilizing training and improving final FID scores.}
    \label{fig:norm}
\end{figure}

\subsection{Improved TARFlow Norm Control}
\label{app:improved_tarflow_norm_exp}

To mitigate the imbalance in loss magnitudes across different blocks of BiFlow, we introduce a simple yet effective modification to the original TARFlow~\cite{tarflow} in \cref{subsec:norm}: clipping the parameters $\mu$ and $\alpha$ within a fixed range. This adjustment stabilizes training and improves final FID performance.

In \cref{fig:norm}, we visualize the norms of intermediate trajectory states during training of the improved TARFlow. Without clipping, the norms across blocks diverge sharply and continue to grow as training progresses. With clipping, the norms remain stable and well-controlled within a reasonable range. Such normalization substantially benefits the training of the reverse model in the downstream.

\subsection{Improved TARFlow CFG}\label{app:tarflow_cfg}

For completeness, we also examine classifier-free guidance (CFG) designs for TARFlow, although this component is orthogonal to our main contributions. In the original TARFlow~\cite{tarflow}, the reverse update rule at block $i$ is
\begin{align*}
    z_{t,\text{cfg}}^i=z_t^{i+1}\odot\exp(\alpha_{t, \text{cfg}}^i)+\mu_{t, \text{cfg}}^i,
\end{align*}
where guidance is applied to the predicted parameters by
\begin{equation*}
    \begin{aligned}
    \alpha_{t, \text{cfg}}^i &= \left(1+w_t\right)\alpha_t^i(\cdot\mid\mathbf{c})
    -w_t\,\alpha_t^i(\cdot),\\
    \mu_{t, \text{cfg}}^i &= \left(1+w_t\right)\mu_t^i(\cdot\mid\mathbf{c})
    -w_t\,\mu_t^i(\cdot),
    \end{aligned}
\end{equation*}
with a linearly increasing guidance schedule $w_t = \frac{t}{T-1} w$ along the token dimension. Here, subscript $t$ denotes the token dimension, superscript $i$ denotes the block dimension, and $(\cdot\mid\mathbf{c})$ and $(\cdot)$ represent the conditional and unconditional counterparts, respectively.

\begin{table}[t]
\tablestyle{4pt}{1.2}
\resizebox{\linewidth}{!}{
\begin{tabular}{y{30}|x{40}x{40}x{40}x{40}}
 & \multicolumn{2}{c}{Linear} & \multicolumn{2}{c}{Const}\\
 & $\mu,\alpha$ & $z$ & $\mu,\alpha$ & $z$ \\
\shline
Online & \ \ \textbf{6.83}$_{\g{(2.8)}}$ & \ \ \textbf{6.82}$_{\g{(2.8)}}$ & \ \ 7.26$_{\g{(1.3)}}$ & \ \ 7.24$_{\g{(1.3)}}$ \\
\hline
Offline & 22.14$_{\g{(1.2)}}$ & 22.03$_{\g{(1.2)}}$ & 18.23$_{\g{(1.0)}}$ & 18.11$_{\g{(1.0)}}$ \\
\end{tabular}
}
\caption{Improved TARFlow CFG ablation. Online guidance substantially outperforms the offline variants, whereas the choice of guidance schedule (linear vs. constant) and the level at which guidance is applied ($\mu,\alpha$ vs. $z$) has only minor impact. Numbers in gray parentheses denote the corresponding optimal CFG scale. (Settings: improved TARFlow-B/2, FID w/o CFG: 44.46.)}
\label{tab:tarflow_cfg}
\vspace{-0.5em}
\end{table}

Following prior CFG studies in diffusion models, we decompose the design space into three orthogonal choices:

\emph{Schedule.}
We can replace the original linearly increasing $w_t$ with a constant guidance scale: $w_t=w$.

\emph{Space for applying guidance}.
Parameter-space CFG $(\mu, \alpha)$ vs. pixel-space CFG applied directly to $z$:
\begin{align*}
    z_{t, \text{cfg}}^i=(1+w_t)z^i_t(\cdot\mid\mathbf{c})-w_tz_t^i(\cdot).
\end{align*}
We denote these two settings by ``$\mu,\alpha$'' and ``$z$'', respectively.

\emph{Online vs. Offline.}
We distinguish between online and offline CFG strategies. The online approach (TARFlow's practice) applies guidance at each generation step; the offline approach generates the entire conditional and unconditional sequences independently and performs extrapolation only once on the final outputs. The difference lies only in how guidance interacts with intermediate states.

While both approaches have similar runtimes, \cref{tab:tarflow_cfg} shows that online CFG significantly outperforms the offline variant. Regarding other hyperparameters, a linear schedule offers a slight advantage over a constant one, while applying guidance in parameter space versus pixel space yields similar performance. Overall, TARFlow's original CFG formulation is close to optimal.

Based on these results, we use the original TARFlow CFG formulation (Online, Linear, Parameter-space) as our baseline. It is important to note that this CFG setting only affects the inference quality of the forward model; the training of our BiFlow reverse model always relies on the unguided forward trajectory.

\section{Training-free Image Editing}

BiFlow naturally supports several training-free image editing applications by explicitly modeling a bidirectional mapping between the data and noise domains. We showcase two representative applications: inpainting and class editing. For brevity, we omit the VAE encoder/decoder in the following descriptions, as they always serve as pre-/post-processing steps in our experiments.

\begin{figure}[t]
\centering
\tablestyle{0pt}{3.0}
\begin{tabular}{
    @{} ccc @{\hspace{16pt}} ccc @{}
}
\includegraphics[width=0.15\linewidth]{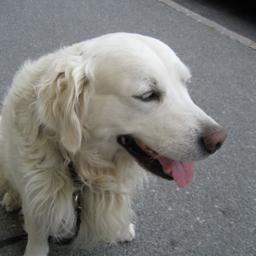} & 
\includegraphics[width=0.15\linewidth]{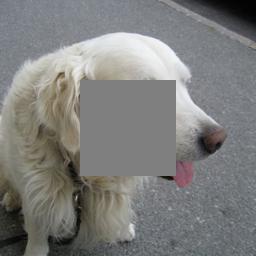} & 
\includegraphics[width=0.15\linewidth]{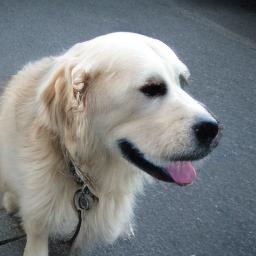} &
\includegraphics[width=0.15\linewidth]{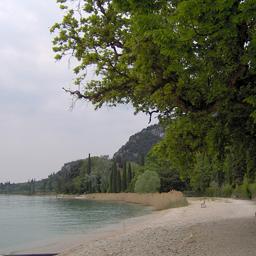} & 
\includegraphics[width=0.15\linewidth]{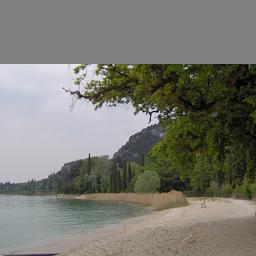} & 
\includegraphics[width=0.15\linewidth]{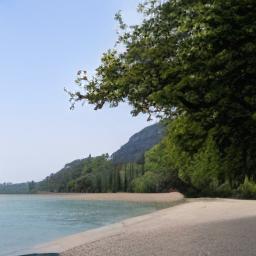}\\
\includegraphics[width=0.15\linewidth]{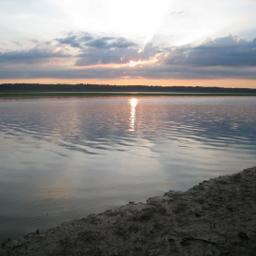} & 
\includegraphics[width=0.15\linewidth]{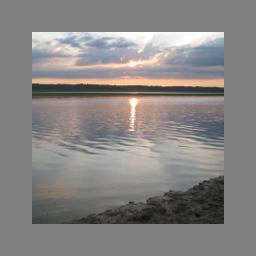} & 
\includegraphics[width=0.15\linewidth]{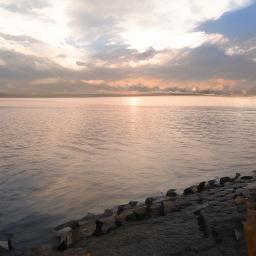} & 
\includegraphics[width=0.15\linewidth]{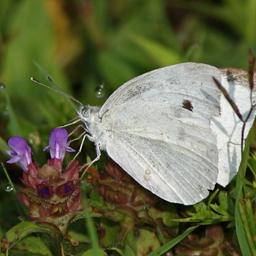} & 
\includegraphics[width=0.15\linewidth]{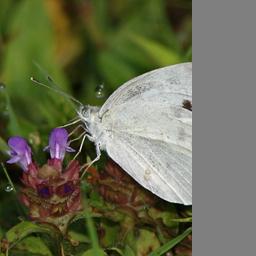} & 
\includegraphics[width=0.15\linewidth]{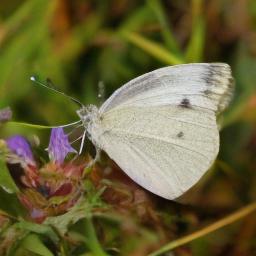} \\
\includegraphics[width=0.15\linewidth]{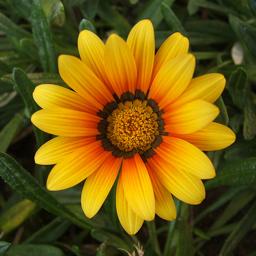} & 
\includegraphics[width=0.15\linewidth]{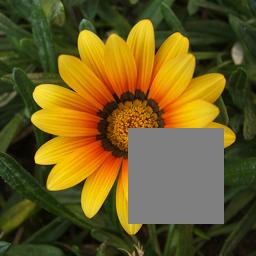} & 
\includegraphics[width=0.15\linewidth]{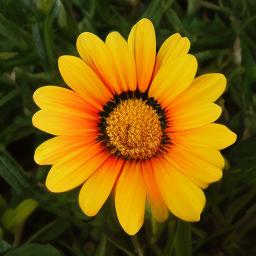} & 
\includegraphics[width=0.15\linewidth]{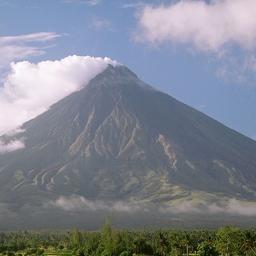} & 
\includegraphics[width=0.15\linewidth]{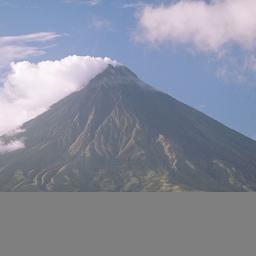} & 
\includegraphics[width=0.15\linewidth]{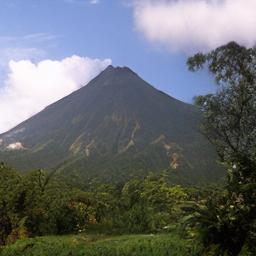} \\
\end{tabular}
\caption{\textbf{Inpainting}. BiFlow enables efficient image inpainting by leveraging its bidirectional mapping between images and noise. By resampling the masked part of the noise, BiFlow can perform training-free inpainting on various image masks. Each triplet contains ground-truth image (left), masked image (middle), and reconstructed image (right).}
\label{fig:inpaint}
\end{figure}

\begin{figure}[t]
\centering
\tablestyle{4pt}{1.1}
\begin{tabular}{
    c@{\hspace{6pt}}c@{\hspace{3pt}}c @{\hspace{4pt}} | @{\hspace{8pt}} c@{\hspace{6pt}}c@{\hspace{3pt}}c
}
egyptian cat & \multirow{2}{*}[-5ex]{$\longrightarrow$} & kit fox &
hen & \multirow{2}{*}[-5ex]{$\longrightarrow$} & flamingo\\
\includegraphics[width=0.15\linewidth]{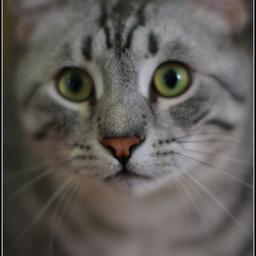} & & \includegraphics[width=0.15\linewidth]{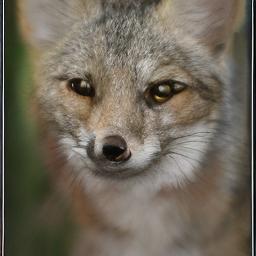} &
\includegraphics[width=0.15\linewidth]{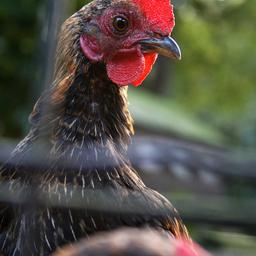} & & \includegraphics[width=0.15\linewidth]{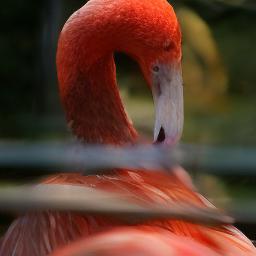}\\
[2mm]
tennis ball & \multirow{2}{*}[-5ex]{$\longrightarrow$} & golden retriever &
daisy & \multirow{2}{*}[-5ex]{$\longrightarrow$} & custard apple\\
\includegraphics[width=0.15\linewidth]{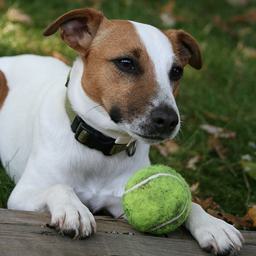} & & \includegraphics[width=0.15\linewidth]{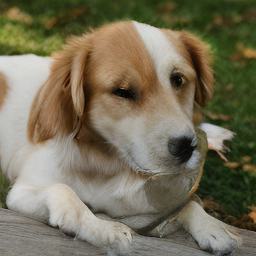} &
\includegraphics[width=0.15\linewidth]{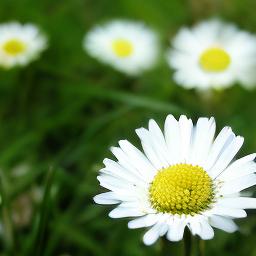} & & \includegraphics[width=0.15\linewidth]{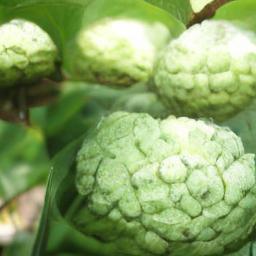}\\
\end{tabular}
\caption{\textbf{Class Editing}. BiFlow constructs an explicit bidirectional mapping between images and noise. With this property, BiFlow is able to conduct training-free class editing by modifying only the label condition in the forward and reverse process.}
\vspace{-0.3em}
\label{fig:classedit}
\end{figure}

\paragraph{Inpainting.}
The forward model $\mathcal{F}_\theta$ encodes an image $x$ into noise $z=\mathcal{F}_\theta(x)$. We empirically observe that localized perturbations in $z$ predominantly affect corresponding spatial regions in the reconstructed image.

Based on this property, BiFlow enables inpainting with an arbitrary binary mask $\mathcal{M}\in\{0,1\}^{H\times W}$.
Given a masked image $x_{\text{mask}}=\mathcal{M}\odot x$, we first map it to the noise domain using the forward model: $z_{\text{mask}}=\mathcal{F}_\theta(x_{\text{mask}})$. We then \textit{resample} the masked portion of the prior as
\begin{align*}
z'=\mathcal M\odot z_{\text{mask}}+(1-\mathcal M)\odot \epsilon,\quad \epsilon\sim \mathcal N(\bm0,\mI).
\end{align*}
Finally, the modified noise $z'$ is mapped back to the image domain by the reverse model $\mcal G_\phi$. This procedure fills the masked region with content coherent with the context. Representative examples are shown in \cref{fig:inpaint}.

\paragraph{Class Editing.}
The reverse model $\mathcal{G}_\phi$ allows us to generate images from noise $z$ under different class conditions.
For a fixed $z$, changing the class label $\mathbf{c}$ primarily modifies the class-dependent appearance while largely preserving the global spatial structure.

Concretely, given an image $x$ with label $\mathbf{c}$, we obtain its prior variable $z=\mathcal{F}_\theta(x\mid \mathbf{c})$, and reconstruct it using a different label $\mathbf{c}'$, writing $x'=\mathcal{G}_\phi(z\mid\mathbf{c}')$. As illustrated in \cref{fig:classedit}, BiFlow effectively alters class-specific attributes while maintaining the overall structure, enabling intuitive class editing without retraining.

\paragraph{Efficiency.}
Both inpainting and class editing require only a single forward pass from data to noise and a single reverse pass from noise to data, making BiFlow a lightweight and efficient tool for training-free image manipulation.

\section{Visualizations}

\begin{figure*}[t]
    \centering

    \begin{minipage}[t]{0.46\linewidth}
        \centering
        \includegraphics[width=\textwidth]{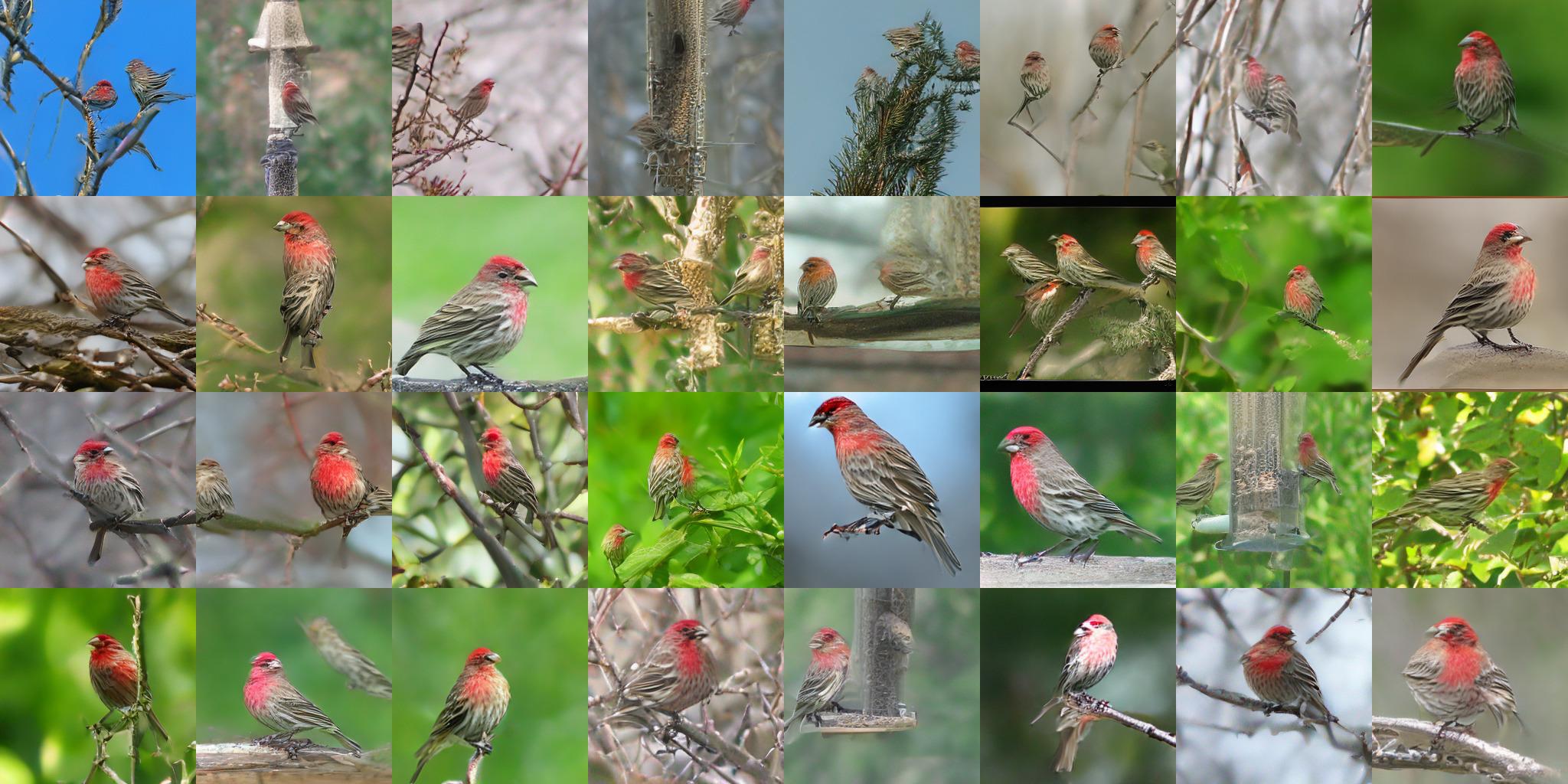}
        {\scriptsize class 12: house finch, linnet, Carpodacus mexicanus}
        \vspace{1em}
    \end{minipage}
    \hfill
    \begin{minipage}[t]{0.46\linewidth}
        \centering
        \includegraphics[width=\textwidth]{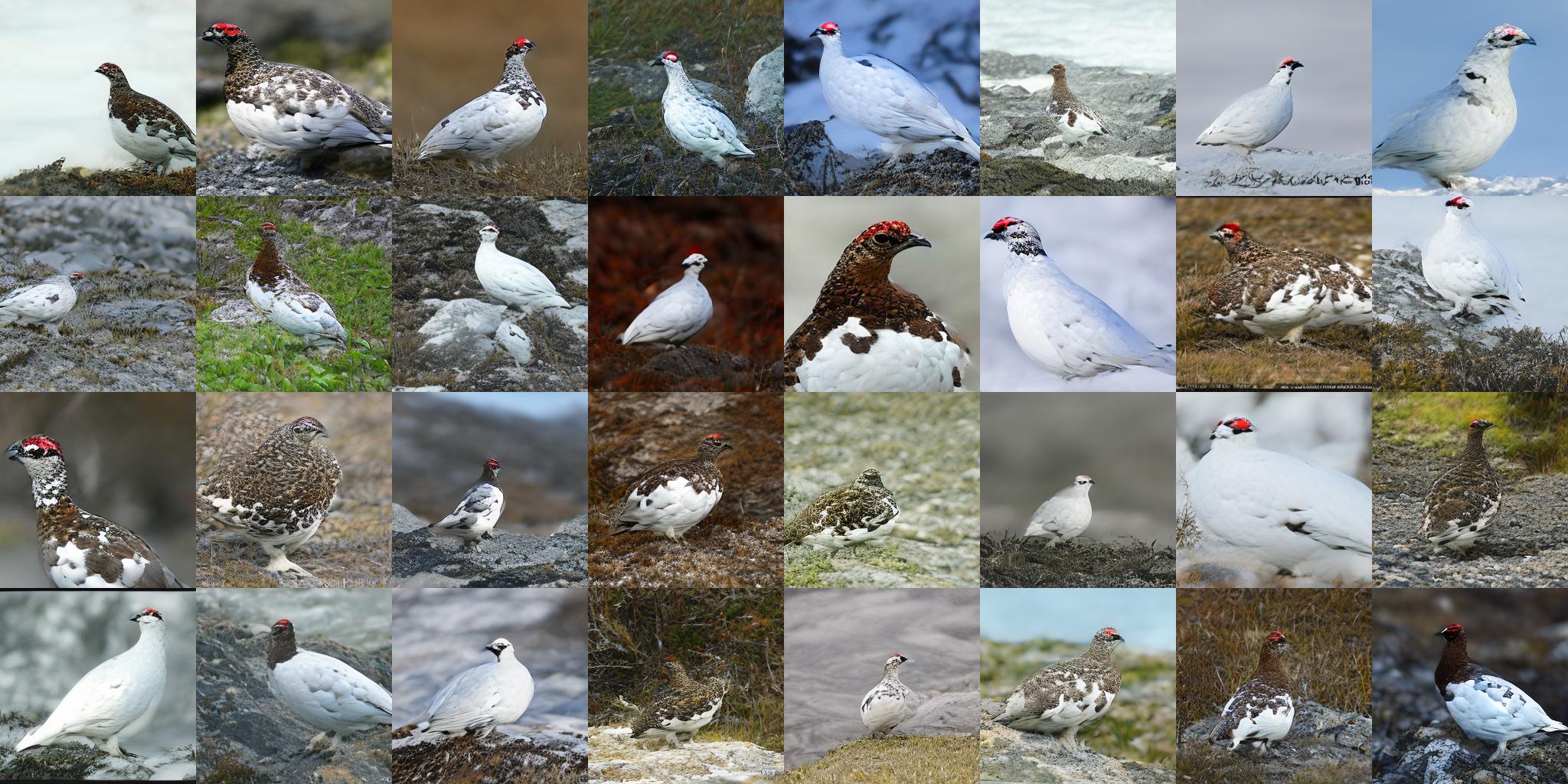}
        {\scriptsize class 81: ptarmigan}
        \vspace{1em}
    \end{minipage}

    \begin{minipage}[t]{0.46\linewidth}
        \centering
        \includegraphics[width=\textwidth]{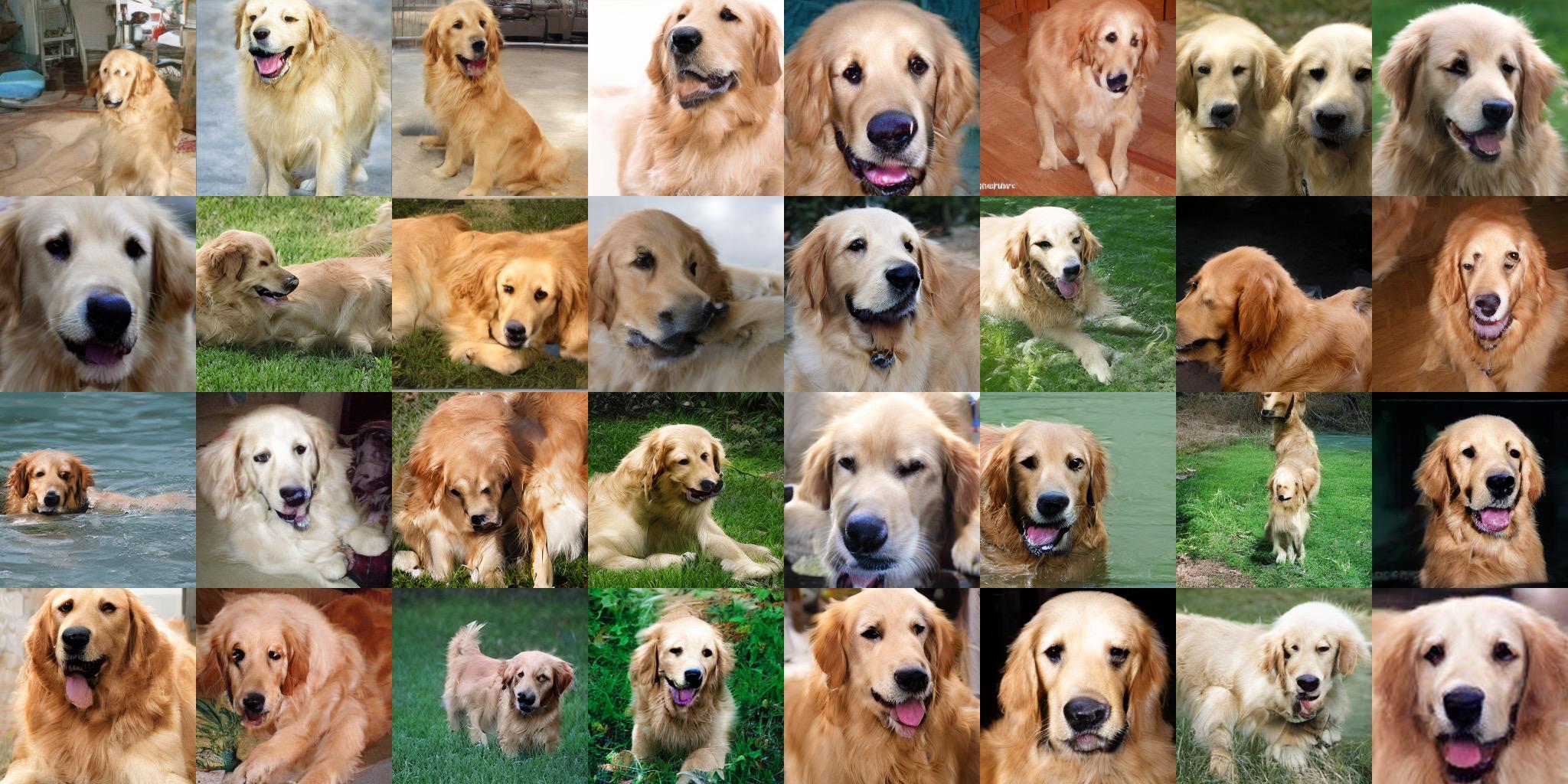}
        {\scriptsize class 207: golden retriever}
        \vspace{1em}
    \end{minipage}
    \hfill
    \begin{minipage}[t]{0.46\linewidth}
        \centering
        \includegraphics[width=\textwidth]{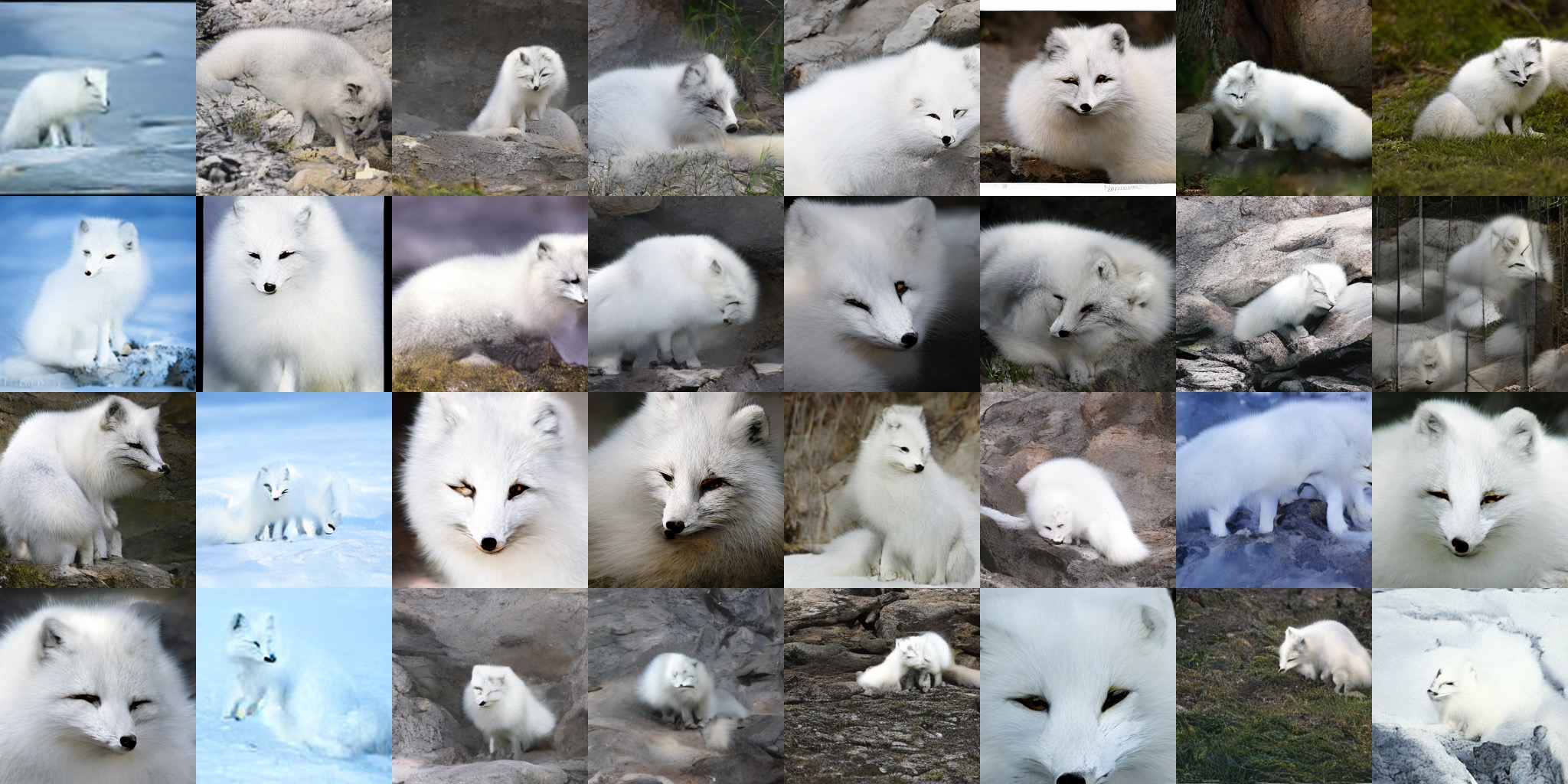}
        {\scriptsize class 279: Arctic fox, white fox, Alopex lagopus}
        \vspace{1em}
    \end{minipage}

    \begin{minipage}[t]{0.46\linewidth}
        \centering
        \includegraphics[width=\textwidth]{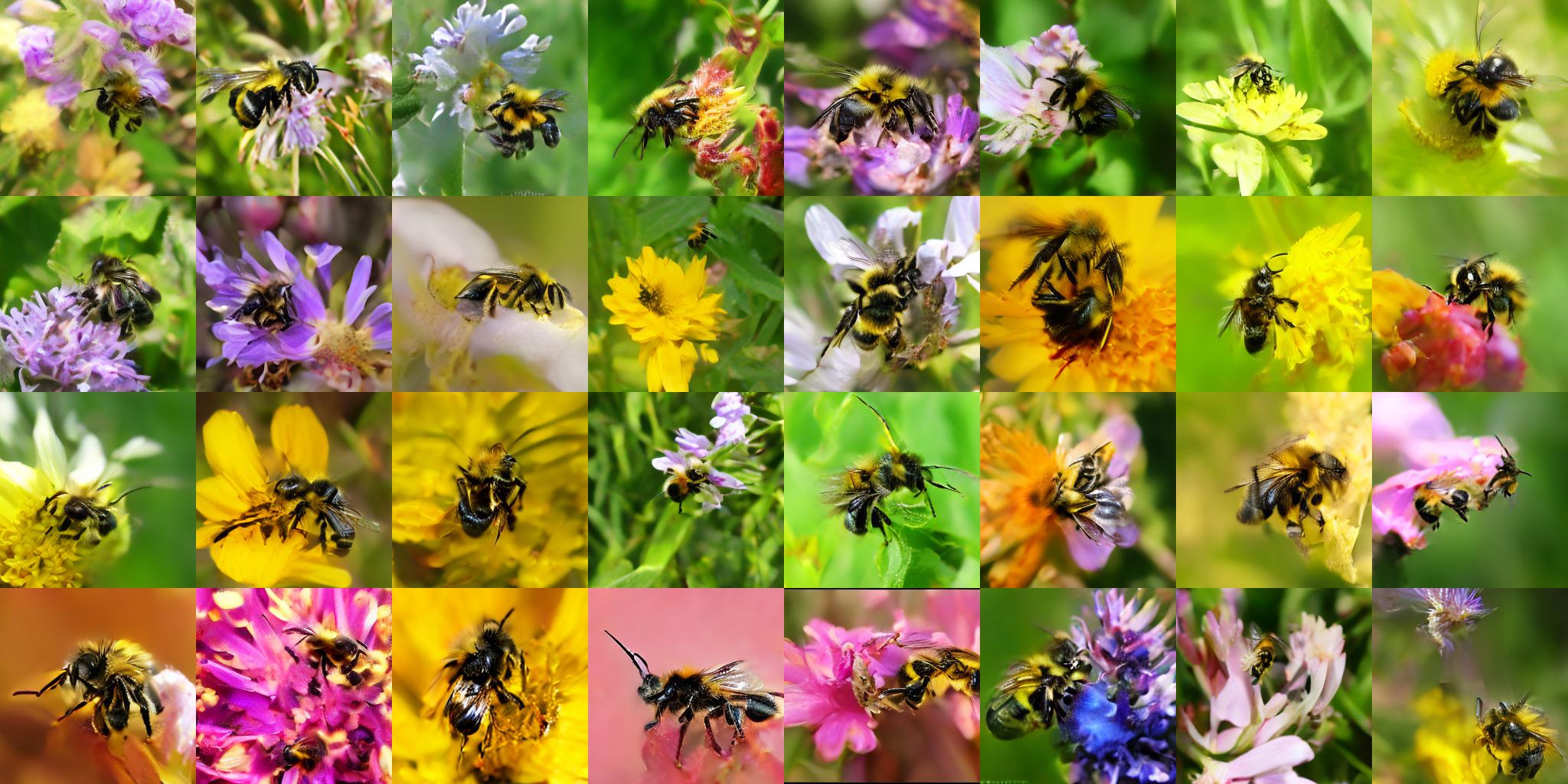}
        {\scriptsize class 309: bee}
        \vspace{1em}
    \end{minipage}
    \hfill
    \begin{minipage}[t]{0.46\linewidth}
        \centering
        \includegraphics[width=\textwidth]{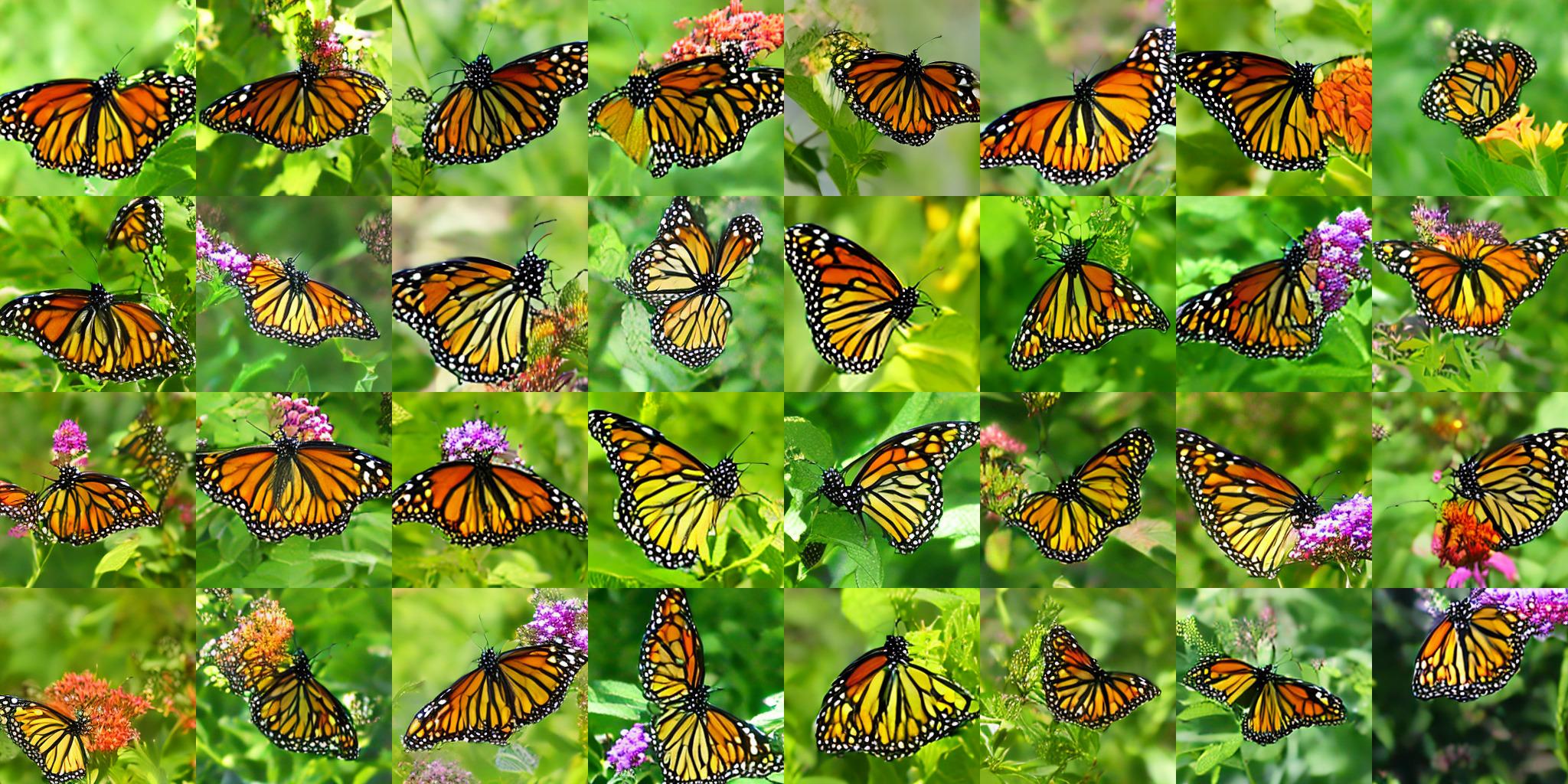}
        {\scriptsize class 323: monarch, monarch butterfly, milkweed butterfly, Danaus plexippus}
        \vspace{1em}
    \end{minipage}

    \begin{minipage}[t]{0.46\linewidth}
        \centering
        \includegraphics[width=\textwidth]{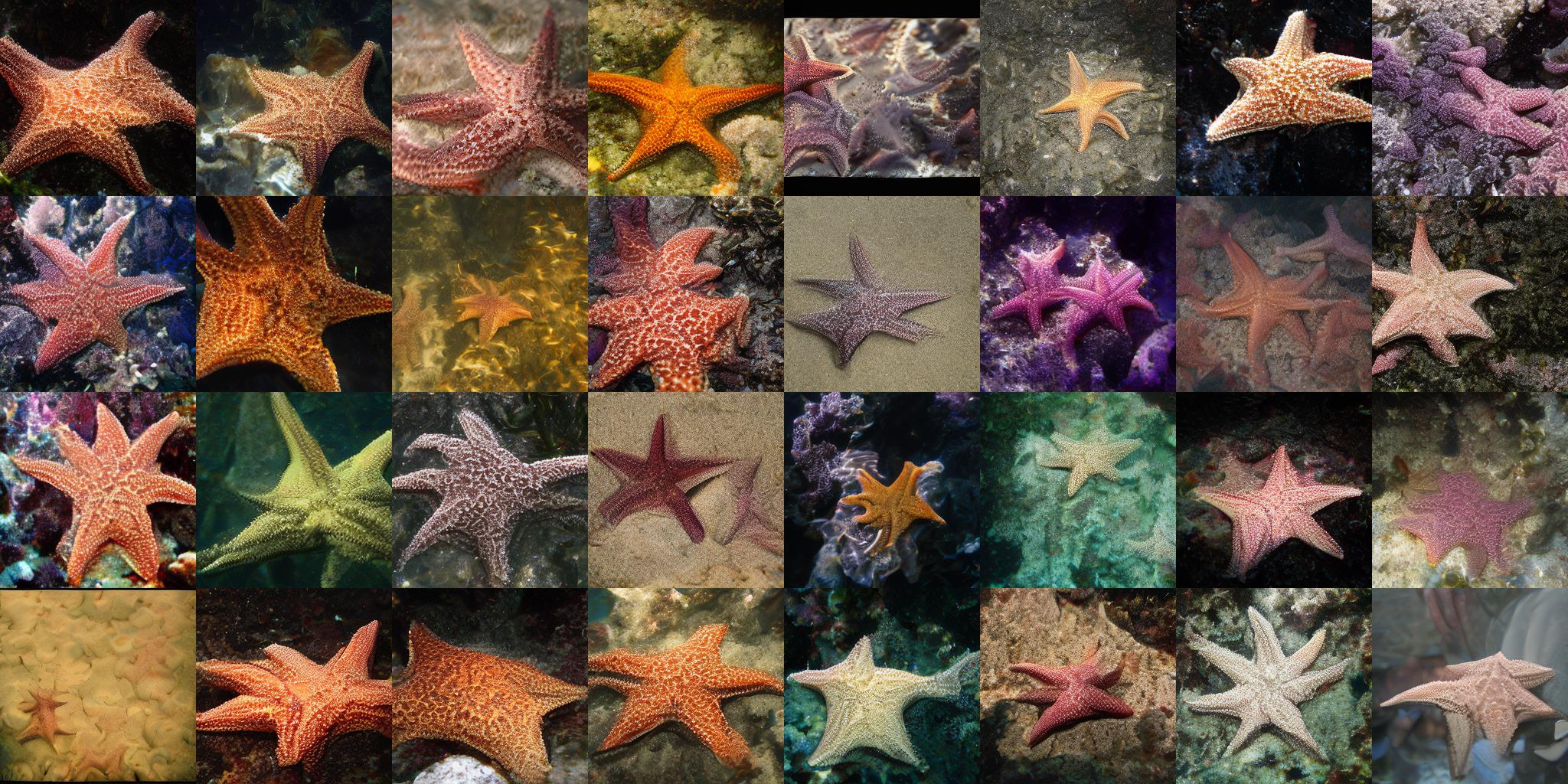}
        {\scriptsize class 327: starfish, sea star}
        \vspace{1em}
    \end{minipage}
    \hfill
    \begin{minipage}[t]{0.46\linewidth}
        \centering
        \includegraphics[width=\textwidth]{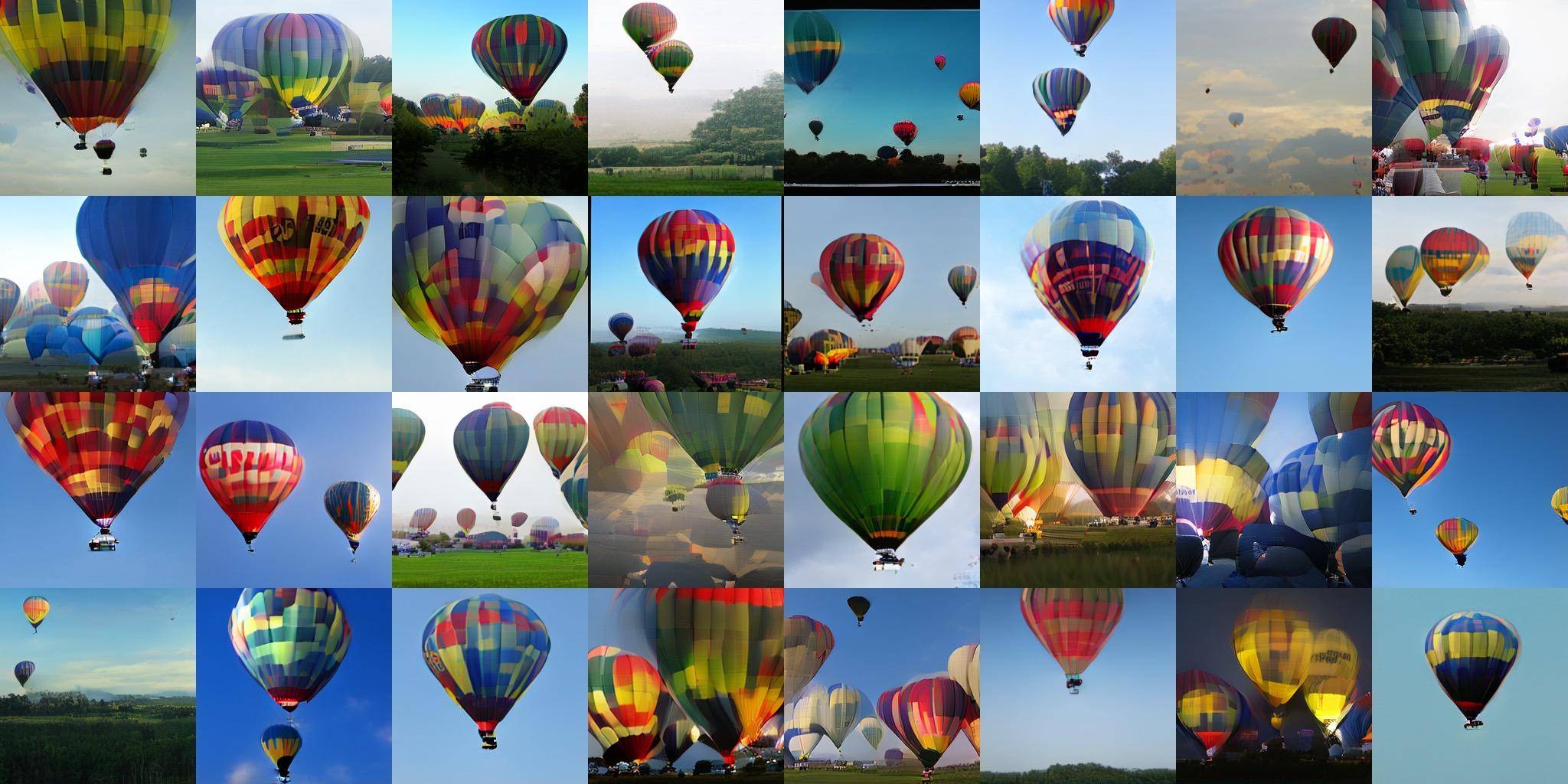}
        {\scriptsize class 417: balloon}
        \vspace{1em}
    \end{minipage}
    \caption{\emph{Uncurated} 1-NFE class-conditional generation samples of BiFlow-B/2 on ImageNet 256$\times$256. CFG scale: 2.0}
    \label{fig:uncurated_1}
\end{figure*}

\begin{figure*}[t]
    \centering

    \begin{minipage}[t]{0.46\linewidth}
        \centering
        \includegraphics[width=\textwidth]{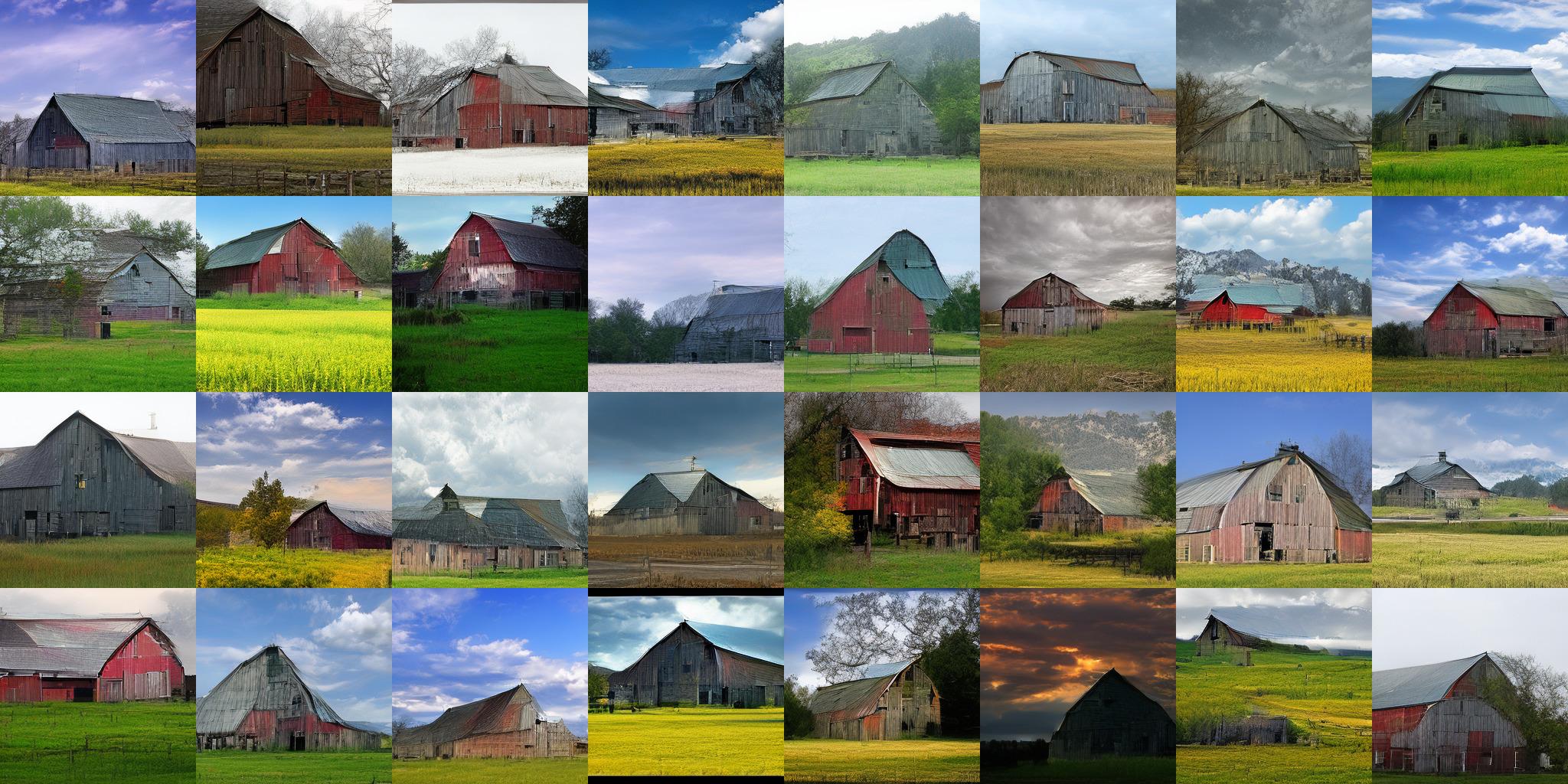}
        {\scriptsize class 425: barn}
        \vspace{1em}
    \end{minipage}
    \hfill
    \begin{minipage}[t]{0.46\linewidth}
        \centering
        \includegraphics[width=\textwidth]{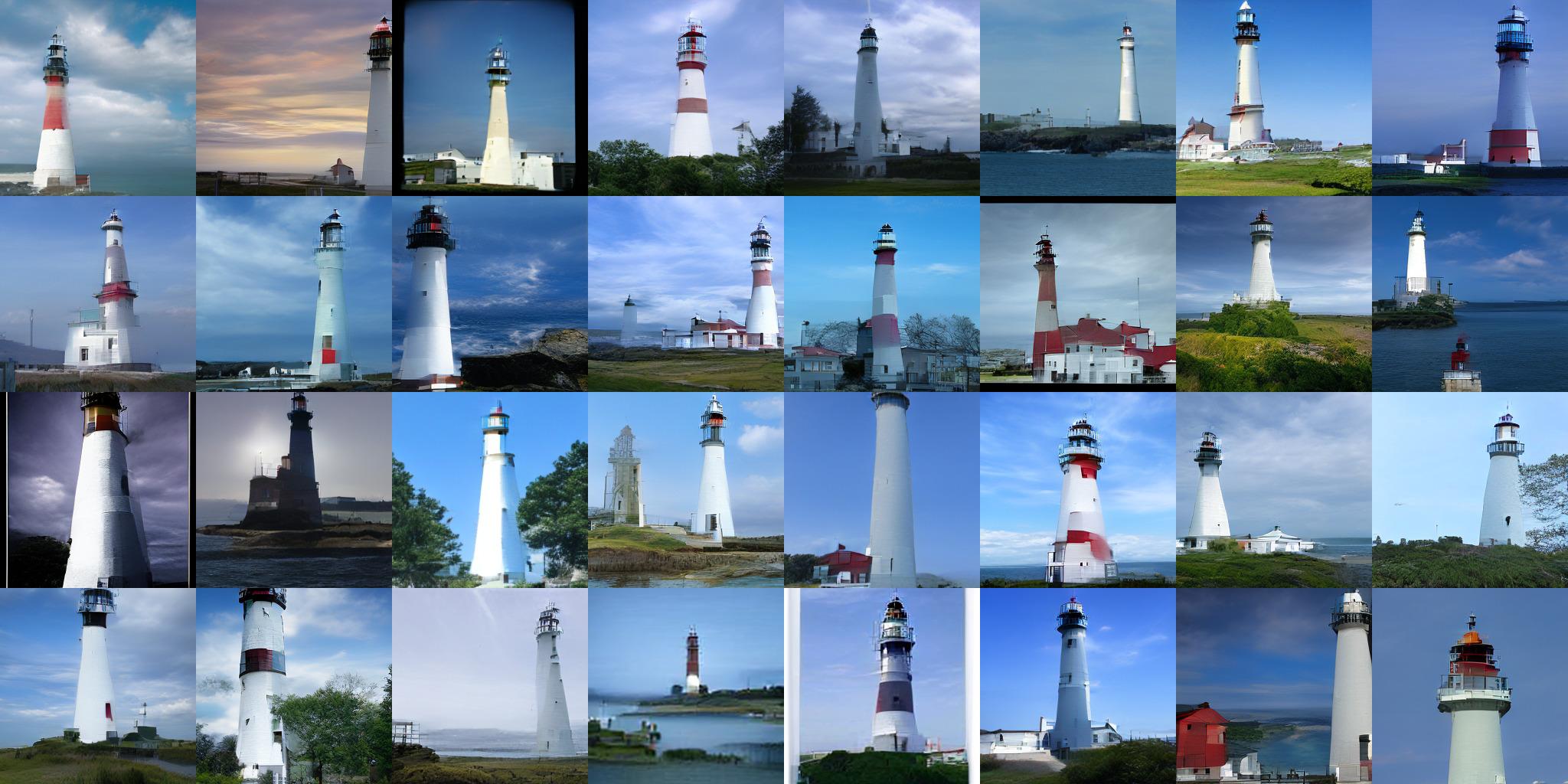}
        {\scriptsize class 437: beacon, lighthouse, beacon light, pharos}
        \vspace{1em}
    \end{minipage}

    \begin{minipage}[t]{0.46\linewidth}
        \centering
        \includegraphics[width=\textwidth]{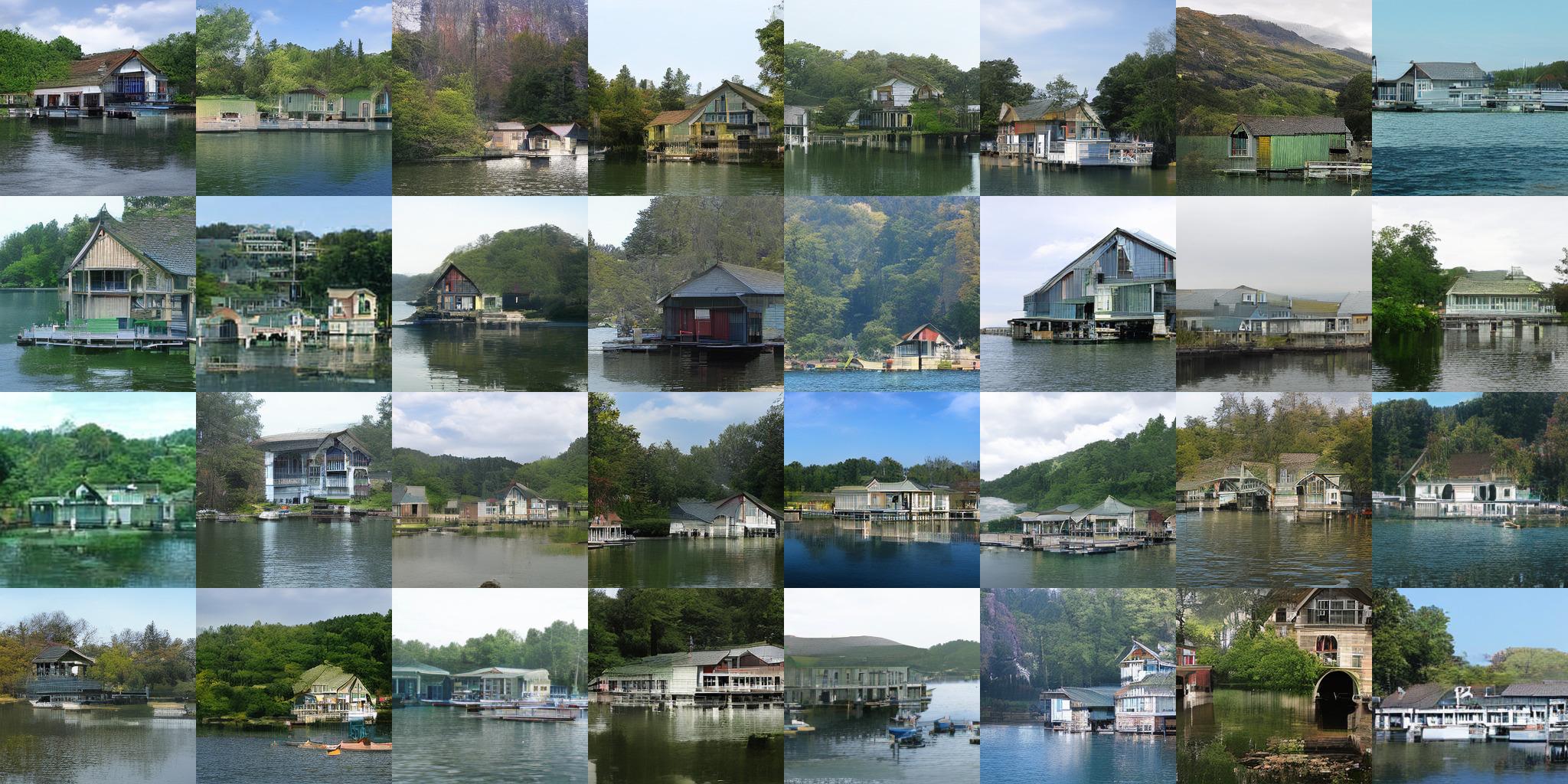}
        {\scriptsize class 449: boathouse}
        \vspace{1em}
    \end{minipage}
    \hfill
    \begin{minipage}[t]{0.46\linewidth}
        \centering
        \includegraphics[width=\textwidth]{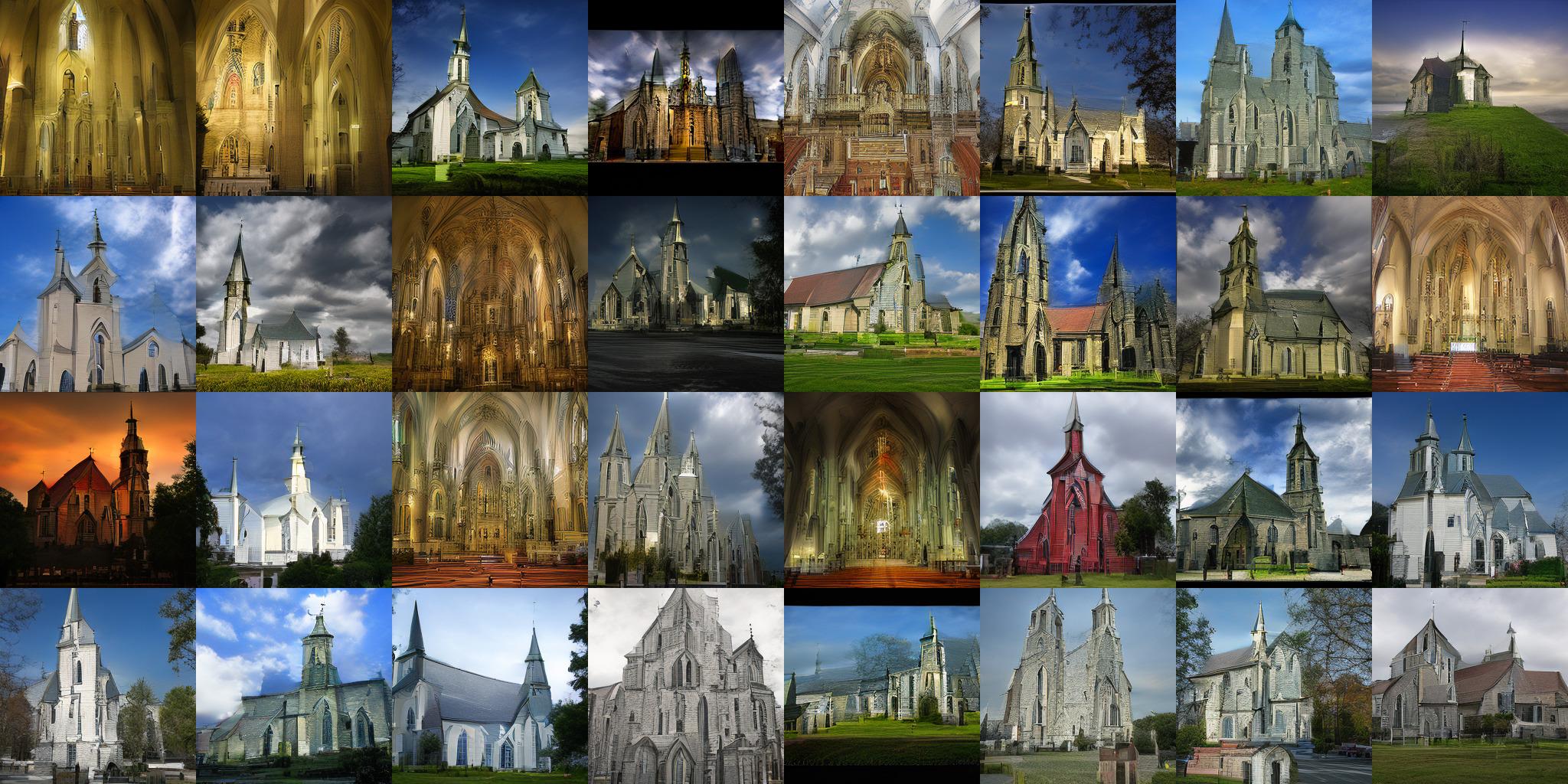}
        {\scriptsize class 497: church, church building}
        \vspace{1em}
    \end{minipage}

    \begin{minipage}[t]{0.46\linewidth}
        \centering
        \includegraphics[width=\textwidth]{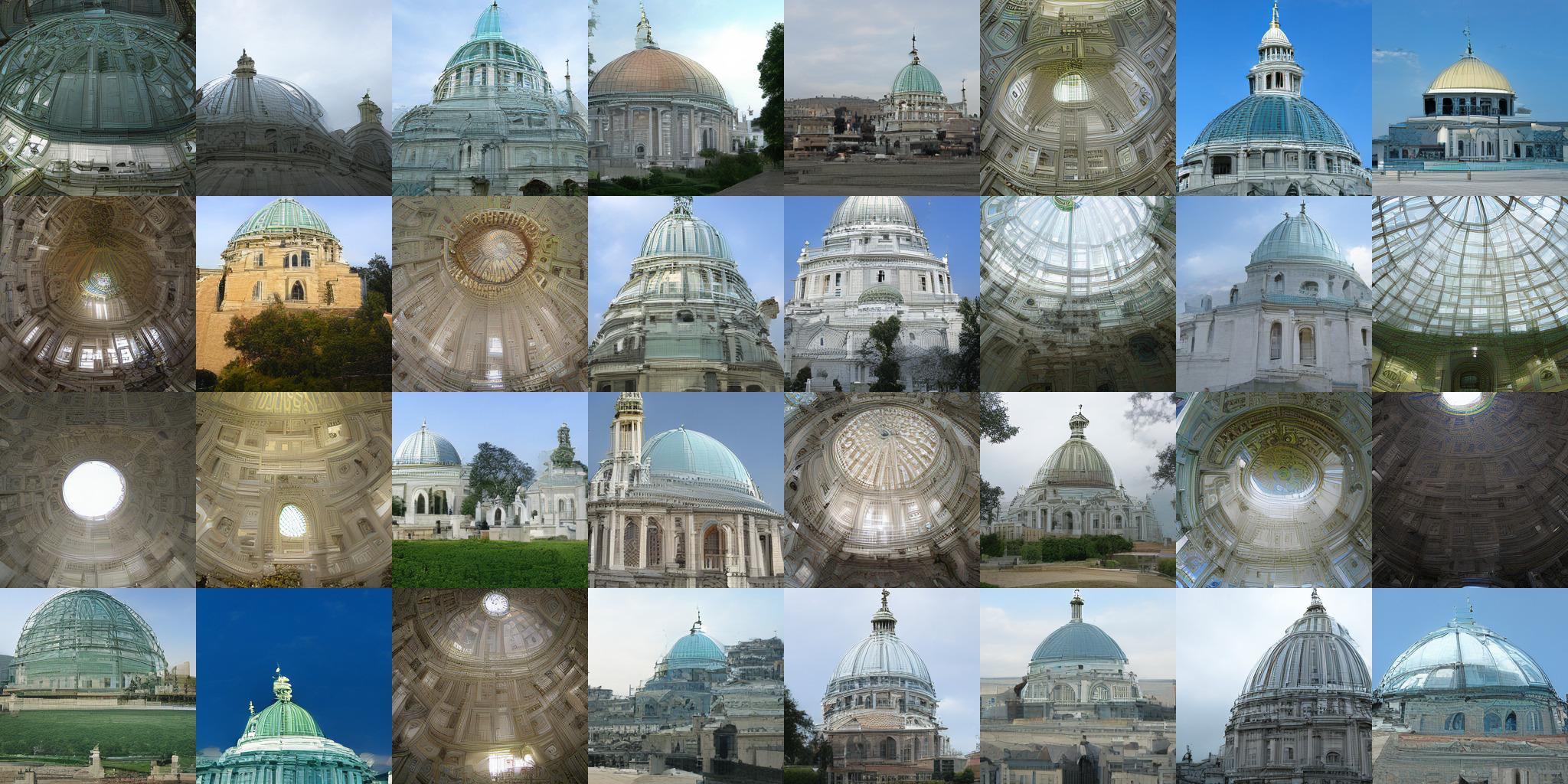}
        {\scriptsize class 538: dome}
        \vspace{1em}
    \end{minipage}
    \hfill
    \begin{minipage}[t]{0.46\linewidth}
        \centering
        \includegraphics[width=\textwidth]{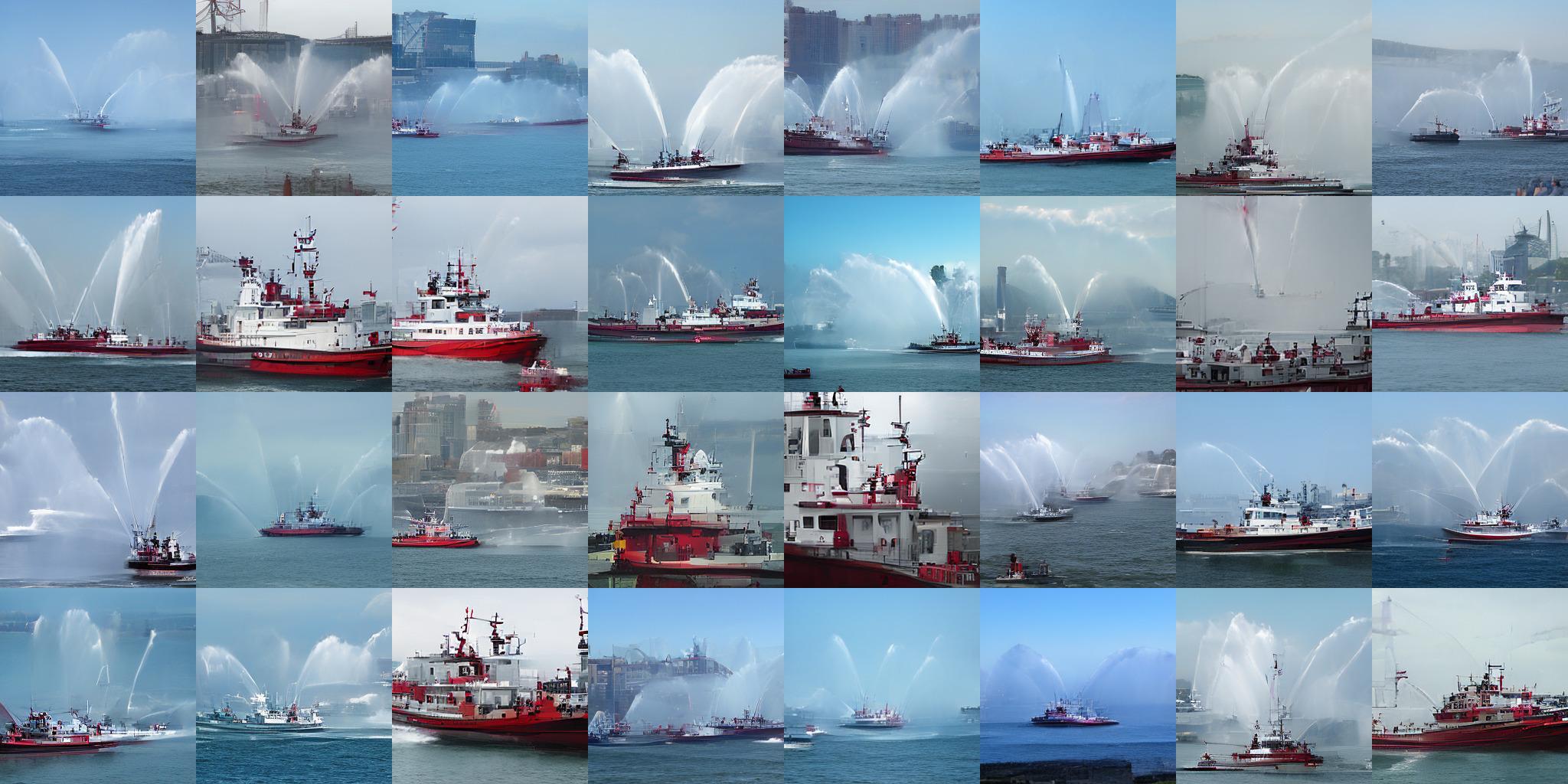}
        {\scriptsize class 554: fireboat}
        \vspace{1em}
    \end{minipage}
    
    \begin{minipage}[t]{0.46\linewidth}
        \centering
        \includegraphics[width=\textwidth]{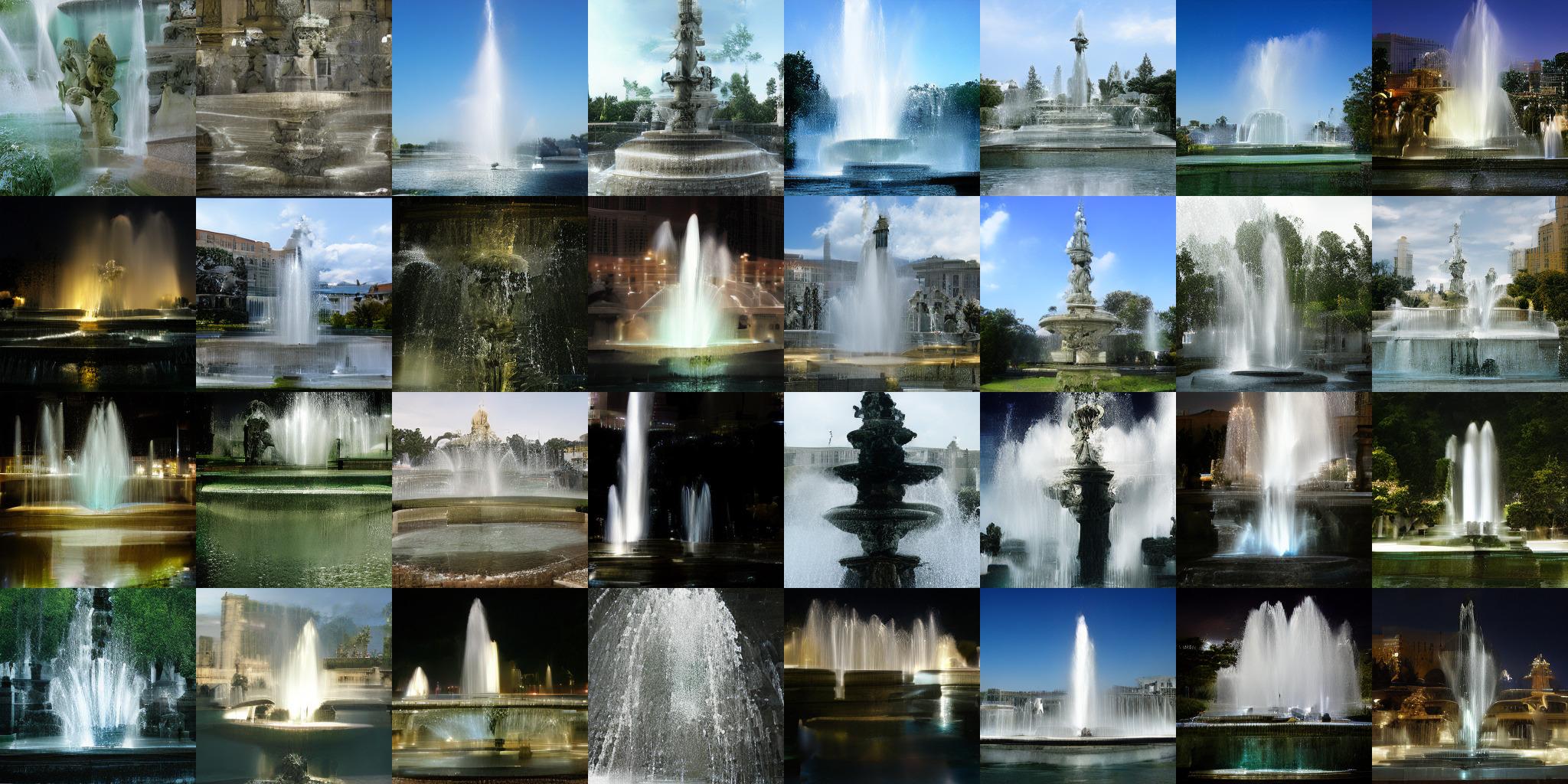}
        {\scriptsize class 562: fountain}
        \vspace{1em}
    \end{minipage}
    \hfill
    \begin{minipage}[t]{0.46\linewidth}
        \centering
        \includegraphics[width=\textwidth]{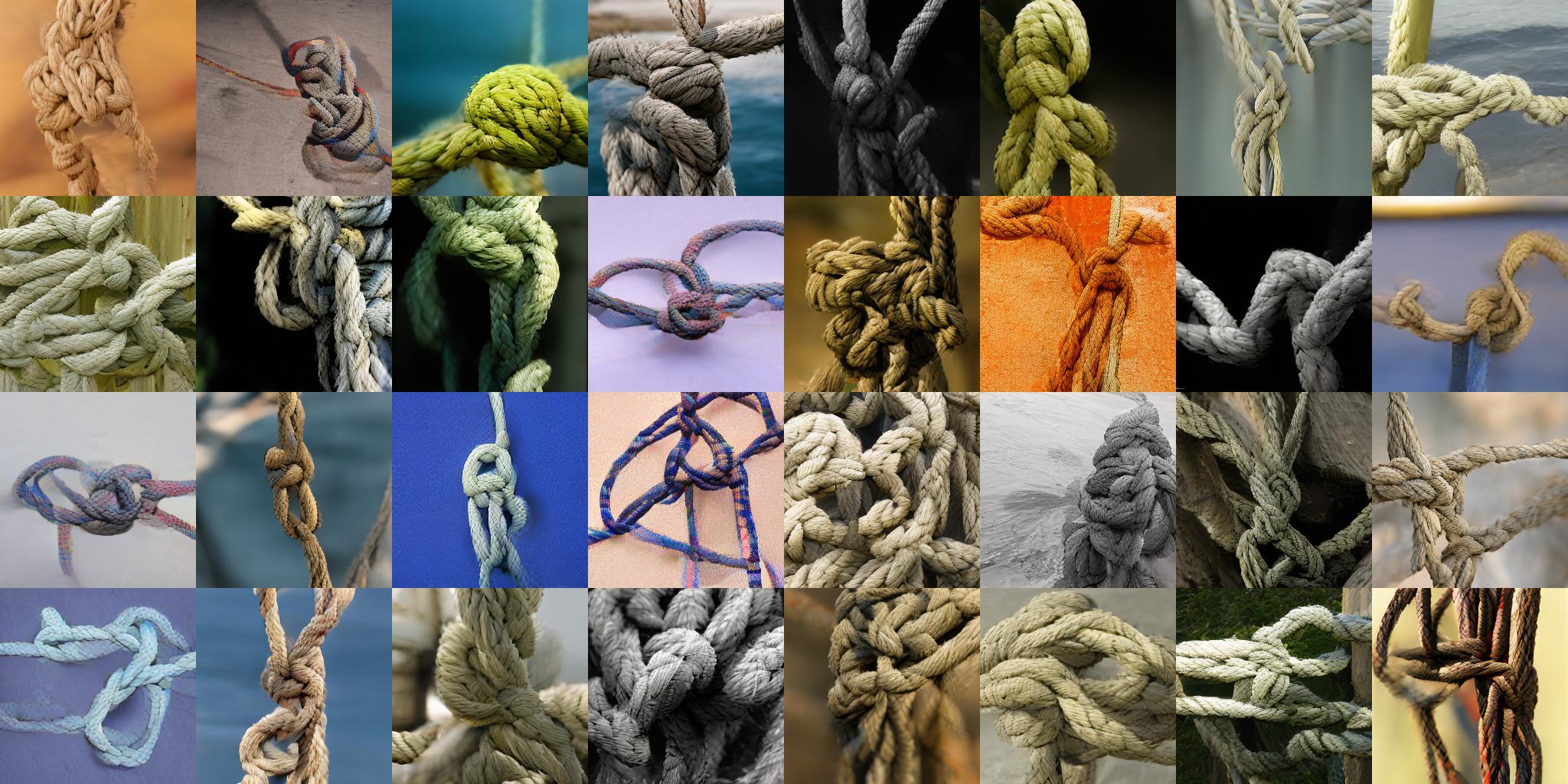}
        {\scriptsize class 616: knot}
        \vspace{1em}
    \end{minipage}
    
    \caption{\emph{Uncurated} 1-NFE class-conditional generation samples of BiFlow-B/2 on ImageNet 256$\times$256. CFG scale: 2.0}
    \label{fig:uncurated_2}
\end{figure*}

\begin{figure*}[t]
    \centering

    \begin{minipage}[t]{0.46\linewidth}
        \centering
        \includegraphics[width=\textwidth]{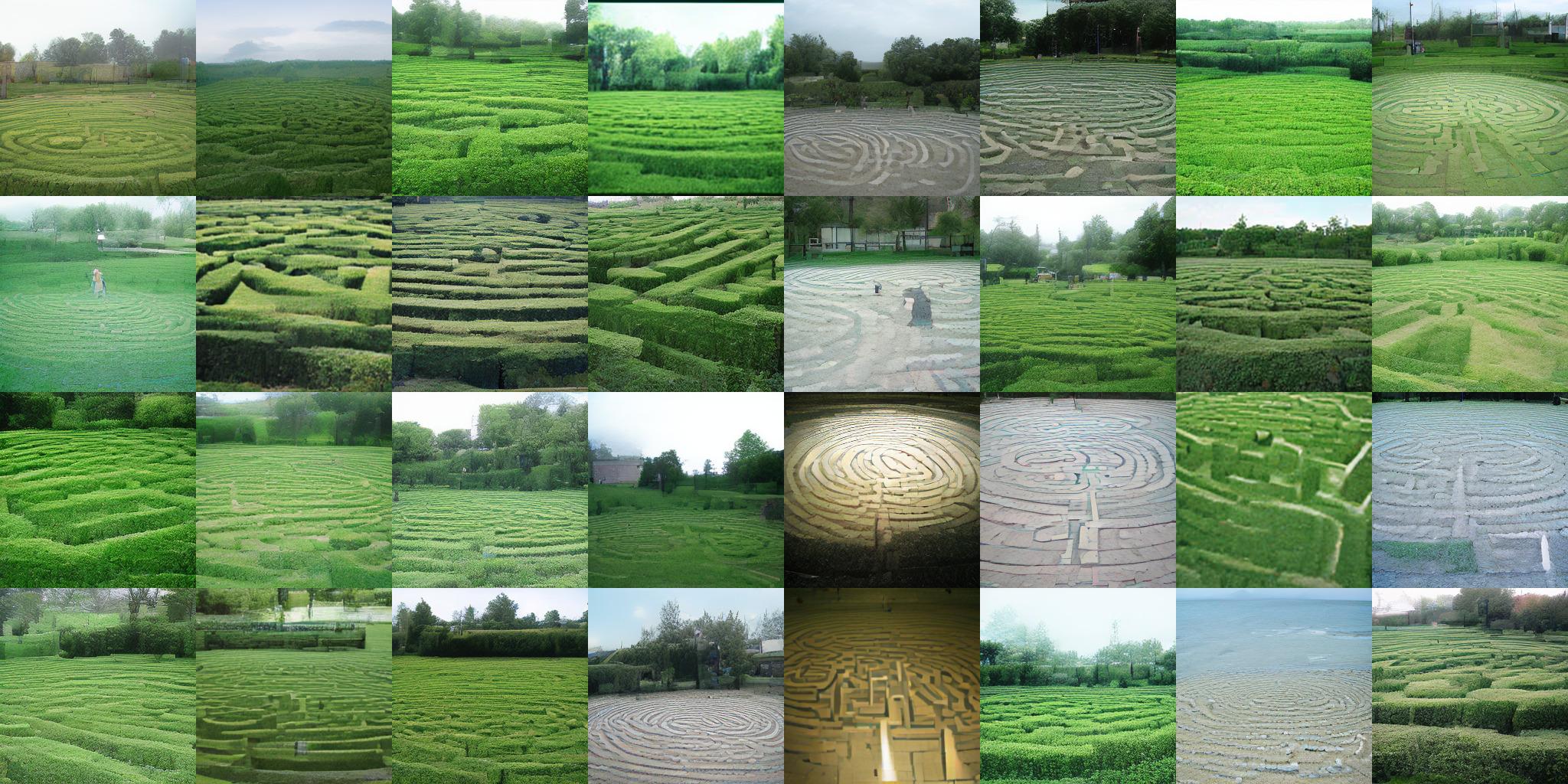}
        {\scriptsize class 646: maze, labyrinth}
        \vspace{1em}
    \end{minipage}
    \hfill
    \begin{minipage}[t]{0.46\linewidth}
        \centering
        \includegraphics[width=\textwidth]{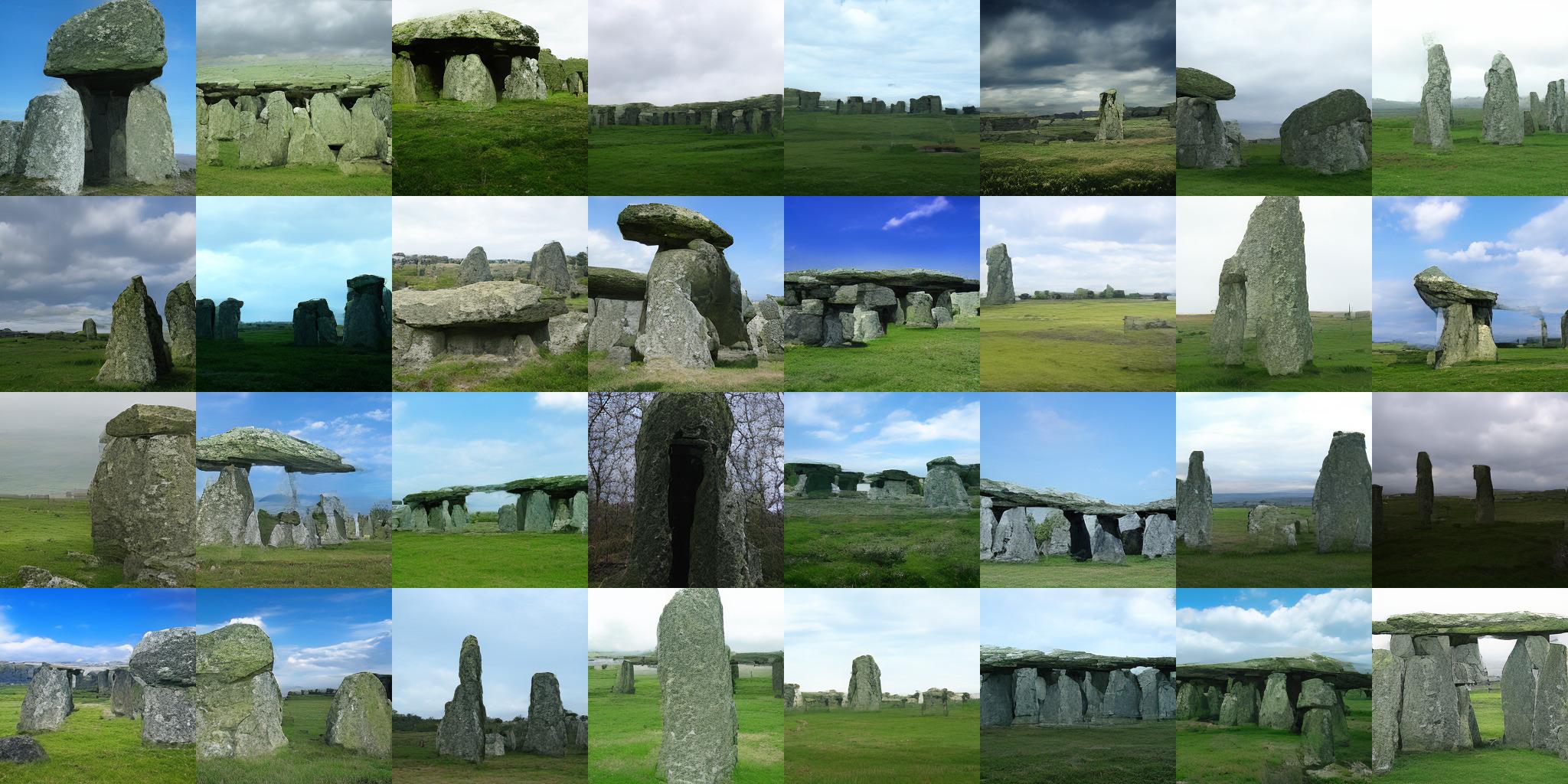}
        {\scriptsize class 649: megalith, megalithic structure}
        \vspace{1em}
    \end{minipage}

    \begin{minipage}[t]{0.46\linewidth}
        \centering
        \includegraphics[width=\textwidth]{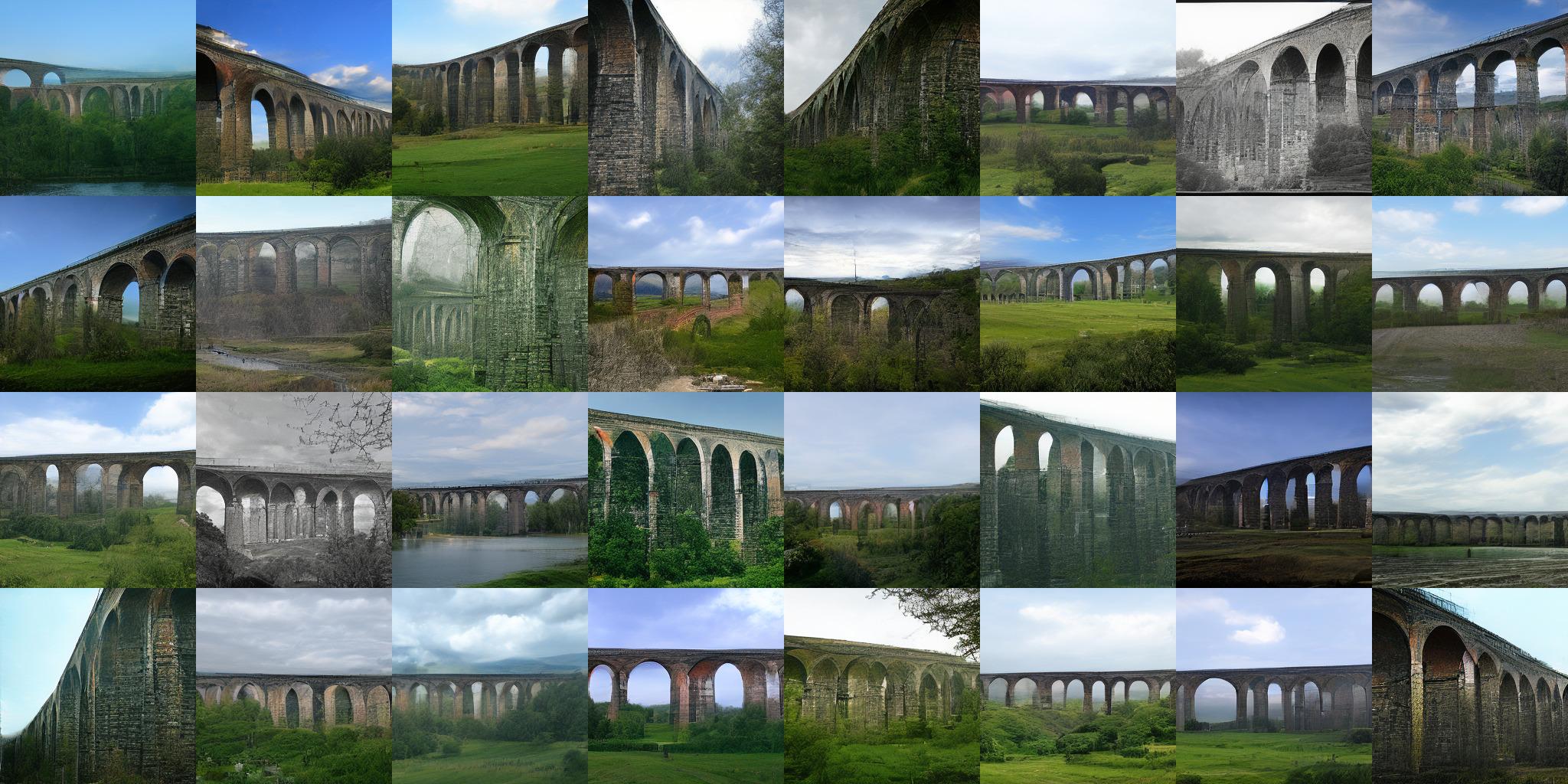}
        {\scriptsize class 888: viaduct}
        \vspace{1em}
    \end{minipage}
    \hfill
    \begin{minipage}[t]{0.46\linewidth}
        \centering
        \includegraphics[width=\textwidth]{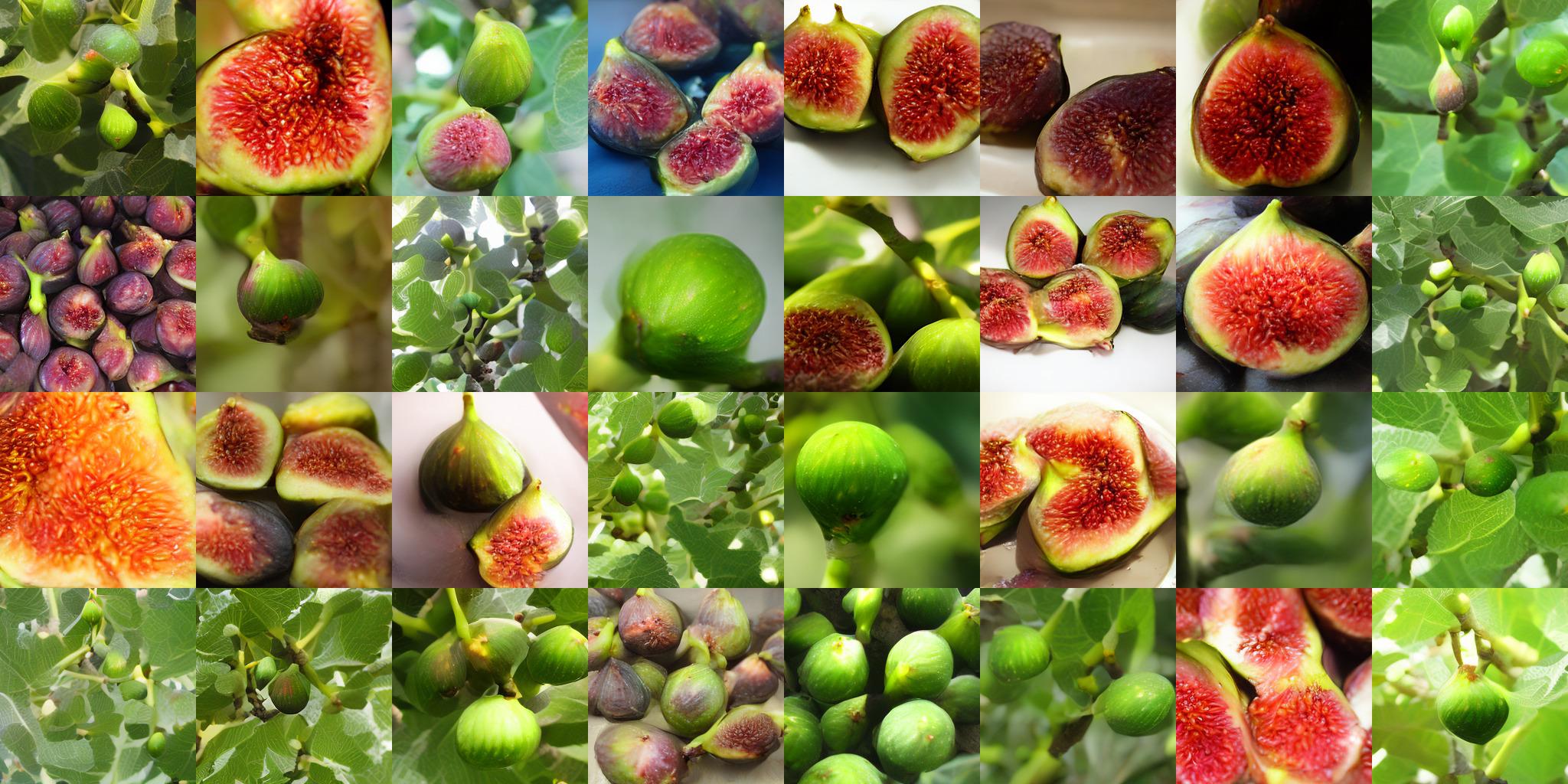}
        {\scriptsize class 952: fig}
        \vspace{1em}
    \end{minipage}

    \begin{minipage}[t]{0.46\linewidth}
        \centering
        \includegraphics[width=\textwidth]{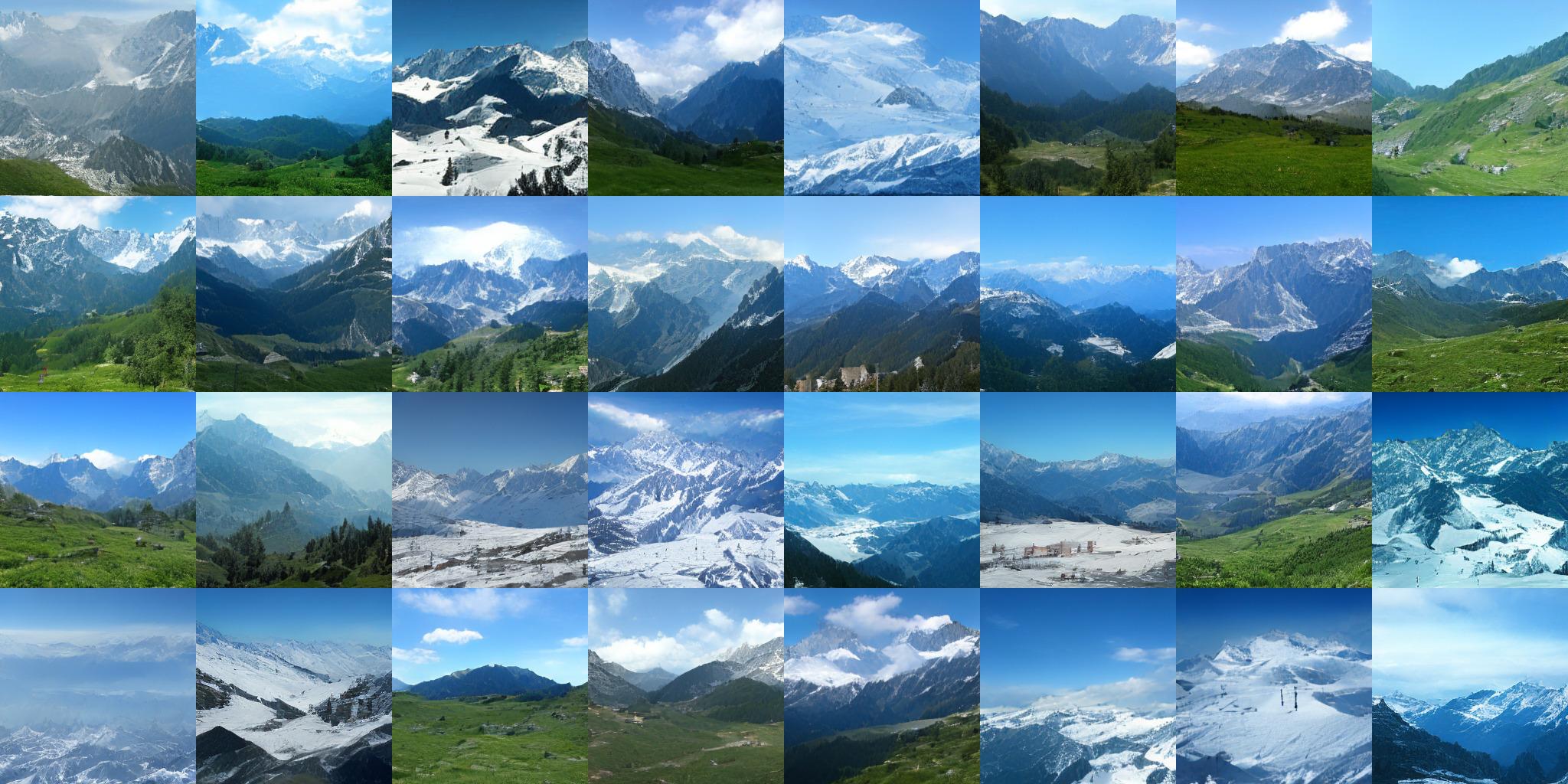}
        {\scriptsize class 970: alp}
        \vspace{1em}
    \end{minipage}
    \hfill
    \begin{minipage}[t]{0.46\linewidth}
        \centering
        \includegraphics[width=\textwidth]{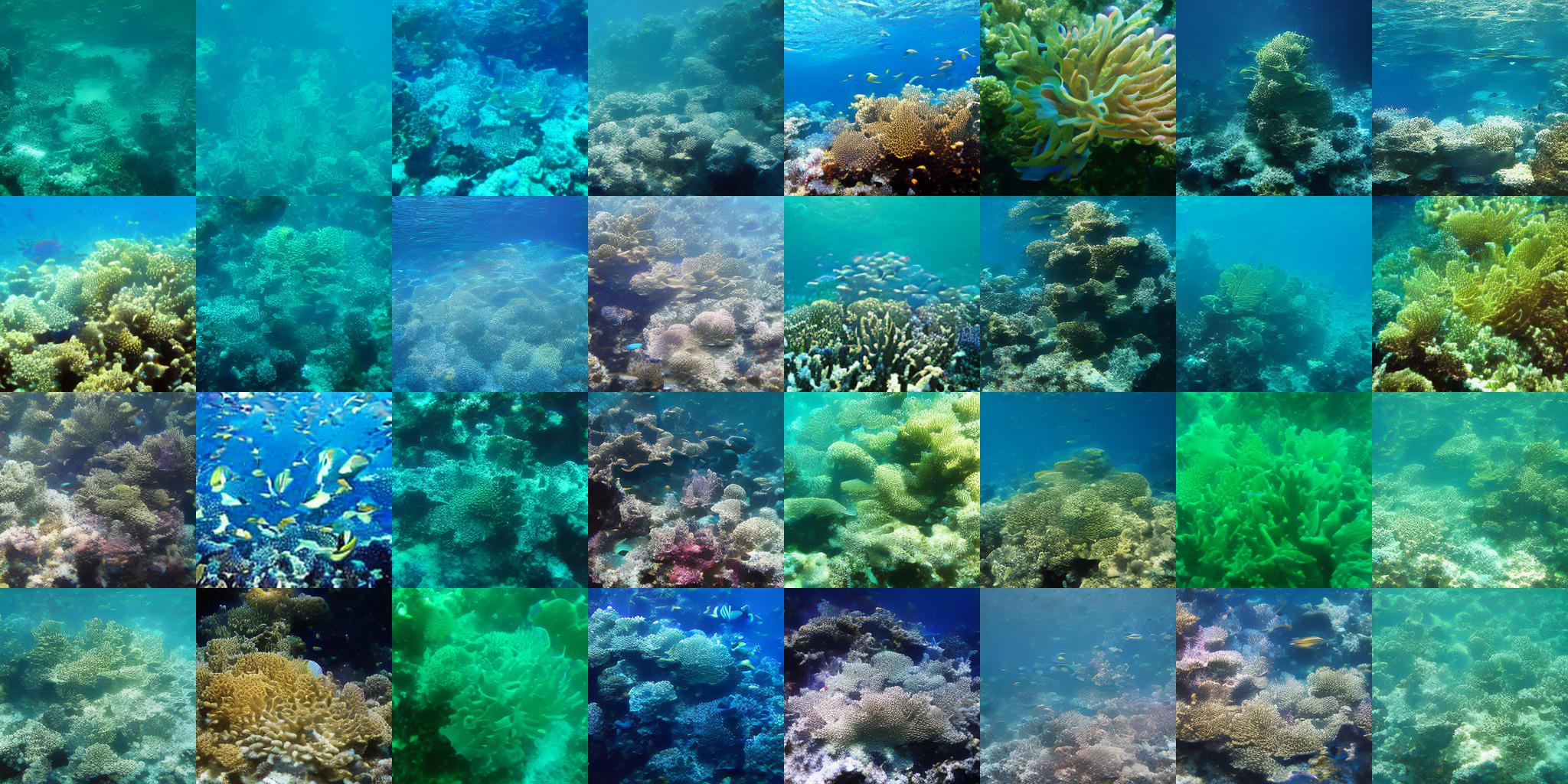}
        {\scriptsize class 973: coral reef}
        \vspace{1em}
    \end{minipage}

    \begin{minipage}[t]{0.46\linewidth}
        \centering
        \includegraphics[width=\textwidth]{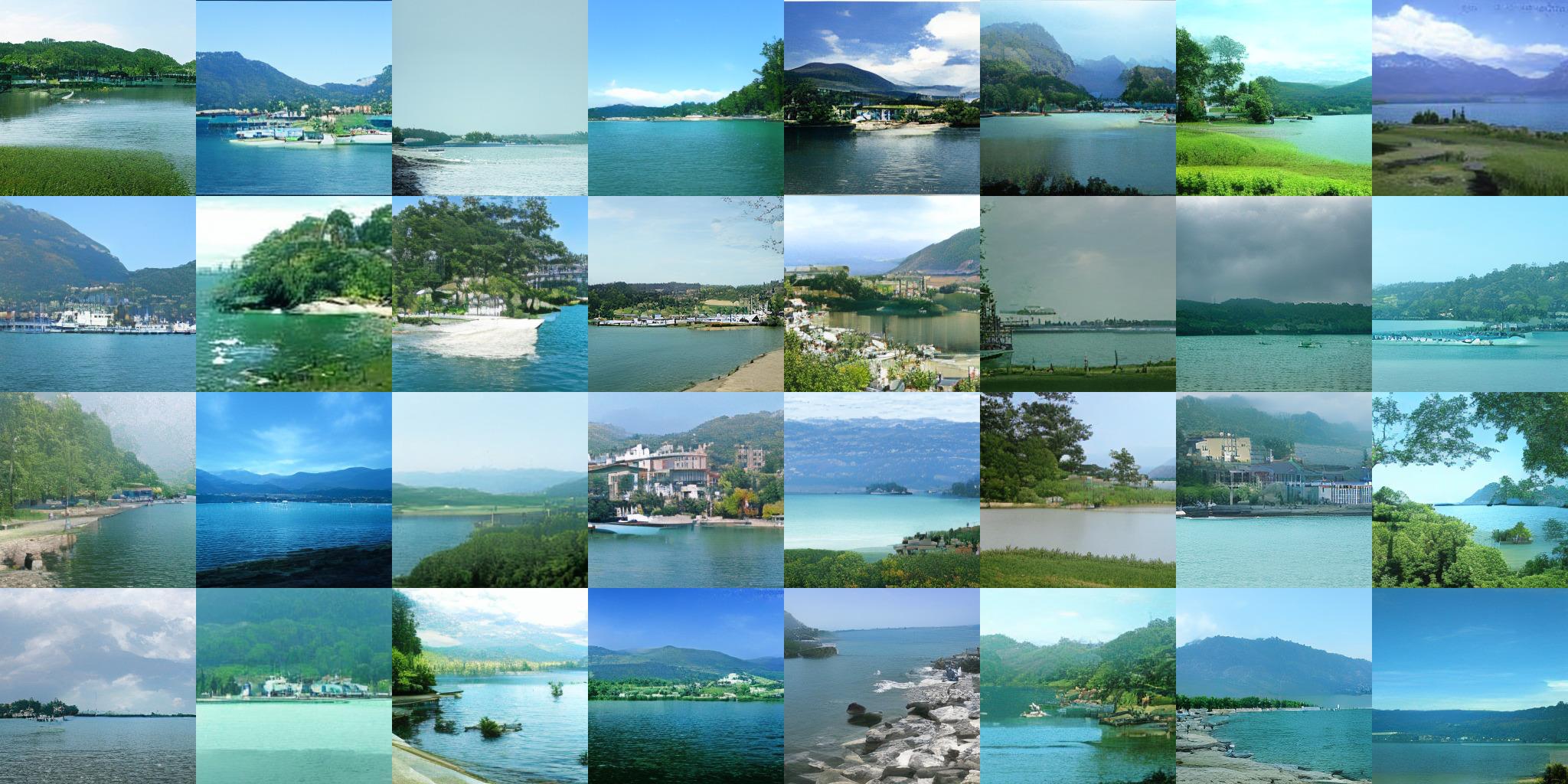}
        {\scriptsize class 975: lakeside, lakeshore}
        \vspace{1em}
    \end{minipage}
    \hfill
    \begin{minipage}[t]{0.46\linewidth}
        \centering
        \includegraphics[width=\textwidth]{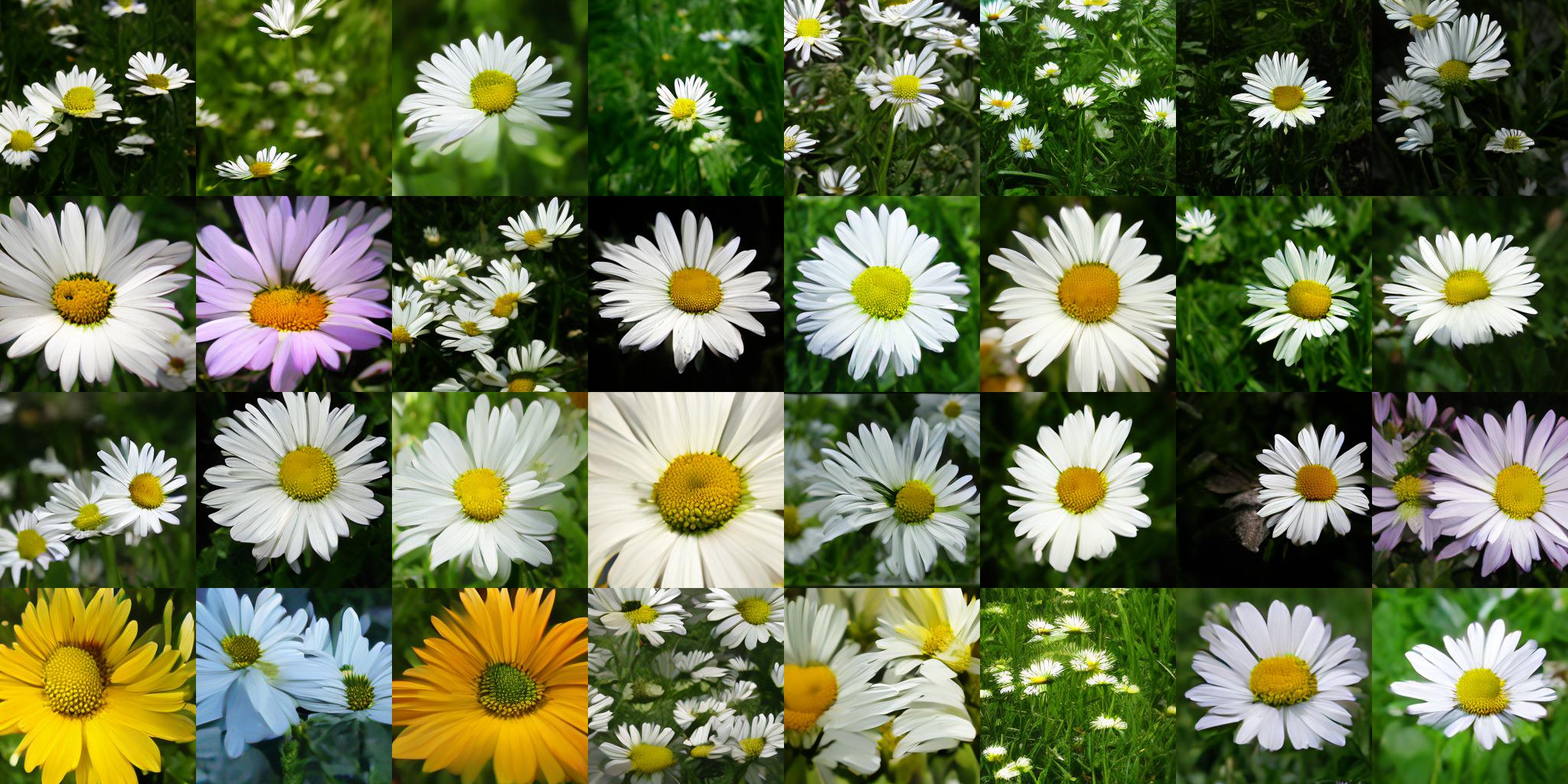}
        {\scriptsize class 985: daisy}
        \vspace{1em}
    \end{minipage}
    \caption{\emph{Uncurated} 1-NFE class-conditional generation samples of BiFlow-B/2 on ImageNet 256$\times$256. CFG scale: 2.0}
    \label{fig:uncurated_3}
\end{figure*}

We provide \textit{uncurated} 1-NFE generation results of BiFlow-B/2 in \cref{fig:uncurated_1} to \cref{fig:uncurated_3}.

\vspace{1em}

\paragraph{Acknowledgements.} 
We greatly thank Google TPU Research Cloud (TRC) for granting us access to TPUs. Q. Sun, X. Wang, Z. Jiang, and H. Zhao are supported by the MIT Undergraduate Research Opportunities Program (UROP). We thank Zhengyang Geng, Tianhong Li, and our other group members for their helpful discussions and feedback on the draft.

\newpage
{
    \small
    \bibliographystyle{ieee_fullname}
    \bibliography{biflow}
}

\newpage

\end{document}